\pdfoutput=1
\documentclass[twoside]{article}
\usepackage[preprint]{aistats2027}

\usepackage[round]{natbib}

\usepackage{amsmath,amssymb,amsthm,mathtools}
\usepackage{bm}
\usepackage{booktabs}
\usepackage{multirow}
\usepackage{graphicx}
\usepackage{xcolor}
\usepackage[hidelinks]{hyperref}
\usepackage[capitalize,noabbrev]{cleveref}
\usepackage{algorithm}
\usepackage[noend]{algorithmic}
\usepackage{enumitem}
\usepackage{subcaption}
\usepackage{placeins}

\newtheorem{theorem}{Theorem}
\newtheorem{lemma}[theorem]{Lemma}
\newtheorem{proposition}[theorem]{Proposition}
\newtheorem{corollary}[theorem]{Corollary}
\theoremstyle{definition}

\newtheorem{assumption}{Assumption}
\theoremstyle{remark}
\newtheorem{remark}{Remark}

\newcommand{\R}{\mathbb{R}}
\newcommand{\E}{\mathbb{E}}
\renewcommand{\P}{\mathbb{P}}
\newcommand{\1}{\mathbf{1}}
\newcommand{\sal}{s}           
\newcommand{\Ptil}{\widetilde{P}}
\newcommand{\err}{\mathrm{err}}
\newcommand{\WGA}{\mathrm{WGA}}
\newcommand{\sgn}{\operatorname{sign}}
\newcommand{\peel}{\textsc{Peel}}

\newcommand{\Mmarg}{M}

\begin{document}

\runningtitle{Peel, Don't Whack: Intervention-Based Shortcut Mitigation}

\twocolumn[
\aistatstitle{Peel, Don't Whack: Partial Coverage, Imperfect Validators,\\
and Intervention-Based Shortcut Mitigation}
\runningauthor{Ali Alavi}
\aistatsauthor{ Ali Alavi }
\aistatsaddress{ Department of Computer Science and Engineering \\
  The Ohio State University, Columbus, OH, USA \\
  \texttt{alavibajestan.1@osu.edu} }
 ]

\begin{abstract}
Without group labels, shortcut mitigation increasingly relies on \emph{interventions}: label-preserving edits, such as background or watermark replacement, that are proposed automatically and accepted by a validator. Invariance to a \emph{complete} family of valid edits yields robustness, but in practice coverage is partial and validators err.

We analyze both in a stylized linear model. For the population logistic solution and the max-margin limit of gradient descent, the decision on every shortcut-flip group follows one additive rule. Hence worst-group accuracy never decreases with coverage, while individual groups can provably degrade, a mechanism for the Whac-a-Mole effect; full coverage is not always needed. In the logistic regime the validator's false-accept rate must stay below a closed-form tolerance that grows with coverage. With noiseless training data, one false accept can make the max-margin solution ignore the core. In this model, sensitivity-based discovery removes shortcuts in decreasing salience. We also give a model-agnostic certificate, valid under content-independent validation, whose validator-error term comes from an audit.

Simulations match the theory, and deep networks on synthetic and real images show the predicted Whac-a-Mole signs, monotone worst-group accuracy and validator fragility. The resulting procedure, \peel{} (Propose--Edit--Evaluate--Learn), can propose candidate edits from the model's own mistakes. Given an oracle library of the benchmarks' intervention operators, it selects the right ones and beats group-label methods on Waterbirds and UrbanCars. With fully generated edits, run end to end from three discovery seeds, it improves on empirical risk minimization (ERM) on five of six benchmarks but trails exact edits.
\end{abstract}

\section{Introduction}
\label{sec:intro}

Neural networks readily exploit \emph{shortcuts}: features that predict the label in training but not under deployment shift \citep{geirhos2020shortcut}. Group distributionally robust optimization (GroupDRO; \citealp{sagawa2020distributionally}), reweighting as in Just Train Twice (JTT; \citealp{liu2021just}), and last-layer retraining by deep feature reweighting (DFR; \citealp{kirichenko2023last}) mitigate known shortcuts, but need group annotations for training or validation. Without such side information, shortcut identification is impossible in general: two causal models with the same training distribution can disagree on which feature is stable \citep{lin2022zin}.

A growing line of work supplies \emph{interventions}: label-preserving edits that change a suspected shortcut, such as replacing the background, removing a watermark, or rewriting an irrelevant phrase. These edits are increasingly produced by generative models and filtered by vision--language or language models \citep{dunlap2023diversify,wu2023discover,prabhu2023lance}. Invariance to a \emph{complete} family of such edits is well understood: it makes the predictor a function of the core alone, robust to any shift of the shortcut distribution \citep{veitch2021counterfactual,vonkugelgen2021self}.

Practice never meets these conditions, and two gaps matter. First, \textbf{coverage is partial}: real data contain several shortcuts, and any edit bank covers only some of them. \citet{li2023whac} documented a puzzling consequence, the \emph{Whac-a-Mole} effect, in which mitigating one shortcut amplifies reliance on another. We lack a theory of when it happens and what partial coverage does and does not buy. Second, \textbf{validators err}: an automatic validator sometimes accepts an edit that silently changed the label-relevant content, as documented for counterfactual augmentation in natural language processing (NLP) \citep{joshi2022investigation}. How such errors propagate into shortcut reliance has not been analyzed.

\paragraph{Contributions.}
\begin{itemize}[leftmargin=*,itemsep=1pt,topsep=2pt]
\item \textbf{Partial-coverage theory} (\cref{thm:partial}, \cref{cor:wam,cor:suff}). In a stylized linear model, for the population logistic solution and the max-margin limit of gradient descent (which is known to produce shortcuts; \citealp{puli2023dont}), we derive the exact decision rule on every shortcut-flip group as a function of the covered set. Worst-group accuracy never decreases with coverage, while specific groups can provably \emph{degrade} (the Whac-a-Mole effect, with a necessary and sufficient condition). Full coverage is unnecessary once the core outweighs the uncovered shortcuts.
\item \textbf{Validator-error theory} (\cref{thm:validator}). In the logistic regime we give a closed-form false-accept tolerance $\delta^\star$ that grows with coverage. If the training data are noiseless, a \emph{single} false accept can make the max-margin solution a pure shortcut classifier.
\item \textbf{Discovery and certification} (\cref{prop:greedy}, \cref{thm:cert}). In the stylized model, greedy, sensitivity-based discovery removes shortcuts in decreasing salience. We also give a finite-sample certificate for any model, relative to the discovered edit family and under content-independent validation, in which the validator's error is \emph{estimated} by an audit.
\item \textbf{\peel{} and MD-\peel{}.} We instantiate the theory as \peel{} (Propose--Edit--Evaluate--Learn), and as MD-\peel{} (mistake-driven \peel), which proposes candidate edits from the model's own mistakes. Given an oracle library of the benchmarks' intervention operators, MD-\peel{} selects the right ones from the model's mistakes alone and beats every group-label baseline on Waterbirds and UrbanCars. With fully generated edits, run end to end, it is weaker.
\end{itemize}

\section{Setup: Shortcuts, Interventions, and Validators}
\label{sec:setup}

\paragraph{Data.} Labels $Y\in\{\pm1\}$ are balanced. An input carries a \emph{core} factor and $k$ \emph{shortcut} factors. In the stylized model used for our theory,
\begin{equation}
X=\big(a\,Z_cY,\; b_1Z_1Y,\;\dots,\; b_kZ_kY\big)\in\R^{1+k},
\label{eq:model}
\end{equation}
where $a,b_j>0$ are feature magnitudes and $Z_c,Z_1,\dots,Z_k\in\{\pm1\}$ are mutually independent given $Y$, with $\P(Z_c=1)=p$ and $\P(Z_j=1)=\rho_j$. Shortcut $j$ is \emph{aligned} with the label when $Z_j=1$. In training $\rho_j>\tfrac12$, so every shortcut is predictive. Following the anti-causal view of spurious features \citep{veitch2021counterfactual,makar2022causally}, the label causes each shortcut $X_j$. Under deployment shift $\rho_j$ may change arbitrarily, including reversing. When $p<1$ the shortcuts carry information about $Y$ beyond the core $X_c$, so robustness has an in-distribution price. We make this explicit rather than assuming the shortcuts are conditionally independent of $Y$ given the core.

\paragraph{Flip groups.} For $F\subseteq[k]$, the \emph{flip group} $G_F$ consists of inputs whose shortcuts in $F$ disagree with the label ($Z_j=-1$ for $j\in F$), whose remaining shortcuts agree, and whose core is intact ($Z_c=1$). Standard worst-group benchmarks are unions of flip groups. The worst-group accuracy ($\WGA$) is the minimum accuracy over all $2^k$ flip groups.

\paragraph{Interventions and coverage.} An \emph{edit} is a label-preserving transformation of an input. Edit type $t$ \emph{randomizes} shortcut $j$ if it replaces $Z_j$ by an independent uniform sign and leaves everything else unchanged. Given a bank of edit types, the \emph{covered set} $K\subseteq[k]$ contains the shortcuts randomized by some bank edit, and $U=[k]\setminus K$ is the uncovered set. For the theory we idealize training on edited data as minimizing the loss under $\Ptil_{K,\delta}$. This is the law of $X$ after every $Z_j$ with $j\in K$ is jointly replaced by an independent uniform sign and, independently of $(Y,Z)$, $Z_c$ is flipped with probability $\delta$ (see \emph{Validators}). In other words, every training input is an accepted edit that randomizes all covered shortcuts, for example a composition of bank edits. If unedited inputs are mixed in, or each edit randomizes a single shortcut, the covered weights shrink but do not vanish in the logistic regime. The max-margin results depend only on the training support and are unaffected.

\paragraph{Validators.} Edits are proposed automatically, for example by a generative editor, and accepted by a \emph{validator} that judges whether the label-relevant content was preserved. A \emph{false accept} is an accepted edit that in fact altered the core while the original label was kept. In $\Ptil_{K,\delta}$ this flips $Z_c$, and $\delta$ is the validator's false-accept rate among accepted edits.

\paragraph{Salience.} With $\lambda(u)=\log\frac{u}{1-u}$, we define the salience of the core and of shortcut $j$ by
\begin{equation}
\begin{aligned}
&\text{logistic:} && \sal_c=\lambda(p),\quad \sal_j=\lambda(\rho_j);\\
&\text{max-margin:} && \sal_c=a^2,\quad\ \ \sal_j=b_j^2.
\end{aligned}
\label{eq:salience}
\end{equation}
We define the margin of $G_F$, in salience units, as
$\Mmarg_K(F)=\sal_c+\sum_{j\in U\setminus F}\sal_j-\sum_{j\in U\cap F}\sal_j$.
Indices of $F$ that are covered do not enter it. A margin of exactly $0$ counts as an error. We write $\err_Q(f)=\P_Q(f(X)\neq Y)$.

\section{Theory}
\label{sec:theory}

Except for \cref{thm:cert}, all results concern homogeneous linear predictors $x\mapsto\sgn(w^\top x)$ on the stylized model \eqref{eq:model} trained on $\Ptil_{K,\delta}$, in two regimes. The \emph{logistic regime} is the infinite-sample, well-specified limit of unregularized logistic regression, for $p,\rho_j\in(\tfrac12,1)$. The \emph{max-margin regime} applies to noiseless data ($p=1$, $\rho_j=1$ for $j\in U$). There, gradient descent on logistic loss converges in direction to the $\ell_2$ max-margin separator \citep{soudry2018implicit}, which relies on shortcuts even when the core is perfectly predictive \citep{nagarajan2021understanding,puli2023dont}.
Proofs: \cref{app:proofs}.

\subsection{Partial coverage and the Whac-a-Mole effect}

\begin{theorem}[Group-wise decision rule under partial coverage]
\label{thm:partial}
Let $\delta=0$.
\begin{enumerate}[label=(\roman*),leftmargin=*,itemsep=0pt,topsep=2pt]
\item \emph{Logistic regime.} The logistic risk under $\Ptil_{K,0}$ has a unique minimizer $w^\star$, with $w^\star_c=\lambda(p)/a$, $w^\star_j=0$ for $j\in K$, and $w^\star_j=\lambda(\rho_j)/b_j$ for $j\in U$.
\item \emph{Max-margin regime.} Suppose $0\in\mathrm{conv}\{(b_jz_{ij})_{j\in K}\}_i$, where $z_i$ ranges over the $K$-patterns of the training points. For instance, some pattern and its negation both occur. For independent and identically distributed (i.i.d.) uniform patterns this holds with high probability once $n\gtrsim4|K|$. Then the max-margin direction is $w^\star\propto(a,\,0_K,\,(b_j)_{j\in U})$.
\end{enumerate}
In both regimes a point of flip group $G_F$ is classified correctly if and only if (iff) $\Mmarg_K(F)>0$. Covered shortcuts never affect the decision.
\end{theorem}

In the logistic regime, conditional independence makes the model well-specified with naive-Bayes log-odds. In the max-margin regime, a convex combination of training patterns cancels every covered weight. The point of \cref{thm:partial} is that one \emph{additive} rule governs the decision for every flip group in both regimes. Two consequences follow, and they reconcile the folklore that ``more coverage helps'' with the empirical Whac-a-Mole effect \citep{li2023whac}.

\begin{corollary}[Monotone worst case, non-monotone groups]
\label{cor:wam}
Under the conditions of \cref{thm:partial}:
\begin{enumerate}[label=(\alph*),leftmargin=*,itemsep=0pt,topsep=2pt]
\item $\min_F \Mmarg_K(F)=\sal_c-\sum_{j\in U}\sal_j$, attained at every $F\supseteq U$, is strictly increasing in $K$. Hence $\WGA$ is non-decreasing in coverage. The same holds if the core is intact only with probability $p_{\rm test}\in(0,1]$ at test time.
\item \emph{(Whac-a-Mole.)} Let $j\in U\setminus F$, so that $j$ is uncovered and aligned on $G_F$. Covering $j$ \emph{lowers} the margin of $G_F$ by $\sal_j$. It turns $G_F$ from correct to incorrect iff $0<\Mmarg_K(F)\le \sal_j$.
\end{enumerate}
\end{corollary}

The mechanism in (b) is \emph{loss of accidental support}: before it is covered, the aligned shortcut $j$ helps outvote the flipped shortcuts in $F$. Benchmarks that report single-shortcut groups, such as the background and co-occurring-object gaps of UrbanCars \citep{li2023whac}, can therefore show one gap widen when the \emph{other} shortcut is mitigated, exactly when $0<\Mmarg_K(F)\le\sal_j$. This happens even though, by (a), the worst case over all flip groups improves.

\begin{corollary}[Sufficient partial coverage]
\label{cor:suff}
All flip groups are classified correctly iff $\sal_c>\sum_{j\in U}\sal_j$. A minimum-cardinality sufficient cover consists of the shortcuts with the largest saliences, added in decreasing order until the residual salience falls below $\sal_c$.
\end{corollary}

Full coverage is therefore unnecessary. It suffices that the core outweighs the \emph{residual} shortcut salience. Each edit type has a cost and, as shown next, a risk.

\subsection{Validator error}

\begin{theorem}[False accepts]
\label{thm:validator}
Let a fraction $\delta$ of training inputs be false accepts, independent of $(Y,Z)$.
\begin{enumerate}[label=(\roman*),leftmargin=*,itemsep=0pt,topsep=2pt]
\item \emph{Logistic regime.} The minimizer is as in \cref{thm:partial}(i), except that $w^\star_c=\lambda(p_\delta)/a$ with $p_\delta=p(1-\delta)+(1-p)\delta$. All flip groups are correct iff $\delta<\delta^\star$, where
\begin{equation}
\delta^\star=\frac{p-\sigma(R_U)}{2p-1},\qquad R_U=\textstyle\sum_{j\in U}\lambda(\rho_j),
\label{eq:deltastar}
\end{equation}
and $\sigma$ is the logistic function. Here $\delta^\star>0$ iff coverage is sufficient (\cref{cor:suff}), and $\delta^\star$ increases as $U$ shrinks.
\item \emph{Max-margin regime.} Let $U\neq\emptyset$. Suppose the sample contains a false accept with $K$-pattern $z'$ and a genuine edit with pattern $-z'$. Then the max-margin direction is $w^\star\propto(0,\,0_K,\,(b_j)_{j\in U})$, a pure shortcut classifier (accuracy $0$ on $G_U$). If $U=\emptyset$, the sample is not linearly separable.
\end{enumerate}
\end{theorem}

Part (i) gives a closed-form error budget for the validator, which \emph{grows} with coverage. \begin{remark}[Class imbalance]
\label{rem:prior}
If $\P(Y=1)=\pi$ and the predictor has an intercept, the logistic-regime minimizer acquires the intercept $\lambda(\pi)$ and is otherwise unchanged. The minority class then loses margin $|\lambda(\pi)|$. All conditions above hold with $\sal_c$ replaced by $\sal_c-|\lambda(\pi)|$, and in particular
$\delta^\star_\pi=\big(p-\sigma(R_U+|\lambda(\pi)|)\big)/(2p-1)$.
Imbalance shrinks the validator's error budget. With full coverage and a near-deterministic core ($p\to1$), $\delta^\star_\pi\to\min(\pi,1-\pi)$.
\end{remark}

Part (ii) has a non-asymptotic core: on the core and uncovered coordinates, a false accept with label $y$ \emph{coincides} with the $G_U$ input of label $-y$. Any predictor that ignores covered coordinates and fits the false accept must misclassify that input. Interpolating edited data therefore fails the worst group as soon as the false accept is fitted. In our simulations this happens within ${\sim}10^3$ gradient steps, even though the direction converges only at rate $O(1/\log t)$ \citep{soudry2018implicit}. The result relies on $\rho_j=1$ for $j\in U$: one genuine training point with a misaligned uncovered shortcut restores $w^\star_c>0$. Regularization, early stopping, and auditing the validator are therefore advisable (\cref{sec:exp}).

\subsection{Discovery order and an audited certificate}

\begin{proposition}[Greedy discovery peels by salience]
\label{prop:greedy}
In either regime, under the current model $w^\star(K)$, randomizing shortcut $j\in U$ changes the logit of any point by $2\sal_j/\kappa$ in absolute value with probability $\tfrac12$, and leaves it unchanged otherwise. Covered-shortcut and null edits leave every logit unchanged. Here $\kappa=1$ in the logistic regime and $\kappa=\|(a,b_U)\|$ for the unit-norm max-margin predictor. Covering $j$ changes the remaining weights only by a common positive factor. Hence: (i) ranking edits by mean absolute logit change covers shortcuts in decreasing salience, so after $r$ greedy rounds $K=K_r$, the $r$ most salient shortcuts; (ii) the rule ``stop when no candidate changes the logit more than the null edits'' stops exactly when every shortcut with $\sal_j>0$ is covered.
\end{proposition}

\peel{} uses exactly rules (i) and (ii) (\cref{alg:peel}). Rule (ii) never stops while a used shortcut is uncovered, but it may cover more than \cref{cor:suff} requires. A cheaper rule stops when no edit changes a \emph{prediction}. In the logistic regime it stops at a minimum sufficient cover: with $m^\star=\min\{r:\sal_c>\sum_{j\notin K_r}\sal_j\}$, some uncovered edit changes a prediction with positive probability iff $r<m^\star$ (ties aside). In the max-margin regime, however, two shortcuts can jointly outvote the core while neither flip alone changes a prediction, so the flip rule can stop before the cover is sufficient (\cref{app:proofs}). \Cref{sec:e2} observes exactly this.

Finally, what can be \emph{certified} about a trained model, for any architecture? Let edit types $t=1,\dots,\tau$ act through kernels $\kappa_t(\cdot\mid x)$. Write $X'\sim\kappa_t(\cdot\mid X)$ for an edit of $X\sim P$, $A_t$ for the event that the validator accepts it, and $V_t\subseteq A_t$ for the event that it is accepted and label-preserving. Let
\begin{itemize}[leftmargin=*,itemsep=0pt,topsep=2pt]
\item $\delta_t=\P(A_t\setminus V_t\mid A_t)$, the false-accept rate of type $t$;
\item $\Gamma_t(f)=\P(f(X')\neq f(X)\mid V_t)$ and $\Gamma^{\rm acc}_t(f)=\P(f(X')\neq f(X)\mid A_t)$, the prediction-change rates on valid and on all accepted edits;
\item $Q_t$ be the law of $(X',Y)$ given $V_t$, and $\mathcal Q=\{\pi_0P+\sum_t\pi_tQ_t:\pi\in\Delta_\tau\}$.
\end{itemize}

\begin{assumption}[Content-independent validation]
\label{ass:accept}
For every $t$, the law of $(X,Y)$ given $V_t$ equals $P$.
\end{assumption}

\begin{theorem}[Audited certificate]
\label{thm:cert}
Let \cref{ass:accept} hold, and let $f$ and the edit family be independent of the following samples, each i.i.d.\ given them, with at most one edit per base input (samples may overlap):
\begin{itemize}[leftmargin=*,itemsep=0pt,topsep=2pt]
\item $n_{\rm v}$ labelled points from $P$;
\item $n_t$ accepted edits of each type, used to estimate $\widehat\Gamma^{\rm acc}_t$;
\item $m_t$ audited accepted edits of each type, labelled without error, giving a Clopper--Pearson upper bound $\bar\delta_t$ at level $\alpha/(3\tau)$.
\end{itemize}
Then with probability at least $1-\alpha$, simultaneously for all $Q\in\mathcal Q$,
\begin{equation}
\err_Q(f)\le \widehat{\err}_P(f)+\varepsilon_{n_{\rm v}}^{\alpha/3}+\max_{t}\frac{\widehat\Gamma^{\rm acc}_t+\varepsilon_{n_t}^{\alpha/3\tau}}{1-\bar\delta_t},
\label{eq:cert}
\end{equation}
where $\varepsilon_n^{\beta}=\sqrt{\log(1/\beta)/(2n)}$.
\end{theorem}

The certificate is deliberately modest. Its family $\mathcal Q$ consists of the distributions the \emph{accepted} valid edits generate. A validator that preferentially rejects valid edits that change $f$ makes $\mathcal Q$ under-represent the shift, and the family covers mixtures of single edits, not compositions. It bounds neither shortcuts outside the edit family nor reversed-shortcut groups. No observational procedure could do better: two causal models with the same $P(X,Y)$ can disagree on every flip group (\cref{app:impossible}; cf.\ \citealp{lin2022zin}). Its virtue is that every term is \emph{estimable}, including the validator's error, which is corrected by the factor $(1-\bar\delta_t)^{-1}$. Discovery chooses $f$ and the family, so the certificate needs data and an audit that discovery never used.

\section{\peel: Propose--Edit--Evaluate--Learn}
\label{sec:method}

The theory suggests a concrete procedure. It needs no group labels, it audits its validator, and it emits a certificate (\cref{alg:peel} in \cref{app:mainfigs}).
A pool of candidate edit types $\mathcal C$ is \emph{proposed} (e.g.\ background replacement, overlay removal, recoloring), applied, and filtered by the validator. Each candidate is \emph{evaluated} by the mean absolute change it causes in the current model's logit margin, against \emph{null} edits known to be irrelevant (one-sided Welch test, Holm). By \cref{prop:greedy}(i) this ranks uncovered shortcuts by salience. A small random sample of accepted edits of each surviving type is \emph{audited}, giving the Clopper--Pearson bound $\bar\delta_t$, and types with $\bar\delta_t>\delta_{\max}$ are discarded, which guards against \cref{thm:validator}(ii). Finally the model is \emph{retrained} on edited inputs with cross-entropy plus a Kullback--Leibler (KL) consistency term $\mu\,\mathrm{KL}(p_\theta(\cdot\mid x)\,\|\,p_\theta(\cdot\mid Tx))$, using weight decay and a fixed schedule rather than interpolation.
The loop stops when no candidate moves the logit significantly more than the null edits. By \cref{prop:greedy}(ii) this happens only once every shortcut the model uses is covered; a flip-rate rule could stop earlier. A cheaper variant, \peel-head, retrains only the last layer, like DFR \citep{kirichenko2023last} but without group labels.

\paragraph{Mistake-driven proposals (MD-\peel).} The candidate pool $\mathcal C$ need not be hand-specified. MD-\peel{} proposes candidates from the model's own failures.
\begin{itemize}[leftmargin=*,itemsep=0pt,topsep=2pt]
\item \emph{Contrast.} For each class it compares a \emph{trouble set} with the model's \emph{history}. The trouble set holds held-out examples that the current model misclassifies or predicts with low margin. Without labels, it holds low-confidence examples and examples where two empirical risk minimization (ERM) models of different seeds disagree. The history holds the model's highest-margin correct training examples.
\item \emph{Hypotheses.} A vision--language model (VLM) captions both groups, and concepts are scored with CLIP (contrastive language--image pre-training). Differences that are significant under a permutation test are merged by a large language model (LLM) into hypotheses about the \emph{context}; an LLM filter drops object properties.
\item \emph{Edits.} Each hypothesis becomes a REMOVE/ADD pair of edit instructions. An LLM maps each onto one of the benchmark's known intervention operators or onto none (an \emph{oracle library}), or a text-guided editor (CLIPSeg segmentation plus inpainting) realizes it, checked by a VLM validator.
\item \emph{Test.} As in \peel, hypotheses are tested by the change in the logit margin. The null must be a same-operation placebo (e.g.\ another background of the same type), because history and trouble sets have extreme margins that any large edit moves.
\end{itemize}

\section{Experiments}
\label{sec:exp}

We test the theory at three levels of realism: the stylized model itself with finite samples, gradient descent, and a noisy validator (E1); deep networks on a four-shortcut image benchmark (E2); a two-shortcut Waterbirds variant built from real photographs (E3); and five standard benchmarks, where MD-\peel{} proposes the candidate shortcuts from the model's own mistakes (E4). In E2 and E3 every intervention is \emph{exact} because images are composed from their factors. This lets us evaluate every flip group on the full test set and inject validator errors at controlled rates. No group labels are used for training or model selection unless marked~$\dagger$. By default every training input is edited ($q=1$), matching $\Ptil_{K,\delta}$. Details: \cref{app:experiments}.

\subsection{E1: the theory is quantitatively sharp}

We fit logistic regression on samples from $\Ptil_{K,\delta}$ ($k\in\{3,5\}$, random $\rho_j$, 10 seeds; figures in \cref{app:mainfigs}).
\emph{Decision rule.} At $n=10^5$ the rule of \cref{thm:partial} disagrees with the fitted model on $0.39\%$ of 7{,}680 flip-group evaluations, all near-ties. \emph{Whac-a-Mole.} Across 19{,}200 (cover, shortcut, group) triples, the sign of every accuracy change matches the theory in $99.4\%$, and $98.4\%$ of the predicted degradations occur. \emph{Validator tolerance.} The empirical false-accept threshold lies within $0.01$ of $\delta^\star$ in every row. \emph{Interpolation.} With noiseless data, one false accept among $1{,}000$ points drives worst-group accuracy to $0$ after $873$ gradient steps, the moment it is fitted. $\ell_2$ regularization restores a finite tolerance. \emph{Discovery and certificate.} Greedy discovery recovers the salience order in $94$--$98\%$ of instances, and the certificate was violated in $0$ of $200$ replications. \emph{Partial editing.} If only half of the training inputs are edited, covered shortcuts keep weight and only $35\%$ of predicted degradations occur.

\subsection{E2: deep networks, four shortcuts}
\label{sec:e2}

\paragraph{Setup.} The core task is CIFAR-10 cat vs.\ dog \citep{krizhevsky2009learning}, with $4{,}000$ training images per class. Four binary shortcuts are rendered exactly: a corner patch ($S_1$), a color tint ($S_2$), a border ($S_3$), and a low-amplitude grating ($S_4$). Each agrees with the label with probability $\rho_j=0.95$. We train a ResNet-18 \citep{he2016deep} from scratch for 40 epochs and select the last epoch. We use all 16 coverage sets (5 seeds) and all 16 flip groups. The core is weaker than every shortcut: ERM has $99.2$ IID (independent and identically distributed) test accuracy but $0.0$ worst-group accuracy, and the core alone supports ${\approx}73\%$.

\paragraph{Coverage.} Covered shortcuts have measured salience $0.10$ on average, against $5.27$ for uncovered ones (\cref{thm:partial}). \cref{cor:wam}(a) holds: worst-group accuracy rises from $0.0$ to $70.9\pm4.0$ with full coverage, and the worst group's margin rises along all 32 cover-one-more steps.

For \cref{cor:wam}(b), $72.5\%$ of the 512 (cover, shortcut, group) triples change significantly (paired $t$-test over seeds), and $97.6\%$ of these have the predicted sign. Where the newly covered shortcut is aligned and an uncovered flip remains, accuracy drops by $32.7$ points on average. $95\%$ of significant decreases have the covered shortcut aligned, which is exactly the loss-of-support mechanism. A per-example version of the rule, which subtracts each example's measured shortcut effect from the current margin, predicts the size of the change with $r=0.85$ (\cref{fig:e2}, left, \cref{app:mainfigs}).

\paragraph{One prediction fails in its intact-core form.} Groups whose flipped shortcuts are all covered should be unaffected, yet they lose $8.5$ points, mostly when the \emph{last} shortcut is covered. These are examples where the weak core errs and aligned shortcuts had rescued them: the noisy-core case and the in-distribution price of invariance (\cref{sec:setup}).

The additive structure itself is accurate: predicting each multi-flip group's accuracy from per-example single-flip effects gives $R^2=0.996$ over all coverage sets, seeds, and groups (\cref{fig:e2}, right).

\paragraph{Validator error and \peel.} Worst-group accuracy falls with $\delta$ in every setting (full coverage: $73.3\to54.7$ at $\delta=0.3$), but gradually, without the sharp threshold of the linear theory. Interpolation is more fragile only for the largest partial cover $\{1,3,4\}$ ($27.4\to15.5$ already at $\delta=0.02$, vs.\ $20.5\to17.1$ regularized). These are partial confirmations.

\peel{} finds all four shortcuts in every seed, in salience order (Spearman $0.96$; \cref{prop:greedy}), and the audit rejects core swap every time. \peel{} reaches $71.3\pm4.0$ worst-group accuracy, matching oracle coverage ($70.9$); without the audit it reaches $60.8$. Group-label baselines fail because the worst groups are nearly empty in training (JTT$^\dagger$ $0.2$, GroupDRO$^\dagger$ $34.4$, DFR$^\dagger$ $43.6$).

\peel-head on the frozen ERM backbone reaches only $19.0$: trained from scratch, the network never represents the core (a representation failure), unlike the pretrained backbone of E3. The certificate held in every case (mean bound $0.40$ vs.\ measured $0.27$).

\paragraph{Coverage does not transfer; redundant shortcuts hide from flip tests.} We add four more exact shortcuts, giving eight families, and cover only the first four (A--D) exactly. Accuracy on groups where only E--H are flipped then \emph{falls} by $42.4\pm1.6$ points: 11 of 15 such groups drop significantly and none rise (5 seeds). This is \cref{cor:wam}(b) at scale, and it rules out ``transfer'' to uncovered families. With eight redundant shortcuts, a single randomized shortcut rarely flips a prediction, because the others outvote it. A flip-rate test then finds nothing in 3 of 5 seeds, the early stopping noted after \cref{prop:greedy}, whereas \peel's margin test finds all eight families and reaches $71.2$ worst-group accuracy (oracle coverage: $68.6$).

\begin{table*}[!t]
\centering
\caption{Worst-group accuracy (\%; Spawrious: class$\times$location groups; average accuracy in \cref{tab:e4-spaw}), mean$\pm$standard deviation (std) over 3 seeds (Waterbirds-2S: 5). $\dagger$: group labels. n/a: no exact edit library; --: not run. Oracle library: hypotheses mapped onto the benchmark's known operators. Generated edits: three end-to-end runs (own edited set and classifier per discovery seed). O2O/M2M: one-to-one/many-to-many splits. CSI: certified shortcut immunization. ShortcutProbe \citep{zheng2025shortcutprobe} uses validation \emph{class} labels. $^\ast$CelebA: head-only retraining, 20k-image subset (no-edit control 43.5).}
\label{tab:main}
\small
\setlength{\tabcolsep}{4pt}
\begin{tabular}{lcccccc}
\toprule
Method & Waterbirds-2S & Waterbirds & UrbanCars & CelebA & Spawrious O2O & Spawrious M2M \\
\midrule
ERM & 66.5{\scriptsize$\pm$1.3} & 74.1{\scriptsize$\pm$1.4} & 20.0{\scriptsize$\pm$1.3} & 44.8{\scriptsize$\pm$3.0} & 55.1{\scriptsize$\pm$4.1} & 34.1{\scriptsize$\pm$2.3} \\
ShortcutProbe & -- & 85.6{\scriptsize$\pm$2.3} & 88.5{\scriptsize$\pm$0.4} & 82.4{\scriptsize$\pm$0.8} & 55.1{\scriptsize$\pm$4.0} & 32.7{\scriptsize$\pm$1.3} \\
CSI (learned adversary) & 67.7{\scriptsize$\pm$4.0} & 74.2{\scriptsize$\pm$1.2} & 24.0{\scriptsize$\pm$7.9} & 48.7{\scriptsize$\pm$1.8} & 55.1{\scriptsize$\pm$4.1} & 34.1{\scriptsize$\pm$2.3} \\
\midrule
\peel{} (hand list) & 98.2{\scriptsize$\pm$0.1} & 94.0{\scriptsize$\pm$0.4} & \textbf{94.1}{\scriptsize$\pm$0.4} & n/a & n/a & n/a \\
MD-\peel{} (oracle library) & 98.0{\scriptsize$\pm$0.1} & 93.5{\scriptsize$\pm$0.3} & \textbf{94.1}{\scriptsize$\pm$1.0} & n/a & n/a & n/a \\
MD-\peel{} (generated edits) & 78.9{\scriptsize$\pm$1.6} & 80.0{\scriptsize$\pm$0.5} & 72.0{\scriptsize$\pm$5.2} & 42.8{\scriptsize$\pm$5.7}$^\ast$ & 73.2{\scriptsize$\pm$3.5} & \textbf{46.5}{\scriptsize$\pm$1.6} \\
\midrule
GroupDRO$^\dagger$ & 85.6{\scriptsize$\pm$1.4} & 86.3{\scriptsize$\pm$1.4} & 80.5{\scriptsize$\pm$1.6} & 87.6{\scriptsize$\pm$3.8} & \textbf{80.1}{\scriptsize$\pm$3.6} & 40.2{\scriptsize$\pm$1.3} \\
JTT$^\dagger$ & 83.1{\scriptsize$\pm$3.8} & 80.6{\scriptsize$\pm$0.8} & 71.2{\scriptsize$\pm$4.6} & 49.8{\scriptsize$\pm$5.8} & 53.7{\scriptsize$\pm$4.5} & 26.6{\scriptsize$\pm$4.9} \\
DFR$^\dagger$ & 89.6{\scriptsize$\pm$0.6} & 92.2{\scriptsize$\pm$0.2} & 86.9{\scriptsize$\pm$1.0} & \textbf{89.9}{\scriptsize$\pm$0.1} & 71.9{\scriptsize$\pm$1.2} & 44.1{\scriptsize$\pm$1.9} \\
\bottomrule
\end{tabular}
\end{table*}

\subsection{E3: real photographs, two shortcuts}
\label{sec:e3}

\paragraph{Waterbirds-2S.} We composite CUB-200 birds \citep{wah2011caltech} with their segmentation masks onto Places365 land and water scenes \citep{zhou2018places}, as in Waterbirds \citep{sagawa2020distributionally}. The shortcuts are the background ($\rho=0.95$) and a semi-transparent ``\copyright{} PHOTOSTOCK'' watermark ($\rho=0.90$; cf.\ ImageNet-W, \citealp{li2023whac}). The edit library contains exact background swaps and watermark toggles; null edits (flip, color jitter, crop); and \emph{core swap}, which pastes a bird of the other class while keeping the label and so plays the false accept. We fine-tune an ImageNet ResNet-50 for 20 epochs, take the last epoch, and average over 5 seeds. ERM relies on both shortcuts: its saliences are $\hat\lambda_{\rm bg}=4.4$ and $\hat\lambda_{\rm wm}=2.6$, and its worst-group accuracy is $66.5$ against $98.9$ IID.

\begin{figure*}[t]
\centering
\includegraphics[width=0.85\textwidth]{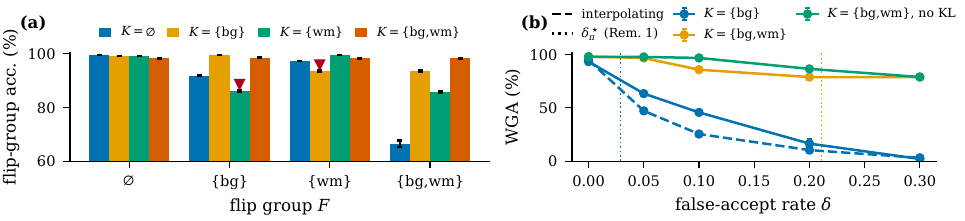}
\caption{E3 (Waterbirds-2S, ResNet-50, 5 seeds). (a) Flip-group accuracy per covered set. Covering one shortcut lowers accuracy on the group where the \emph{other} shortcut is flipped (triangles; \cref{cor:wam}(b)), while worst-group accuracy rises monotonically. (b) WGA vs.\ injected false-accept rate. Partial coverage collapses fastest; interpolation (dashed) is worse. Dropping the KL increases tolerance. Dotted lines mark $\delta^\star_\pi$ (\cref{rem:prior}).}
\label{fig:e3}
\end{figure*}

\paragraph{Whac-a-Mole on real images.} \Cref{fig:e3}(a) reproduces \cref{cor:wam} on photographs. Covering the background lowers the $\{$wm$\}$ group from $97.3$ to $93.5$ ($-3.8\pm0.3$). Covering the watermark lowers $\{$bg$\}$ from $91.9$ to $86.2$ ($-5.7\pm0.4$). Worst-group accuracy rises monotonically: $66.5\to93.4$/$85.8\to98.2$. Across all coverage transitions, 12 of 16 (transition, group) cells change by at least one point with a consistent sign, and \emph{all 12} have the sign predicted by \cref{thm:partial}. For a CLIP vision transformer (ViT-B/16), both fine-tuned and linearly probed, all mean changes have the predicted sign (\cref{app:experiments}).

\paragraph{Validator error.} \Cref{fig:e3}(b) injects false accepts. As \cref{thm:validator} predicts, partial coverage is fragile. With $K=\{$bg$\}$, $\delta=0.05$ drops worst-group accuracy from $93.4$ to $63.4$, and the fraction of core-swapped images classified by the bird falls from $90\%$ to $18\%$. Full coverage tolerates more. Interpolating training is worse under partial coverage ($47.2$ vs.\ $63.4$ at $\delta=0.05$) but not under full coverage, where the data are inseparable. Our pre-registered balanced-class tolerances ($\delta^\star=0.10$, $0.5$) were too optimistic. Accounting for the $21\%$ class prior (\cref{rem:prior}, derived after observing this failure) gives $\delta^\star_\pi=0.03$ and $0.21$, which matches where the curves break: at $\delta\ge0.2$ the fully covered model predicts the majority class. Finally, the consistency term amplifies false accepts. Without it, worst-group accuracy is $96.9$ at $\delta=0.1$, vs.\ $85.9$ with it.

\paragraph{\peel{} vs.\ baselines.} \peel{} flags background, watermark, and core swap in round~1. The audit rejects core swap ($50/50$ bad) and keeps the other two ($\bar\delta=0.058$), and round~2 finds nothing new. \peel{} reaches the oracle-coverage model ($98.2$; \cref{tab:main}). Applying random library edits under the same budget reaches $76.3$, and group-label methods reach at most $89.6$. Without the audit, core swaps enter the bank and worst-group accuracy falls to $85.6$. \peel-head, which retrains only the last layer, recovers most of the gain ($95.7$), so on a pretrained backbone the shortcut is largely a readout failure \citep{kirichenko2023last}. On held-out test birds, \cref{thm:cert} holds in all 20 (model, edit type) cases but is loose ($7.8$--$8.9\%$ vs.\ a measured $1.6$--$1.8\%$).

\subsection{E4: standard benchmarks, and finding shortcuts from mistakes}
\label{sec:e4}

\paragraph{Setup.} We use Waterbirds, UrbanCars \citep{li2023whac} (background and co-occurring object), CelebA (blond hair, spurious gender), and the one-to-one (O2O) and many-to-many (M2M) hard splits of Spawrious \citep{lynch2023spawrious}, with an ImageNet ResNet-50, 3 seeds, the last epoch, and one optimizer for all methods. Baselines: environment inference for invariant learning (EIIL; \citealp{creager2021environment}), SPARE \citep{yang2024identifying}, ShortcutProbe \citep{zheng2025shortcutprobe}, GroupDRO$^\dagger$, JTT$^\dagger$, DFR$^\dagger$ and invariant risk minimization (IRM$^\dagger$; \citealp{arjovsky2019invariant}). We also run a learned-adversary baseline, certified shortcut immunization (CSI): a U-Net adversary proposes edits, a CLIP zero-shot validator accepts them, and a novelty reward drives discovery. Hand-list \peel{} needs exact edits, which exist only for Waterbirds-2S, Waterbirds, and UrbanCars. MD-\peel{} starts from ERM with no candidate list and uses the oracle library (background swap, watermark toggle, co-object swap) or generated edits, validated by Qwen2.5-VL and audited by InternVL2.5 and CLIP (\cref{app:e4}).

\paragraph{MD-\peel{} names the shortcuts.} From the model's mistakes alone, MD-\peel{}'s confirmed hypotheses name the background and the watermark on Waterbirds-2S (on one seed the watermark appears only in round~2, after the background is covered, as \cref{prop:greedy} predicts), the background on Waterbirds, both shortcuts on UrbanCars, gender on CelebA, and the background on Spawrious O2O. With the oracle library the LLM only \emph{selects} among the benchmark's own operators: this tests semantic selection, not discovery of new interventions. On Spawrious M2M it confirms ``lush, tropical background'' and ``scattered background'', which our pre-registered keywords do not count as hits. Without any labels (low confidence plus seed disagreement) it confirms the same edits on all 9 exact-edit runs. Our pre-registered keyword score (seed 0) credits 4 of 5 datasets (lenient rule) or 2 of 5 (strict; mostly plurals). \emph{The test needs a placebo null.} With the specified pixel nulls it rejected no round-1 hypothesis, but it also confirmed $10$ of $30$ same-operation placebos and $9$ of $28$ re-tests of covered operations on fixed models. With a placebo null (post hoc) both fall to $0$, and all round-1 confirmations and final banks are unchanged (\cref{app:e4-md}).

\paragraph{\ldots and fixes them with the oracle library.} With the oracle library (\cref{tab:main}), MD-\peel{} matches hand-list \peel{} (98.0/93.5/94.1 vs.\ 98.2/94.0/94.1) and beats every group-label baseline, including DFR$^\dagger$ (89.6/92.2/86.9). On UrbanCars it closes all three Whac-a-Mole gaps; JTT$^\dagger$ closes the background gap but widens the co-object gap to $-16.4$. Generated edits are weaker evidence. We ran them end to end from three discovery seeds (round~2 not run). On Spawrious, where no library exists, they raise worst-group accuracy over ERM ($73.2$ vs.\ $55.1$ on O2O, $46.5$ vs.\ $34.1$ on M2M). That is the best M2M result, but GroupDRO$^\dagger$ is better on O2O ($80.1$). On average accuracy, the official metric, MD-\peel{} is best on M2M ($75.2$ vs.\ $72.2$) and ties GroupDRO$^\dagger$ on O2O ($89.1$ vs.\ $89.3$). It improves on ERM on Waterbirds-2S, Waterbirds and UrbanCars (78.9, 80.0, 72.0), but trails exact edits there, and discovery variance is large on UrbanCars ($64.8$--$76.8$). Its edits rarely damage the object (at most $0.3\%$ by ground-truth masks), but often fail to change the shortcut: only 14--43\% of accepted edits flip the true background. On CelebA it fails (42.8 vs.\ ERM 44.8). With seed 0 the audit removed the direct gender edit, leaving ``beard'' and an irrelevant hypothesis; seeds 1 and 2 kept nothing, so the loop returned ERM. The editor, not discovery, is the bottleneck.

\paragraph{The learned adversary fails, and coverage is what matters.} CSI never improves on ERM by more than seed noise: its adversary paints over the object, CLIP still accepts $79$--$92\%$ of those edits, and with a \emph{perfect} validator it finds almost nothing ($\le0.4$ bank members per run). Discovery is not always needed. In natural language inference (NLI) from Multi-Genre NLI (MNLI) to the HANS challenge set (Heuristic Analysis for NLI Systems; \citealp{mccoy2019right}), uniform training on all 12 rewrite families matches or beats any selection (HANS $64.7$ vs.\ $62.8$ MD-\peel, $62.4$ \peel, $62.3$ CSI; ERM $59.0$), and mistake-driven discovery confirms only style features. What matters: shortcut coverage and bad-edit filtering.

\section{Related Work}
\label{sec:related}

\paragraph{Group robustness without full annotations.} GroupDRO \citep{sagawa2020distributionally} needs training-group labels; JTT \citep{liu2021just}, DFR \citep{kirichenko2023last}, automatic feature reweighting (AFR; \citealp{qiu2023simple}) and SPARE \citep{yang2024identifying} still use group-labelled validation data; simple balancing is a strong baseline \citep{idrissi2022simple}.

\paragraph{Invariance, augmentation, and causality.} Complete intervention families yield counterfactual invariance \citep{veitch2021counterfactual,makar2022causally}; augmentations that vary all style factors isolate content \citep{vonkugelgen2021self} and can simulate interventions \citep{ilse2021selecting,gao2023out}; limits of IRM \citep{arjovsky2019invariant,kamath2021does,lin2022zin} motivate side information.

\paragraph{Multiple shortcuts and simplicity bias.} \citet{li2023whac} documented the Whac-a-Mole effect on UrbanCars and ImageNet variants and proposed a last-layer ensemble. \citet{khani2021removing} showed in regression that removing a spurious feature can hurt accuracy and affect groups disproportionately. \citet{zheng2022causally} showed that removing a sufficient set of shortcuts suffices. Our \cref{cor:wam,cor:suff} give the exact per-group condition for classification. Simplicity bias \citep{shah2020pitfalls} and the max-margin bias of gradient descent \citep{soudry2018implicit,puli2023dont} explain why shortcuts are learned.

\paragraph{Generative discovery and augmentation.} Diffusion-based augmentation \citep{dunlap2023diversify}, concept-based and error-slice discovery \citep{wu2023discover,eyuboglu2022domino,kim2024discovering}, and edit-based stress tests \citep{prabhu2023lance,wiles2022discovering} propose edits automatically; MD-\peel{} instead contrasts failures with the model's own confident history and \emph{tests} each hypothesis by intervention first. Counterfactually augmented data can be ineffective or harmful \citep{kaushik2020learning,joshi2022investigation}. \citet{bai2025invariant} bound out-of-distribution (OOD) error when noise perturbs counterfactual features; we analyze label-changing false accepts. Our audit correction resembles prediction-powered inference \citep{angelopoulos2023prediction}.

\section{Discussion}
\label{sec:conclusion}

Intervention-based mitigation trades group labels for coverage and validation; our analysis quantifies this trade: partial coverage helps the worst case but redistributes errors, and the validator's error budget grows with coverage and shrinks with class imbalance. In practice: edit every input, audit the validator, regularize, not interpolate, and report every flip group.


\subsection*{AI use statement}

In this work, we used generative artificial intelligence (AI) tools (LLM assistants, including agentic coding assistants) for the following tasks with required disclosure: refining the research question and hypotheses; developing the theoretical model, formulating the mathematical claims and drafting their proofs; designing the experiments; implementing the methods and baselines, including the code that generates the synthetic and composited datasets (E1--E3) and preprocesses the public datasets; and interpreting results. We have not used generative AI tools for translation, and qualitative or thematic data analysis is not applicable to this work. Additionally, we used them for tasks with recommended disclosure: surveying the literature, brainstorming, running experiments on our cluster, producing figures, verifying references, and drafting and editing the text. The idea of proposing candidate edits from the model's own mistakes (MD-\peel) came from an author. Generative models (Qwen2.5-VL, Qwen2.5, InternVL2.5, Stable Diffusion inpainting) are also \emph{components} of MD-\peel{} and of our NLP experiments (\cref{sec:e4}).

We have reviewed all AI-assisted work. An independent model-based review checked the proofs adversarially, and the issues it found were corrected; the theorems were also validated numerically (E1). Every reported number comes from logged runs included in the supplementary material; we did not use generative AI to fabricate or alter experimental data. Bibliographic entries were verified against publisher or arXiv records. We take responsibility for the final content of this work, including text, claims, and artifacts produced with the aid of generative AI.

\bibliography{references}

@article{geirhos2020shortcut,
  author    = {Robert Geirhos and J{\"{o}}rn{-}Henrik Jacobsen and Claudio Michaelis and Richard Zemel and Wieland Brendel and Matthias Bethge and Felix A. Wichmann},
  title     = {Shortcut Learning in Deep Neural Networks},
  journal   = {Nature Machine Intelligence},
  volume    = {2},
  number    = {11},
  pages     = {665--673},
  year      = {2020},
  doi       = {10.1038/s42256-020-00257-z}
}

@inproceedings{sagawa2020distributionally,
  author    = {Shiori Sagawa and Pang Wei Koh and Tatsunori B. Hashimoto and Percy Liang},
  title     = {Distributionally Robust Neural Networks for Group Shifts: On the Importance of Regularization for Worst-Case Generalization},
  booktitle = {International Conference on Learning Representations ({ICLR})},
  year      = {2020},
  url       = {https://openreview.net/forum?id=ryxGuJrFvS}
}

@inproceedings{liu2021just,
  author    = {Evan Zheran Liu and Behzad Haghgoo and Annie S. Chen and Aditi Raghunathan and Pang Wei Koh and Shiori Sagawa and Percy Liang and Chelsea Finn},
  title     = {Just Train Twice: Improving Group Robustness without Training Group Information},
  booktitle = {Proceedings of the 38th International Conference on Machine Learning ({ICML})},
  series    = {Proceedings of Machine Learning Research},
  volume    = {139},
  pages     = {6781--6792},
  publisher = {PMLR},
  year      = {2021}
}

@inproceedings{kirichenko2023last,
  author    = {Polina Kirichenko and Pavel Izmailov and Andrew Gordon Wilson},
  title     = {Last Layer Re-Training is Sufficient for Robustness to Spurious Correlations},
  booktitle = {International Conference on Learning Representations ({ICLR})},
  year      = {2023},
  url       = {https://openreview.net/forum?id=Zb6c8A-Fghk}
}

@inproceedings{puli2023dont,
  author    = {Aahlad Puli and Lily Zhang and Yoav Wald and Rajesh Ranganath},
  title     = {Don't Blame Dataset Shift! {S}hortcut Learning due to Gradients and Cross Entropy},
  booktitle = {Advances in Neural Information Processing Systems ({NeurIPS})},
  volume    = {36},
  pages     = {71874--71910},
  year      = {2023}
}

@inproceedings{shah2020pitfalls,
  author    = {Harshay Shah and Kaustav Tamuly and Aditi Raghunathan and Prateek Jain and Praneeth Netrapalli},
  title     = {The Pitfalls of Simplicity Bias in Neural Networks},
  booktitle = {Advances in Neural Information Processing Systems ({NeurIPS})},
  volume    = {33},
  pages     = {9573--9585},
  year      = {2020}
}

@article{soudry2018implicit,
  author    = {Daniel Soudry and Elad Hoffer and Mor Shpigel Nacson and Suriya Gunasekar and Nathan Srebro},
  title     = {The Implicit Bias of Gradient Descent on Separable Data},
  journal   = {Journal of Machine Learning Research},
  volume    = {19},
  number    = {70},
  pages     = {1--57},
  year      = {2018}
}

@inproceedings{li2023whac,
  author    = {Zhiheng Li and Ivan Evtimov and Albert Gordo and Caner Hazirbas and Tal Hassner and Cristian Canton Ferrer and Chenliang Xu and Mark Ibrahim},
  title     = {A Whac-a-Mole Dilemma: Shortcuts Come in Multiples Where Mitigating One Amplifies Others},
  booktitle = {Proceedings of the {IEEE/CVF} Conference on Computer Vision and Pattern Recognition ({CVPR})},
  pages     = {20071--20082},
  year      = {2023}
}

@inproceedings{veitch2021counterfactual,
  author    = {Victor Veitch and Alexander D'Amour and Steve Yadlowsky and Jacob Eisenstein},
  title     = {Counterfactual Invariance to Spurious Correlations in Text Classification},
  booktitle = {Advances in Neural Information Processing Systems ({NeurIPS})},
  volume    = {34},
  pages     = {16196--16208},
  year      = {2021}
}

@inproceedings{vonkugelgen2021self,
  author    = {Julius von K{\"{u}}gelgen and Yash Sharma and Luigi Gresele and Wieland Brendel and Bernhard Sch{\"{o}}lkopf and Michel Besserve and Francesco Locatello},
  title     = {Self-Supervised Learning with Data Augmentations Provably Isolates Content from Style},
  booktitle = {Advances in Neural Information Processing Systems ({NeurIPS})},
  volume    = {34},
  pages     = {16451--16467},
  year      = {2021}
}

@inproceedings{lin2022zin,
  author    = {Yong Lin and Shengyu Zhu and Lu Tan and Peng Cui},
  title     = {{ZIN}: When and How to Learn Invariance Without Environment Partition?},
  booktitle = {Advances in Neural Information Processing Systems ({NeurIPS})},
  volume    = {35},
  pages     = {24529--24542},
  year      = {2022}
}

@article{arjovsky2019invariant,
  author    = {Martin Arjovsky and L{\'{e}}on Bottou and Ishaan Gulrajani and David Lopez{-}Paz},
  title     = {Invariant Risk Minimization},
  journal   = {arXiv preprint arXiv:1907.02893},
  year      = {2019}
}

@inproceedings{kamath2021does,
  author    = {Pritish Kamath and Akilesh Tangella and Danica J. Sutherland and Nathan Srebro},
  title     = {Does Invariant Risk Minimization Capture Invariance?},
  booktitle = {Proceedings of the 24th International Conference on Artificial Intelligence and Statistics ({AISTATS})},
  series    = {Proceedings of Machine Learning Research},
  volume    = {130},
  pages     = {4069--4077},
  publisher = {PMLR},
  year      = {2021}
}

@inproceedings{creager2021environment,
  author    = {Elliot Creager and J{\"{o}}rn{-}Henrik Jacobsen and Richard Zemel},
  title     = {Environment Inference for Invariant Learning},
  booktitle = {Proceedings of the 38th International Conference on Machine Learning ({ICML})},
  series    = {Proceedings of Machine Learning Research},
  volume    = {139},
  pages     = {2189--2200},
  publisher = {PMLR},
  year      = {2021}
}

@inproceedings{ilse2021selecting,
  author    = {Maximilian Ilse and Jakub M. Tomczak and Patrick Forr{\'{e}}},
  title     = {Selecting Data Augmentation for Simulating Interventions},
  booktitle = {Proceedings of the 38th International Conference on Machine Learning ({ICML})},
  series    = {Proceedings of Machine Learning Research},
  volume    = {139},
  pages     = {4555--4562},
  publisher = {PMLR},
  year      = {2021}
}

@inproceedings{makar2022causally,
  author    = {Maggie Makar and Ben Packer and Dan Moldovan and Davis Blalock and Yoni Halpern and Alexander D'Amour},
  title     = {Causally Motivated Shortcut Removal Using Auxiliary Labels},
  booktitle = {Proceedings of the 25th International Conference on Artificial Intelligence and Statistics ({AISTATS})},
  series    = {Proceedings of Machine Learning Research},
  volume    = {151},
  pages     = {739--766},
  publisher = {PMLR},
  year      = {2022}
}

@inproceedings{qiu2023simple,
  author    = {Shikai Qiu and Andres Potapczynski and Pavel Izmailov and Andrew Gordon Wilson},
  title     = {Simple and Fast Group Robustness by Automatic Feature Reweighting},
  booktitle = {Proceedings of the 40th International Conference on Machine Learning ({ICML})},
  series    = {Proceedings of Machine Learning Research},
  volume    = {202},
  pages     = {28448--28467},
  publisher = {PMLR},
  year      = {2023}
}

@inproceedings{yang2024identifying,
  author    = {Yu Yang and Eric Gan and Gintare Karolina Dziugaite and Baharan Mirzasoleiman},
  title     = {Identifying Spurious Biases Early in Training through the Lens of Simplicity Bias},
  booktitle = {Proceedings of the 27th International Conference on Artificial Intelligence and Statistics ({AISTATS})},
  series    = {Proceedings of Machine Learning Research},
  volume    = {238},
  pages     = {2953--2961},
  publisher = {PMLR},
  year      = {2024}
}

@inproceedings{idrissi2022simple,
  author    = {Badr Youbi Idrissi and Mart{\'{\i}}n Arjovsky and Mohammad Pezeshki and David Lopez{-}Paz},
  title     = {Simple Data Balancing Achieves Competitive Worst-Group-Accuracy},
  booktitle = {Proceedings of the First Conference on Causal Learning and Reasoning ({CLeaR})},
  series    = {Proceedings of Machine Learning Research},
  volume    = {177},
  pages     = {336--351},
  publisher = {PMLR},
  year      = {2022}
}

@inproceedings{zheng2025shortcutprobe,
  author    = {Guangtao Zheng and Wenqian Ye and Aidong Zhang},
  title     = {{ShortcutProbe}: Probing Prediction Shortcuts for Learning Robust Models},
  booktitle = {Proceedings of the Thirty-Fourth International Joint Conference on Artificial Intelligence ({IJCAI})},
  pages     = {7146--7154},
  year      = {2025},
  doi       = {10.24963/ijcai.2025/795}
}

@inproceedings{dunlap2023diversify,
  author    = {Lisa Dunlap and Alyssa Umino and Han Zhang and Jiezhi Yang and Joseph E. Gonzalez and Trevor Darrell},
  title     = {Diversify Your Vision Datasets with Automatic Diffusion-Based Augmentation},
  booktitle = {Advances in Neural Information Processing Systems ({NeurIPS})},
  volume    = {36},
  pages     = {79024--79034},
  year      = {2023}
}

@inproceedings{wu2023discover,
  author    = {Shirley Wu and Mert Yuksekgonul and Linjun Zhang and James Zou},
  title     = {Discover and Cure: Concept-aware Mitigation of Spurious Correlation},
  booktitle = {Proceedings of the 40th International Conference on Machine Learning ({ICML})},
  series    = {Proceedings of Machine Learning Research},
  volume    = {202},
  pages     = {37765--37786},
  publisher = {PMLR},
  year      = {2023}
}

@inproceedings{gao2023out,
  author    = {Irena Gao and Shiori Sagawa and Pang Wei Koh and Tatsunori Hashimoto and Percy Liang},
  title     = {Out-of-Domain Robustness via Targeted Augmentations},
  booktitle = {Proceedings of the 40th International Conference on Machine Learning ({ICML})},
  series    = {Proceedings of Machine Learning Research},
  volume    = {202},
  pages     = {10800--10834},
  publisher = {PMLR},
  year      = {2023}
}

@inproceedings{eyuboglu2022domino,
  author    = {Sabri Eyuboglu and Maya Varma and Khaled Kamal Saab and Jean{-}Benoit Delbrouck and Christopher Lee{-}Messer and Jared Dunnmon and James Zou and Christopher R{\'{e}}},
  title     = {Domino: Discovering Systematic Errors with Cross-Modal Embeddings},
  booktitle = {International Conference on Learning Representations ({ICLR})},
  year      = {2022},
  url       = {https://openreview.net/forum?id=FPCMqjI0jXN}
}

@inproceedings{prabhu2023lance,
  author    = {Viraj Prabhu and Sriram Yenamandra and Prithvijit Chattopadhyay and Judy Hoffman},
  title     = {{LANCE}: Stress-testing Visual Models by Generating Language-guided Counterfactual Images},
  booktitle = {Advances in Neural Information Processing Systems ({NeurIPS})},
  volume    = {36},
  pages     = {25165--25184},
  year      = {2023}
}

@inproceedings{kim2024discovering,
  author    = {Younghyun Kim and Sangwoo Mo and Minkyu Kim and Kyungmin Lee and Jaeho Lee and Jinwoo Shin},
  title     = {Discovering and Mitigating Visual Biases through Keyword Explanation},
  booktitle = {Proceedings of the {IEEE/CVF} Conference on Computer Vision and Pattern Recognition ({CVPR})},
  pages     = {11082--11092},
  year      = {2024}
}

@article{wiles2022discovering,
  author    = {Olivia Wiles and Isabela Albuquerque and Sven Gowal},
  title     = {Discovering Bugs in Vision Models using Off-the-shelf Image Generation and Captioning},
  journal   = {arXiv preprint arXiv:2208.08831},
  year      = {2022},
  note      = {NeurIPS 2022 ML Safety Workshop}
}

@inproceedings{mccoy2019right,
  author    = {R. Thomas McCoy and Ellie Pavlick and Tal Linzen},
  title     = {Right for the Wrong Reasons: Diagnosing Syntactic Heuristics in Natural Language Inference},
  booktitle = {Proceedings of the 57th Annual Meeting of the Association for Computational Linguistics},
  pages     = {3428--3448},
  publisher = {Association for Computational Linguistics},
  year      = {2019},
  doi       = {10.18653/v1/P19-1334}
}

@inproceedings{kaushik2020learning,
  author    = {Divyansh Kaushik and Eduard Hovy and Zachary C. Lipton},
  title     = {Learning the Difference that Makes a Difference with Counterfactually-Augmented Data},
  booktitle = {International Conference on Learning Representations ({ICLR})},
  year      = {2020},
  url       = {https://openreview.net/forum?id=Sklgs0NFvr}
}

@inproceedings{joshi2022investigation,
  author    = {Nitish Joshi and He He},
  title     = {An Investigation of the (In)effectiveness of Counterfactually Augmented Data},
  booktitle = {Proceedings of the 60th Annual Meeting of the Association for Computational Linguistics (Volume 1: Long Papers)},
  pages     = {3668--3681},
  publisher = {Association for Computational Linguistics},
  year      = {2022},
  doi       = {10.18653/v1/2022.acl-long.256}
}

@article{lynch2023spawrious,
  author    = {Aengus Lynch and Gb{\`{e}}tondji J.{-}S. Dovonon and Jean Kaddour and Ricardo Silva},
  title     = {Spawrious: A Benchmark for Fine Control of Spurious Correlation Biases},
  journal   = {arXiv preprint arXiv:2303.05470},
  year      = {2023}
}

@techreport{wah2011caltech,
  author      = {Catherine Wah and Steve Branson and Peter Welinder and Pietro Perona and Serge Belongie},
  title       = {The {Caltech-UCSD} Birds-200-2011 Dataset},
  institution = {California Institute of Technology},
  number      = {CNS-TR-2011-001},
  year        = {2011}
}

@article{zhou2018places,
  author    = {Bolei Zhou and Agata Lapedriza and Aditya Khosla and Aude Oliva and Antonio Torralba},
  title     = {Places: A 10 Million Image Database for Scene Recognition},
  journal   = {{IEEE} Transactions on Pattern Analysis and Machine Intelligence},
  volume    = {40},
  number    = {6},
  pages     = {1452--1464},
  year      = {2018},
  doi       = {10.1109/TPAMI.2017.2723009}
}

@inproceedings{he2016deep,
  author    = {Kaiming He and Xiangyu Zhang and Shaoqing Ren and Jian Sun},
  title     = {Deep Residual Learning for Image Recognition},
  booktitle = {Proceedings of the {IEEE} Conference on Computer Vision and Pattern Recognition ({CVPR})},
  pages     = {770--778},
  year      = {2016},
  doi       = {10.1109/CVPR.2016.90}
}

@techreport{krizhevsky2009learning,
  author      = {Alex Krizhevsky},
  title       = {Learning Multiple Layers of Features from Tiny Images},
  institution = {University of Toronto},
  year        = {2009}
}

@inproceedings{khani2021removing,
  author    = {Fereshte Khani and Percy Liang},
  title     = {Removing Spurious Features can Hurt Accuracy and Affect Groups Disproportionately},
  booktitle = {Proceedings of the 2021 {ACM} Conference on Fairness, Accountability, and Transparency (FAccT)},
  pages     = {196--205},
  publisher = {{ACM}},
  year      = {2021},
  doi       = {10.1145/3442188.3445883}
}

@inproceedings{zheng2022causally,
  author    = {Jiayun Zheng and Maggie Makar},
  title     = {Causally motivated multi-shortcut identification and removal},
  booktitle = {Advances in Neural Information Processing Systems 35 (NeurIPS)},
  year      = {2022}
}

@article{bai2025invariant,
  author    = {Ruqi Bai and Yao Ji and Zeyu Zhou and David I. Inouye},
  title     = {From Invariant Representations to Invariant Data: Provable Robustness to Spurious Correlations via Noisy Counterfactual Matching},
  journal   = {CoRR},
  volume    = {abs/2505.24843},
  year      = {2025},
  eprint    = {2505.24843},
  archivePrefix = {arXiv}
}

@article{angelopoulos2023prediction,
  author    = {Anastasios N. Angelopoulos and Stephen Bates and Clara Fannjiang and Michael I. Jordan and Tijana Zrnic},
  title     = {Prediction-powered inference},
  journal   = {Science},
  volume    = {382},
  number    = {6671},
  pages     = {669--674},
  year      = {2023},
  doi       = {10.1126/science.adi6000}
}

@inproceedings{nagarajan2021understanding,
  title     = {Understanding the Failure Modes of Out-of-Distribution Generalization},
  author    = {Nagarajan, Vaishnavh and Andreassen, Anders and Neyshabur, Behnam},
  booktitle = {International Conference on Learning Representations (ICLR)},
  year      = {2021}
}


\clearpage
\appendix
\thispagestyle{empty}
\onecolumn
\aistatstitle{Peel, Don't Whack:\\
Supplementary Materials}
\section{Proofs}
\label{app:proofs}

Throughout, $x=(x_c,x_1,\dots,x_k)$ follows \eqref{eq:model}. For a feature $f\in\{c,1,\dots,k\}$ we write $m_f$ for its magnitude ($m_c=a$, $m_j=b_j$) and $p_f$ for its agreement probability under the training distribution. Under $\Ptil_{K,\delta}$ these are $p_c=p_\delta$, $p_j=\tfrac12$ for $j\in K$, and $p_j=\rho_j$ for $j\in U$.

\subsection{Proof of \cref{thm:partial} and \cref{thm:validator}(i) (logistic regime)}

\begin{lemma}
\label{lem:nb}
Under $\Ptil_{K,\delta}$ the coordinates of $X$ are conditionally independent given $Y$, and
$\log\frac{\P(Y=1\mid X=x)}{\P(Y=-1\mid X=x)}=\sum_{f}\frac{x_f}{m_f}\lambda(p_f).$
\end{lemma}
\begin{proof}
Randomizing $Z_j$ replaces it by a fresh independent sign. Flipping $Z_c$ with probability $\delta$, independently, yields a sign with $\P(Z_c=1)=p(1-\delta)+(1-p)\delta=p_\delta$. Hence the $Z$'s remain independent given $Y$. For each feature, $x_f\in\{\pm m_f\}$ and $\P(x_f=m_f\mid Y=1)=p_f=\P(x_f=-m_f\mid Y=-1)$. So the log-likelihood ratio of $x_f$ is $\frac{x_f}{m_f}\lambda(p_f)$. Sum over features and use the balanced prior.
\end{proof}

\begin{proof}[Proof of \cref{thm:partial}(i) and \cref{thm:validator}(i)]
The logistic loss is a strictly proper scoring rule. Its population risk $R(w)=\E\log(1+e^{-Yw^\top X})$ is therefore minimized by any $w$ with $\sigma(w^\top x)=\P(Y=1\mid X=x)$ almost surely. By \cref{lem:nb}, $w^\star_f=\lambda(p_f)/m_f$ satisfies this. Uniqueness: $R$ is strictly convex in every direction $v$ with $\E(v^\top X)^2>0$. Moreover
$\E XX^\top=\mathrm{diag}\big(m_f^2(1-(2p_f-1)^2)\big)+uu^\top$ with $u_f=m_f(2p_f-1)$,
which is positive definite because $p_f<1$ for every $f$. Plugging in $p_c=p_\delta$ (with $\delta=0$ for \cref{thm:partial}), $p_j=\tfrac12\Rightarrow\lambda=0$ for $j\in K$, and $p_j=\rho_j$ for $j\in U$ gives the weights.

For a point of $G_F$ with label $y$, we have $x_c=ay$ and $x_j=-b_jy$ for $j\in F$, $x_j=b_jy$ otherwise. Thus
$y\,w^{\star\top}x=\lambda(p_\delta)+\sum_{j\in U\setminus F}\lambda(\rho_j)-\sum_{j\in U\cap F}\lambda(\rho_j)$,
and covered coordinates contribute $0$. With $\delta=0$ this is $\Mmarg_K(F)$, proving the decision rule.

For \cref{thm:validator}(i), the worst group is any $F\supseteq U$, with margin $\lambda(p_\delta)-R_U$. This is positive iff $p_\delta>\sigma(R_U)$. Since $p_\delta=p-\delta(2p-1)$ is strictly decreasing in $\delta$ for $p>\tfrac12$, the condition is equivalent to $\delta<(p-\sigma(R_U))/(2p-1)=\delta^\star$. Finally, $\delta^\star>0$ iff $p>\sigma(R_U)$ iff $\lambda(p)>R_U$, which is the condition of \cref{cor:suff}.
\end{proof}

\begin{proof}[Proof of \cref{rem:prior}]
With prior $\pi$, \cref{lem:nb} gives log-odds $\lambda(\pi)+\sum_f\frac{x_f}{m_f}\lambda(p_f)$, which the logistic model with intercept represents exactly. A point of $G_F$ with label $y$ has margin $y\lambda(\pi)+(\text{margin of \cref{thm:partial}})$. The minority label therefore has margin $\Mmarg_K(F)-|\lambda(\pi)|$. The worst group is again $F\supseteq U$, and the condition $\lambda(p_\delta)-R_U-|\lambda(\pi)|>0$ is equivalent to $\delta<\delta^\star_\pi$, as in the balanced case. As $p\to1$, $\delta^\star_\pi\to1-\sigma(R_U+|\lambda(\pi)|)$, which equals $\min(\pi,1-\pi)$ when $U=\emptyset$.
\end{proof}

\subsection{Proof of \cref{thm:partial}(ii) and \cref{thm:validator}(ii) (max-margin regime)}

Recall that for linearly separable data, gradient descent on the (unregularized) empirical logistic loss with a sufficiently small step size satisfies $w(t)/\|w(t)\|\to\hat w$, where $\hat w$ is the unique solution of $\max_{\|w\|=1}\min_i y_iw^\top x_i$ \citep{soudry2018implicit}.

\begin{proof}[Proof of \cref{thm:partial}(ii)]
With $p=1$ and $\rho_j=1$ on $U$, every training point satisfies
$y_iw^\top x_i=v^\top\beta+w_K^\top(b_K\circ z_i)$, where $v=(w_c,w_U)$, $\beta=(a,b_U)$, $z_i$ is the $K$-pattern of the point, and $\circ$ is the entrywise product.
Let $\sum_i\alpha_i(b_K\circ z_i)=0$ with $\alpha$ in the simplex, as assumed. Then the minimum margin satisfies
$\phi(w)=\min_i y_iw^\top x_i\le\sum_i\alpha_i y_iw^\top x_i=v^\top\beta\le\|v\|\|\beta\|\le\|\beta\|$ for $\|w\|=1$.
Equality requires $w_K=0$, $\|v\|=1$, and $v=\beta/\|\beta\|$. That vector attains $\phi=\|\beta\|>0$, so it is the unique maximizer. The decision rule follows by evaluating $y\,\hat w^\top x\propto a^2+\sum_{U\setminus F}b_j^2-\sum_{U\cap F}b_j^2$ on $G_F$.

On sample size: for $n$ i.i.d.\ uniform sign patterns, $0\notin\mathrm{conv}$ iff all patterns lie in an open half-space through the origin. By Wendel's theorem, for points in general position this has probability $2^{-n+1}\sum_{i<|K|}\binom{n-1}{i}$, which is negligible once $n\gtrsim4|K|$. Ties among sign patterns only make this event less likely.
\end{proof}

\begin{proof}[Proof of \cref{thm:validator}(ii)]
Let $x'$ be the false-accepted point. It has $z_c=-1$, $K$-pattern $z'$, and aligned uncovered shortcuts. Let $x$ be a genuine point with $K$-pattern $-z'$. The minimum margin is at most the average of their margins. The $K$-terms and the core terms cancel:
\[
\phi(w)\le\tfrac12\Big[\big(w_ca+B_U-w_K^\top(b_K\circ z')\big)+\big(-w_ca+B_U+w_K^\top(b_K\circ z')\big)\Big]=B_U(w)\le\|w_U\|\,\|b_U\|\le\|b_U\|,
\]
where $B_U(w)=\sum_{j\in U}w_jb_j$. Equality forces $\|w_U\|=1$, hence $w_c=0$, $w_K=0$, and $w_U=b_U/\|b_U\|$. This vector gives every training point, genuine or false-accepted, margin $\|b_U\|>0$, so it is the unique maximizer. On $G_U$ every uncovered shortcut is flipped, so the margin is $-\|b_U\|<0$ and accuracy is $0$.

If $U=\emptyset$, any $w$ correct on $x$ has $w_ca>w_K^\top(b_K\circ z')$. Then $x'$ has margin $-w_ca+w_K^\top(b_K\circ z')<0$, so no homogeneous separator exists.

\emph{Remarks.}
\begin{itemize}[leftmargin=*,itemsep=0pt,topsep=2pt]
\item The condition on $-z'$ cannot be dropped. With genuine patterns $\{(+,+),(-,-)\}$ and a false accept with pattern $(+,-)$, the max-margin solution keeps $w_c>0$.
\item The result is specific to $\rho_j=1$ on $U$. A genuine point with a flipped uncovered shortcut restores $w_c>0$.
\item Gradient descent approaches the max-margin direction only at rate $O(1/\log t)$ \citep{soudry2018implicit}.
\end{itemize}
\end{proof}

\subsection{Proof of \cref{cor:wam}, \cref{cor:suff}, and \cref{prop:greedy}}

\begin{proof}[Proof of \cref{cor:wam}]
(a) $\Mmarg_K(F)$ is minimized by putting every uncovered index into $F$, giving $\sal_c-\sum_U\sal_j$. Moving $j$ from $U$ to $K$ increases this by $\sal_j>0$. With a noisy core at test time, the accuracy of $G_F$ is $p\cdot\1[\Mmarg_K(F)>0]+(1-p)\cdot\1[\Mmarg_K(F)-2\sal_c>0]$. This is the same non-decreasing function of $\Mmarg_K(F)$ for every $F$, so the minimum accuracy is attained at the minimum margin and is non-decreasing in $K$.
(b) For $j\in U\setminus F$, $\Mmarg_{K\cup\{j\}}(F)=\Mmarg_K(F)-\sal_j$, since $j$ leaves the positive sum. The sign changes from positive to non-positive iff $0<\Mmarg_K(F)\le\sal_j$.
\end{proof}

\begin{proof}[Proof of \cref{cor:suff}]
The first claim is (a) with the minimum margin required to be positive. For the second, the condition $\sum_U\sal_j<\sal_c$ concerns only the residual sum. Among covers of a given size, the residual is minimized by covering the largest saliences (exchange argument). So the smallest feasible size is attained by the greedy prefix in decreasing salience.
\end{proof}

\begin{proof}[Proof of \cref{prop:greedy}]
\emph{Ranking.} Randomizing $Z_j$ leaves it unchanged with probability $\tfrac12$ and flips it otherwise. A flip changes $w^\top x$ by $\pm2w_jb_j$. By \cref{thm:partial}, $w_jb_j=\lambda(\rho_j)$ in the logistic regime and $b_j^2/\|(a,b_U)\|$ for the unit-norm max-margin predictor. The mean absolute change is therefore $\sal_j/\kappa$, maximized by the most salient uncovered shortcut. Covering it leaves the other weights unchanged (logistic) or rescales them by a common factor (max-margin), so the ranking of the rest is preserved. By induction, $K=K_r$ after $r$ rounds.

\emph{Margin stopping, rule (ii).} By \cref{thm:partial} the covered weights are $0$, so covered and null edits leave every logit unchanged, while each uncovered $j$ has mean absolute logit change $\sal_j/\kappa$, which is positive iff $\sal_j>0$. The rule therefore continues while some shortcut with $\sal_j>0$ is uncovered and stops once all are covered.

\emph{Flip-based stopping (logistic regime; the alternative discussed after the proposition).} The logit of an input is $\sum_f z_f\sal_f$, with only $c$ and $U$ contributing.
\begin{itemize}[leftmargin=*,itemsep=0pt,topsep=2pt]
\item If $\sal_c>\sum_U\sal_j$, then $|\sum_U\pm\sal_j|<\sal_c$, so the sign of the logit equals $z_c$ for every input. Flipping uncovered shortcuts never changes a prediction.
\item If $\sal_c<\sum_U\sal_j$, the input with $z_c=1$ and all of $U$ flipped has a negative logit, while the input with all of $U$ aligned has a positive one. Unflipping uncovered shortcuts one at a time, some single flip changes the sign. Because $p,\rho_j<1$, that input has positive probability under $P$. A randomizing edit flips the coordinate with probability $\tfrac12$, so the prediction-change rate of that edit is positive.
\end{itemize}
Combining with the ranking and \cref{cor:suff}, the flip rule stops exactly at $r=m^\star$.

\emph{Max-margin counterexample.} Let $a=1$ and $b_1=b_2=\sqrt{0.8}$, so $\sal_c=1<1.6=\sal_1+\sal_2$. Under $P$ (with $p=\rho_j=1$) every input is fully aligned. Its logit is proportional to $2.6$, and after one flip to $1>0$. No single edit changes a prediction, so the flip rule stops at $K=\emptyset$, although $G_{\{1,2\}}$ is misclassified. Rule (ii) does not: each edit changes the logit by $2\sal_j/\kappa>0$ with probability $\tfrac12$.
\end{proof}

\subsection{Proof of \cref{thm:cert}}

\begin{proof}
Fix $t$, and let $X'$ be an edit of $X$ on the event $V_t$. Since the edit preserves the label,
$\1[f(X')\ne Y]\le\1[f(X)\ne Y]+\1[f(X')\ne f(X)]$.
Take expectations conditional on $V_t$. By \cref{ass:accept}, the first term has expectation $\err_P(f)$, so $\err_{Q_t}(f)\le\err_P(f)+\Gamma_t(f)$. Mixing over $\pi$ gives
$\err_Q(f)\le\err_P(f)+\max_t\Gamma_t(f)$ for every $Q\in\mathcal Q$.

Next, $\Gamma^{\rm acc}_t=(1-\delta_t)\Gamma_t+\delta_t\Gamma^{\rm bad}_t\ge(1-\delta_t)\Gamma_t$, where $\Gamma^{\rm bad}_t$ is the prediction-change rate on false accepts. Hence $\Gamma_t\le\Gamma^{\rm acc}_t/(1-\delta_t)$.

Three high-probability events complete the proof. We condition on $f$ and the edit family throughout. Each event involves only one sample, so the union bound needs no independence between the samples. It does need each sample to be independent of $f$ and of the edit family. A sample that was used to choose either one, for example by discovery, is not valid here.
\begin{itemize}[leftmargin=*,itemsep=0pt,topsep=2pt]
\item by Hoeffding's inequality, $\err_P\le\widehat\err_P+\varepsilon^{\alpha/3}_{n_{\rm v}}$ with probability at least $1-\alpha/3$;
\item by Hoeffding's inequality and a union bound, $\Gamma^{\rm acc}_t\le\widehat\Gamma^{\rm acc}_t+\varepsilon^{\alpha/3\tau}_{n_t}$ for all $t$ with probability at least $1-\alpha/3$;
\item by the Clopper--Pearson construction and a union bound, $\delta_t\le\bar\delta_t$ for all $t$ with probability at least $1-\alpha/3$.
\end{itemize}
On the intersection, which has probability at least $1-\alpha$, \eqref{eq:cert} holds for all $Q\in\mathcal Q$ simultaneously, because the bound does not depend on $\pi$.

\emph{Necessity of \cref{ass:accept}.} Without it the bound can fail. If the validator accepts edits only on the inputs where $f$ errs, and the edits never change $f$'s prediction, then $\err_{Q_t}=1$ while the right-hand side is about $\err_P$.
\end{proof}

\subsection{Non-identifiability from observational data}
\label{app:impossible}

\begin{proposition}
Let $k=1$, $p,\rho_1\in(\tfrac12,1]$, and $a,b_1>0$. Let $M_1$ declare coordinate $1$ the core and coordinate $2$ the shortcut, and let $M_2$ declare the reverse, with the same law $P(X,Y)$. For every classifier $f$, the accuracies of the flip group under $M_1$ and under $M_2$ sum to $1$. In particular one of them is at most $\tfrac12$.
\end{proposition}
\begin{proof}
The flip group of $M_1$ consists of the inputs $(ay,-b_1y)$ labelled $y$. The flip group of $M_2$, in which coordinate $2$ is the core, consists of the inputs $(-ay,b_1y)$ labelled $y$. These are the same two inputs, $(a,-b_1)$ and $(-a,b_1)$, with opposite labels in the two models. Each input is therefore classified correctly in exactly one model. Since each group puts mass $\tfrac12$ on each input, the two accuracies sum to $1$. This is the folklore non-identifiability underlying, e.g., \citet{lin2022zin}.
\end{proof}

\section{Figures for E1 and E2}
\label{app:mainfigs}
These figures and \cref{alg:peel} belong to \cref{sec:method,sec:exp}. They were moved here to save space.

\begin{algorithm}[h!]
\caption{\peel}
\label{alg:peel}
\small
\begin{algorithmic}[1]
\REQUIRE data $D$, candidate edit types $\mathcal C$, null edits $\mathcal N$, validator $V$, audit size $m$, $\delta_{\max}$, level $\alpha$
\STATE $f\gets$ ERM on $D$;\; $B\gets\emptyset$
\REPEAT
  \FOR{$t\in\mathcal C\setminus B$}
    \STATE $E_t\gets$ validator-accepted edits of type $t$ on held-out inputs
    \STATE $\widehat\Delta_t\gets$ mean $|$change of $f$'s logit margin$|$ on $E_t$
  \ENDFOR
  \STATE $S\gets\{t:\widehat\Delta_t>\max_{u\in\mathcal N}\widehat\Delta_u$ (one-sided Welch test, Holm, level $\alpha$)$\}$
  \STATE audit $m$ edits per $t\in S$;\; drop $t$ with $\bar\delta_t>\delta_{\max}$
  \STATE $B\gets B\cup S$;\; $f\gets$ train on $D$ with edits from $B$
\UNTIL{$S=\emptyset$}
\RETURN $f$, bank $B$, certificate \eqref{eq:cert} from a fresh split and audit not used above
\end{algorithmic}
\end{algorithm}

\begin{figure}[h!]
\centering
\includegraphics[width=0.72\linewidth]{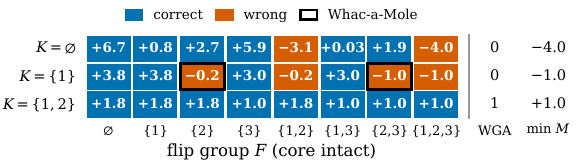}\\[-1pt]
{\small (a)}\\
\includegraphics[width=0.62\linewidth]{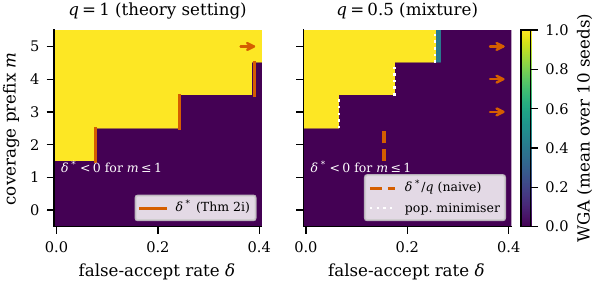}\\[-1pt]
{\small (b)}
\caption{E1 (stylized model). (a) One instance ($p=0.8$, $\rho=(0.95,0.88,0.60)$) under three covered sets $K$. Color: outcome of the fitted model on each core-intact flip group, identical in all 10 seeds. Number: the margin $\Mmarg_K(F)$ of \cref{thm:partial}, whose sign predicts the outcome in all 24 cells. Covering $S_1$ breaks groups $\{2\}$ and $\{2,3\}$ (outlined; Whac-a-Mole, \cref{cor:wam}(b)). WGA stays $0$, but the minimum margin rises ($-4.0\to-1.0$; \cref{cor:wam}(a)), and $K=\{1,2\}$ fixes every group (\cref{cor:suff}). (b) Worst-group accuracy over coverage prefix $m$ and false-accept rate $\delta$. Orange marks are $\delta^\star$ from \cref{thm:validator}(i). The right panel shows that half-edited training ($q=0.5$) is far less tolerant.}
\label{fig:e1}
\end{figure}

\begin{figure}[h!]
\centering
\includegraphics[height=1.75in]{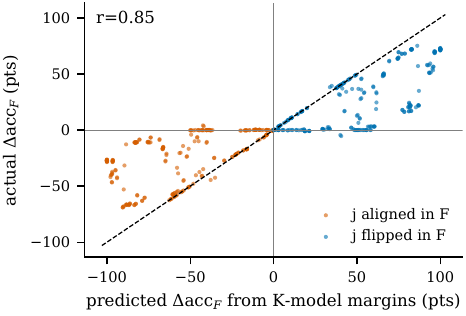}\hspace{0.04\textwidth}
\includegraphics[height=1.75in]{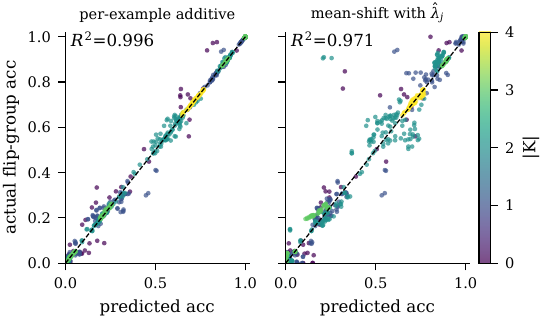}
\caption{E2 (ResNet-18, four shortcuts). Left: change in flip-group accuracy when shortcut $j$ is covered, predicted from per-example margins vs.\ observed ($r=0.85$). Covering an aligned shortcut (orange) lowers accuracy, as in \cref{cor:wam}(b). Right: flip-group accuracy predicted from single-flip effects vs.\ observed, using per-example effects ($R^2=0.996$) or the mean salience $\hat\lambda_j$ ($R^2=0.971$); color is $|K|$.}
\label{fig:e2}
\end{figure}
\FloatBarrier

\IfFileExists{sections/appendix_e4.tex}{\section{E4 and Additional Experiments: Standard Benchmarks, Mistake-Driven Discovery, CSI, Transfer, Scaling, Text}\label{app:e4}

This appendix documents \cref{sec:e4}, the transfer paragraph of \cref{sec:e2}, and the experiments with the learned-adversary baseline (CSI), backbone scaling and text. Accuracies are in \%, mean$\pm$std over seeds. No group labels are used for training or model selection unless a method is marked~$^\dagger$. ``Last epoch'' means no model selection. Graphics processing unit (GPU) time is wall-clock on one NVIDIA A100 (40\,GB) unless stated otherwise. All audits are automatic (ground truth where the data are compositional, otherwise a second model); no human audit was possible.

\newcommand{\apm}[1]{{\scriptsize$\pm$#1}}

\subsection{Standard benchmarks}\label{app:e4-bench}

\paragraph{Datasets and splits.}
\begin{itemize}[leftmargin=*,itemsep=1pt,topsep=2pt]
\item \emph{Waterbirds} \citep{sagawa2020distributionally}: the official tarball, $4{,}795/1{,}199/5{,}794$ train/val/test images, official groups. Images are stored at $256\times256$ and evaluated with a $224$ center crop. WGA is over the 4 test groups; ``avg'' weights the test-group accuracies by the training group frequencies. Exact edits for \peel{} are built from the generative process: the CUB-200-2011 segmentation of the same bird and Places365 images of the same four categories (land: bamboo forest, broadleaf forest; water: ocean, natural lake). The background swap keeps $m\odot x$ and replaces $(1-m)\odot x$ by a fresh Places image of a uniformly random type. Core swap (label-changing; used only as the false-accept candidate in the audit) pastes a bird of the other class onto a fresh image of the original background type. The official composites used aspect-preserving Places images while the $256$\,px Places images are squashed, so edited backgrounds are real Places images of the right category with slightly different geometry.
\item \emph{CelebA}: the official aligned images and attributes, $162{,}770/19{,}867/19{,}962$; $y=$ Blond\_Hair, spurious attribute Male. Gender is not a separable layer, so no exact edit library exists and hand-list \peel{} is not applicable.
\item \emph{Spawrious} O2O-hard and M2M-hard \citep{lynch2023spawrious}: the official archives, splits built exactly as the official \texttt{build\_combination} (total $3{,}168$). There is no official validation split, so $20\%$ of the training set (stratified by class$\times$location$\times$environment, seed 0) is held out as an in-distribution validation set (O2O-hard: $20{,}280/5{,}064/25{,}344$). Metric: average test accuracy (the Spawrious metric), plus worst-class and worst (class$\times$location) group accuracy. No masks are shipped, so hand-list \peel{} is not applicable.
\item \emph{UrbanCars} \citep{li2023whac}: rebuilt with the official generator (Stanford Cars with the official body-type mapping, the 10 official Places365 categories and filter lists, LVIS co-occurring objects on COCO-2017 with the official categories and area thresholds, official seeded orderings, car scale $0.5$, co-object scale $0.25$), $4{,}000/500/500$ per class, training background and co-object ratios $0.95/0.95$, balanced val/test ($125$ per $(y,\text{bg},\text{co})$ cell). Images are stored as lossless layers so that background and co-object swaps are exact. Deviations: car masks from the Hugging Face port of the same MaskFormer Swin-L COCO-panoptic model (official car-picking rule); 19 of $10{,}000$ cars with no detected car replaced by the next unused image in the official order; lossless layers instead of JPEG composites; the background-only augmentation set was not generated. Metrics follow the Whac-a-Mole code: in-distribution (I.D.)\ accuracy weights the class-averaged cell accuracies by the training ratios; background (BG), co-occurring object (CoObj) and BG+CoObj gaps are the accuracy on the corresponding conflicting cells minus I.D.\ accuracy; WGA is over the 8 test groups. Exact edits: background swap, co-object swap, and car swap (label-changing; audit candidate only).
\end{itemize}
The Waterbirds-2S column of \cref{tab:main} comes from E3 (\cref{app:e3}; 5 seeds).

\paragraph{Protocol (fixed; no per-method tuning).} Backbone: torchvision ResNet-50 (IMAGENET1K\_V1) with a new linear head, bfloat16 (bf16) autocast, training augmentation RandomResizedCrop($224$, scale $(0.7,1)$) plus horizontal flip, evaluation on the center crop. One optimizer per dataset for every method: stochastic gradient descent (SGD; momentum $0.9$), cosine schedule, learning rate $10^{-3}$, weight decay $10^{-4}$; Waterbirds 100 epochs (batch 64), CelebA 20 epochs (batch 128), UrbanCars 100 epochs (batch 128), Spawrious 15 epochs (batch 64). Every method gets the same number of SGD steps. Annotation-free methods (ERM, EIIL, SPARE, \peel, CSI) use the last epoch; $^\dagger$ methods select the epoch by worst-group accuracy on the group-labelled validation set. Seeds 0, 1, 2.
\begin{itemize}[leftmargin=*,itemsep=1pt,topsep=2pt]
\item GroupDRO$^\dagger$: $\eta=0.01$, groups $=$ class $\times$ spurious attribute(s).
\item IRM$^\dagger$ (IRMv1, DomainBed defaults): $\lambda=100$ after 500 annealing steps; environments $=$ spurious attribute (Waterbirds: place; CelebA: male; UrbanCars: the 4 (bg, co-object) environments; Spawrious: the two native training environments).
\item JTT$^\dagger$: identification model $=$ the ERM epoch-1 checkpoint ($T=1$), $\lambda_{\rm up}=100$ (50 on CelebA).
\item DFR$^\dagger$: DFR$^{\rm Val}_{\rm Tr}$, $\ell_1$ logistic regression on the final ERM features, group-balanced validation subsets, $C\in\{1,.7,.3,.1,.07,.03,.01\}$ tuned on one half, weights averaged over 20 subsets.
\item EIIL \citep{creager2021environment}: reference model $=$ ERM epoch 1, per-class soft environment assignment maximizing the IRMv1 penalty (Adam $10^{-3}$, 10k steps), two hard environments $\to$ GroupDRO with an environment-balanced sampler; no group labels.
\item SPARE \citep{yang2024identifying}: within-class $k$-means on early-epoch logits (penultimate features on CelebA), $k\in\{2,\dots,6\}$ by silhouette, separation epoch and sampling power from the paper's Table~5, importance sampling $p_i\propto(1/|\text{cluster}|)^{\lambda}$ within class.
\item ShortcutProbe \citep{zheng2025shortcutprobe}: last-layer mode on the final ERM features; the validation set is split into two halves; half 1 trains $K=2$ shortcut base vectors (semantic regularization 5, 50 epochs) and the head is retrained with $5\cdot$CE$_{\rm main}/$CE$_{\rm shortcut}$ (CE: cross-entropy); selection by plain accuracy on half 2. It uses held-out \emph{class} labels but no group labels.
\item CSI (certified shortcut immunization) and CSI-repaired: the learned-adversary baseline and its repaired variant (\cref{app:e4-csi}).
\item Sebra (official code, UrbanCars only): reported at the last epoch and with the official early stopping on group-labelled validation WGA.
\item \peel{}: the E3 protocol (\cref{app:e3}). Library $=$ exact shortcut edits plus the label-changing core swap; nulls $=$ \{horizontal flip, color jitter, random resized crop\}; $\widehat\Delta$ (mean absolute logit-margin change) on 1{,}000 validation images; one-sided Welch test against the strongest null, Holm, $\alpha=0.05$; a simulated validator accepts core swaps with probability $0.1$; audit $m=50$, Clopper--Pearson $\bar\delta$, drop if $\bar\delta>0.15$; Learn $=$ retrain from ImageNet weights with $q=1$, CE on the edited copy $+\,1.0\cdot$KL; at most 5 rounds. \peel-head runs the same loop but retrains only the last layer on the frozen ERM backbone (learning rate $10^{-2}$, 20 epochs).
\end{itemize}

\paragraph{Deviations.}
\begin{itemize}[leftmargin=*,itemsep=1pt,topsep=2pt]
\item \emph{SPARE} has no released code; we re-implemented it from the paper. With the common optimizer the early-epoch logits do not separate the minority groups, so we also run it with the paper's optimizer (Waterbirds: learning rate $10^{-4}$, weight decay $0.1$, batch 128, 300 epochs; CelebA: $10^{-5}$, $1.0$, 20 epochs instead of 50 for budget).
\item \emph{ShortcutProbe}: the released command crashes (the default code path references undefined variables). We patched it by re-implementing the shortcut-walk path from the official code.
\item \emph{Spawrious validation split}: our in-distribution $20\%$ hold-out (above), since none is provided.
\item \emph{JTT schedule}: the published Waterbirds identification schedule ($T=60$, weight decay $1.0$, learning rate $10^{-5}$, 300 epochs) was not affordable; we use $T=1$ with the common optimizer, as JTT and Whac-a-Mole do for CelebA and UrbanCars. Our JTT$^\dagger$ on Waterbirds ($80.6$) is therefore below the published $86.7$.
\item \emph{Sebra}: its CelebA trainer targets a different multi-attribute CelebA variant and the release does not support Waterbirds or Spawrious, so it was run on UrbanCars only.
\item No baseline received a hyper-parameter search; numbers can be below published ones that tuned with group labels. Spurious ImageNet was not run (no ImageNet copy on the cluster).
\end{itemize}

\paragraph{Results.} \Cref{tab:e4-wb,tab:e4-celeba,tab:e4-uc,tab:e4-spaw} give every method on every benchmark. Baselines land near published values: ERM WGA is $74.1$ on Waterbirds and $44.8$ on CelebA; on UrbanCars ERM has I.D.\ $98.5$ and gaps $-8.3/-10.1/-61.4$ (Whac-a-Mole reports $97.6/-15.3/-11.2/-69.2$). \peel{}'s discovery is identical across seeds. On Waterbirds, round~1 keeps the background swap ($\widehat\Delta=5.32$--$5.56$ against at most $1.65$ for the nulls) and the audit drops core swap; round~2 keeps nothing. On UrbanCars, round~1 keeps both the background ($\widehat\Delta=4.94$--$5.15$) and the co-object swap ($4.37$--$4.62$; nulls at most $1.80$) and drops car swap; round~2 keeps nothing. The banks are those of the flip-rate test of our first runs in every seed. Exact-edit sensitivity collapses (Waterbirds background-swap flip rate $13.1\to1.5\%$; UrbanCars background/co-object $22.9/18.0\to1.0/0.3\%$). On UrbanCars, methods that fix one shortcut amplify the other: JTT$^\dagger$ closes the BG gap ($-1.2$) but widens the CoObj gap to $-16.4$ (ERM $-10.1$), and Sebra does the same ($-3.8$ and $-14.8$). \peel{}, MD-\peel{} (\cref{app:e4-md}), DFR$^\dagger$ and ShortcutProbe keep every gap within 5 points; only \peel{} and MD-\peel{} bring all three near 0 with I.D.\ above $97$. On Spawrious no annotation-free method beats ERM, while GroupDRO$^\dagger$ (O2O) and DFR$^\dagger$ (M2M) do. CSI rows are discussed in \cref{app:e4-csi}.

\begin{table}[h!]
\centering
\caption{Waterbirds (ResNet-50, 3 seeds). Groups LL/LW/WL/WW: landbird on land/water, waterbird on land/water. Flip rate: fraction of test predictions changed by an exact background swap. h: wall-clock hours per seed. ``val cls'': uses held-out class labels (no group labels). MD-\peel{} rows: \cref{app:e4-md}; its time includes discovery.}
\label{tab:e4-wb}
\small
\setlength{\tabcolsep}{4pt}
\begin{tabular}{llcccccc}
\toprule
Method & labels & avg acc & WGA & LL/LW/WL/WW & bg flip rate & h \\
\midrule
ERM & -- & 98.0\apm{0.1} & 74.1\apm{1.4} & 99.7/86.0/74.1/95.9 & 13.1\apm{0.6} & 0.09 \\
EIIL & -- & 98.0\apm{0.0} & 76.8\apm{1.3} & 99.7/86.3/76.8/95.6 & 12.2\apm{1.2} & 0.10 \\
SPARE (common optimizer) & -- & 97.9\apm{0.0} & 76.4\apm{2.0} & 99.5/83.0/76.4/96.4 & 14.2\apm{0.3} & 0.09 \\
SPARE (paper optimizer) & -- & 97.8\apm{0.1} & 74.4\apm{1.4} & 99.5/87.6/74.4/94.9 & 11.8\apm{0.2} & 0.25 \\
ShortcutProbe & val cls & 94.9\apm{0.3} & 85.6\apm{2.3} & 97.6/96.8/86.2/86.0 & 5.7\apm{0.5} & 0.01 \\
CSI (selected config) & -- & 98.0\apm{0.1} & 74.2\apm{1.2} & 99.6/84.1/74.2/96.2 & 13.9\apm{0.5} & 0.61 \\
CSI-repaired & -- & 97.9\apm{0.2} & 76.9\apm{2.3} & 99.6/83.0/76.9/96.0 & 14.2\apm{1.0} & 0.48 \\
\peel{} (hand list) & -- & 97.9\apm{0.1} & \textbf{94.0}\apm{0.4} & 99.1/98.9/94.4/94.0 & 1.5\apm{0.1} & 0.17 \\
\peel-head & -- & 96.6\apm{0.0} & 87.7\apm{0.3} & 98.9/97.9/87.7/89.1 & 3.4\apm{0.0} & 0.02 \\
MD-\peel{} (exact edits, aug-only) & -- & 97.5\apm{0.1} & 93.5\apm{0.3} & 98.8/98.7/94.4/93.5 & 1.9\apm{0.2} & 0.4 \\
MD-\peel{} (exact edits, + KL) & -- & 97.9\apm{0.1} & \textbf{94.2}\apm{0.1} & 99.0/98.8/94.3/94.3 & 1.5\apm{0.2} & 0.17 \\
\midrule
GroupDRO$^\dagger$ & train+val & 96.5\apm{1.1} & 86.3\apm{1.4} & 97.8/88.0/86.4/94.4 & 9.6\apm{0.8} & 0.09 \\
JTT$^\dagger$ & val & 95.8\apm{0.4} & 80.6\apm{0.8} & 97.2/83.3/80.8/94.1 & 12.9\apm{0.6} & 0.09 \\
DFR$^\dagger$ & val & 93.8\apm{0.2} & 92.2\apm{0.2} & 94.2/92.2/92.9/92.8 & 8.4\apm{0.1} & 0.00 \\
IRM$^\dagger$ & train+val & 97.5\apm{0.2} & 77.1\apm{0.9} & 99.1/78.7/79.6/96.7 & 15.7\apm{0.5} & 0.10 \\
\bottomrule
\end{tabular}
\end{table}

\begin{table}[h!]
\centering
\caption{CelebA (blond hair $\times$ male; ResNet-50, 3 seeds). Hand-list \peel{} is not applicable (no exact edits).}
\label{tab:e4-celeba}
\small
\begin{tabular}{llcccc}
\toprule
Method & labels & avg acc & WGA & non-blond F/M, blond F/M & h \\
\midrule
ERM & -- & 95.6\apm{0.1} & 44.8\apm{3.0} & 96.2/99.4/85.3/44.8 & 0.57 \\
EIIL & -- & 95.3\apm{0.2} & 44.4\apm{1.8} & 96.2/99.3/83.3/44.4 & 0.60 \\
SPARE (common optimizer) & -- & 95.3\apm{0.0} & 45.4\apm{2.2} & 96.2/99.3/83.4/45.4 & 0.59 \\
SPARE (paper optimizer) & -- & 94.4\apm{0.1} & 20.6\apm{2.8} & 97.7/99.8/70.7/20.6 & 0.61 \\
ShortcutProbe & val cls & 88.1\apm{0.5} & 82.4\apm{0.8} & 82.4/92.0/98.4/88.3 & 0.01 \\
CSI (selected config) & -- & 95.7\apm{0.1} & 48.7\apm{1.8} & 95.5/99.2/88.9/48.7 & 1.14 \\
CSI-repaired & -- & 95.7\apm{0.1} & 48.3\apm{2.8} & 95.7/99.2/87.9/48.3 & 1.12 \\
\midrule
GroupDRO$^\dagger$ & train+val & 92.6\apm{0.7} & 87.6\apm{3.8} & 92.2/93.1/93.2/88.3 & 0.59 \\
JTT$^\dagger$ & val & 88.9\apm{0.9} & 49.8\apm{5.8} & 90.6/96.7/61.6/49.8 & 0.61 \\
DFR$^\dagger$ & val & 91.1\apm{0.2} & \textbf{89.9}\apm{0.1} & 89.9/91.3/95.3/91.7 & 0.01 \\
IRM$^\dagger$ & train+val & 95.5\apm{0.1} & 34.3\apm{0.9} & 95.4/99.6/87.4/34.3 & 0.64 \\
\bottomrule
\end{tabular}
\end{table}

\begin{table}[h!]
\centering
\caption{UrbanCars (ResNet-50, 3 seeds; Whac-a-Mole metrics). Gaps are accuracy on the conflicting cells minus I.D.\ accuracy (0 is best). Flip rates: background / co-object exact swaps.}
\label{tab:e4-uc}
\small
\setlength{\tabcolsep}{3.5pt}
\begin{tabular}{llcccccc}
\toprule
Method & labels & I.D.\ acc & BG gap & CoObj gap & BG+CoObj gap & WGA & flip rate bg/co \\
\midrule
ERM & -- & 98.5\apm{0.0} & $-$8.3\apm{0.5} & $-$10.1\apm{0.3} & $-$61.4\apm{0.8} & 20.0\apm{1.3} & 22.9/18.0 \\
EIIL & -- & 98.0\apm{0.3} & $-$7.2\apm{0.7} & $-$10.1\apm{1.7} & $-$55.9\apm{2.1} & 25.1\apm{2.3} & 21.2/17.2 \\
SPARE (common optimizer) & -- & 98.3\apm{0.4} & $-$8.4\apm{1.1} & $-$8.7\apm{0.8} & $-$59.1\apm{0.2} & 22.9\apm{1.4} & 22.8/16.8 \\
SPARE (paper optimizer) & -- & 98.0\apm{0.3} & $-$7.3\apm{0.8} & $-$6.1\apm{0.4} & $-$45.4\apm{0.8} & 32.3\apm{2.6} & 19.2/12.2 \\
ShortcutProbe & val cls & 94.3\apm{0.5} & $-$3.1\apm{1.2} & $-$3.3\apm{1.4} & $-$2.9\apm{1.1} & 88.5\apm{0.4} & 5.6/3.0 \\
Sebra (last epoch) & -- & 91.9\apm{1.7} & $-$3.8\apm{2.6} & $-$14.8\apm{8.0} & $-$23.2\apm{7.3} & 60.3\apm{12.5} & -- \\
CSI (selected config) & -- & 97.9\apm{0.3} & $-$8.0\apm{1.0} & $-$8.7\apm{2.3} & $-$52.3\apm{2.0} & 24.0\apm{7.9} & 21.1/15.5 \\
CSI-repaired & -- & 98.3\apm{0.0} & $-$7.9\apm{0.4} & $-$7.2\apm{1.0} & $-$51.6\apm{2.8} & 28.3\apm{4.9} & 20.8/14.4 \\
\peel{} (hand list) & -- & 97.3\apm{0.2} & $-$0.5\apm{0.5} & $-$2.1\apm{0.2} & 0.5\apm{0.3} & 94.1\apm{0.4} & 1.0/0.3 \\
\peel-head & -- & 94.6\apm{0.2} & $-$3.0\apm{1.1} & $-$3.1\apm{0.4} & $-$1.3\apm{0.3} & 89.1\apm{0.8} & 4.2/1.8 \\
MD-\peel{} (exact, aug-only) & -- & 97.0\apm{0.2} & $-$0.2\apm{0.5} & $-$1.9\apm{0.3} & 1.0\apm{0.2} & \textbf{94.1}\apm{1.0} & 1.6/0.4 \\
MD-\peel{} (exact, + KL) & -- & 97.0\apm{0.4} & $-$0.6\apm{0.2} & $-$1.5\apm{0.3} & 1.2\apm{0.5} & 93.9\apm{0.8} & 1.2/0.2 \\
\midrule
Sebra (val-WGA stopping) & val & 92.5\apm{1.8} & $-$4.3\apm{1.8} & $-$12.5\apm{5.6} & $-$21.3\apm{4.6} & 67.7\apm{5.6} & -- \\
GroupDRO$^\dagger$ & train+val & 95.0\apm{0.5} & $-$6.8\apm{0.4} & $-$4.3\apm{1.1} & $-$10.7\apm{1.2} & 80.5\apm{1.6} & 8.6/5.4 \\
JTT$^\dagger$ & val & 92.2\apm{1.4} & $-$1.2\apm{3.3} & $-$16.4\apm{4.7} & $-$16.5\apm{4.1} & 71.2\apm{4.6} & 14.6/13.7 \\
DFR$^\dagger$ & val & 94.2\apm{0.5} & $-$4.8\apm{1.9} & $-$3.4\apm{1.1} & $-$3.6\apm{0.8} & 86.9\apm{1.0} & 6.8/3.4 \\
IRM$^\dagger$ & train+val & 95.4\apm{1.0} & $-$10.9\apm{4.1} & $-$6.1\apm{1.3} & $-$26.5\apm{5.7} & 66.7\apm{6.4} & 14.0/8.7 \\
\bottomrule
\end{tabular}
\end{table}

\begin{table}[h!]
\centering
\caption{Spawrious O2O-hard and M2M-hard (ResNet-50, 3 seeds). avg $=$ test accuracy (the official metric); WGA over class$\times$location test groups. On O2O-hard worst-class accuracy equals WGA for every method and is omitted. Hand-list \peel{} is not applicable (no masks).}
\label{tab:e4-spaw}
\small
\begin{tabular}{llccccc}
\toprule
& & \multicolumn{2}{c}{O2O-hard} & \multicolumn{3}{c}{M2M-hard} \\
\cmidrule(lr){3-4}\cmidrule(lr){5-7}
Method & labels & avg & WGA & avg & worst class & WGA \\
\midrule
ERM & -- & 79.6\apm{0.6} & 55.1\apm{4.1} & 61.3\apm{0.5} & 34.9\apm{1.7} & 34.1\apm{2.3} \\
EIIL & -- & 78.8\apm{0.8} & 50.5\apm{5.4} & 61.7\apm{0.0} & 30.8\apm{4.7} & 29.9\apm{4.5} \\
SPARE (common optimizer) & -- & 79.1\apm{1.5} & 49.7\apm{8.0} & 58.5\apm{1.3} & 25.3\apm{5.1} & 24.3\apm{5.8} \\
ShortcutProbe & val cls & 79.6\apm{0.6} & 55.1\apm{4.0} & 60.6\apm{0.7} & 34.3\apm{1.1} & 32.7\apm{1.3} \\
CSI (selected config) & -- & 79.5\apm{0.6} & 55.1\apm{4.1} & 61.3\apm{0.5} & 34.9\apm{1.7} & 34.1\apm{2.3} \\
CSI-repaired & -- & 79.5\apm{0.6} & 55.1\apm{4.1} & 61.3\apm{0.5} & 34.9\apm{1.7} & 34.1\apm{2.3} \\
\midrule
GroupDRO$^\dagger$ & train+val & \textbf{89.3}\apm{0.7} & 80.1\apm{3.6} & 65.5\apm{0.6} & 41.8\apm{0.9} & 40.2\apm{1.3} \\
JTT$^\dagger$ & val & 79.7\apm{1.2} & 53.7\apm{4.5} & 63.5\apm{0.6} & 28.1\apm{4.0} & 26.6\apm{4.9} \\
DFR$^\dagger$ & val & 87.6\apm{0.4} & 71.9\apm{1.2} & \textbf{72.2}\apm{1.1} & 45.3\apm{2.4} & 44.1\apm{1.9} \\
IRM$^\dagger$ & train+val & 75.1\apm{1.9} & 47.0\apm{7.4} & 66.0\apm{1.0} & 38.7\apm{1.3} & 35.4\apm{1.6} \\
\bottomrule
\end{tabular}
\end{table}
\FloatBarrier

\subsection{MD-\peel: discovery from the model's own mistakes}\label{app:e4-md}

\subsubsection{Pipeline as run}
\begin{itemize}[leftmargin=*,itemsep=1pt,topsep=2pt]
\item \emph{Classifiers.} The existing ResNet-50 ERM checkpoint of each seed (E3 protocol for Waterbirds-2S, \cref{app:e3}; the protocol above for the five benchmarks).
\item \emph{Trouble and history sets.} The trouble pool is each dataset's validation split, never trained on (Spawrious: the $20\%$ in-distribution hold-out). Variant L: per class, the misclassified examples plus the lowest-margin correct ones, up to 200 per class. Variant U (no labels): per \emph{predicted} class, the lowest-confidence examples plus those on which the ERM of another seed disagrees. History: the 500 training examples per class with the highest correct margin.
\item \emph{Describe.} Qwen2.5-VL-7B-Instruct (vLLM, greedy) captions every image with the prompt ``Describe this image in one sentence: the main object, the background/setting, any visible text or watermark, and other notable attributes.'' The concept vocabulary is the Natural Language Toolkit (NLTK) nouns and adjectives with document frequency $\ge5$ (220--325 concepts per dataset). Each concept is scored by CLIP ViT-L/14 (OpenAI weights) image--text similarity.
\item \emph{Contrast.} Per class and concept, the history-minus-trouble difference of CLIP scores is tested with a two-sided permutation test ($10^5$ permutations) and Holm correction (9--268 significant concepts per class). In parallel, Qwen2.5-7B-Instruct summarizes differences between 50 history and 50 trouble captions in both directions (``List up to 5 differences in the setting or context (not the identity or parts of the main object) that are common in group A but rare in group B.''). Qwen2.5-32B-Instruct-AWQ merges both lists into at most 12 factor hypotheses per class and keeps the 8 with the largest directional CLIP effect. The same LLM writes the REMOVE/ADD instructions, applies the \emph{core filter} (``property of the object itself'' vs.\ ``context''; only context goes on), and, for the exact editor, maps each hypothesis to one operation $\{$bg-swap, wm-toggle, coobj-swap$\}$ or ``none''. No LLM sees a class name that names the shortcut (``waterbird'', ``urban car''); only the core filter gets a neutral description of the object. The merge LLM (7B $\to$ 32B), the 8-of-12 ranking and the mapper output were fixed during development runs on Waterbirds seed 0, before any scored result; keywords and statistics were not changed.
\item \emph{Test (exact editor).} REMOVE on 100 history images and ADD on 100 trouble images per class. The statistic is the change in $f$'s logit margin toward the class. Null edits: brightness $\times U[0.9,1.1]$ and Gaussian blur ($\sigma=1$). A hypothesis is confirmed if its REMOVE drop or its ADD rise beats both nulls (one-sided Welch), with the expected sign, after Holm across the testable hypotheses of the file. These pixel nulls turned out to be too weak; after the main runs we added a same-operation placebo null (\cref{tab:md-placebo-null}).
\item \emph{Audit.} For exact edits the ground truth is known (they never touch the object), giving $\bar\delta=0.058$. For generative edits see \cref{app:e4-gen}.
\item \emph{Fix.} The union of the confirmed operations is randomized on every training image ($q=1$), with CE on the edited copy only (augmentation, no consistency KL, because E3 showed that KL amplifies false accepts), from ImageNet weights with the dataset's protocol, last epoch. Rounds repeat on the fixed model until no new operation is confirmed (at most 3). A KL ablation ($\lambda_{\rm cons}=1$, the original \peel{} objective) is run on the final bank.
\end{itemize}
CelebA and Spawrious have no exact editor; their hypotheses are tested with the generative editor (\cref{app:e4-gen}).

\paragraph{Pre-registered scoring.} A confirmed hypothesis \emph{hits} a shortcut if its text contains a keyword of that shortcut (case-insensitive, word level), using lists fixed before discovery. Background (Waterbirds, Waterbirds-2S): water, ocean, sea, lake, beach, wave(s), river, pond, shore, coast, forest, tree(s), bamboo, wood(s), grass, land, field, jungle, vegetation. Watermark: watermark, text, logo, copyright, letter(s), writing, word(s), caption, overlay. UrbanCars background: city, urban, street, road, building(s), alley, downtown, crosswalk, parking, garage, gas station, country, countryside, rural, farm, field, forest, nature, dirt, grass. UrbanCars co-object: the UrbanCars co-object category names plus ``animal'', ``sign'', ``hydrant''. CelebA gender: man, men, male, boy, guy, gentleman, beard, mustache, moustache, woman, women, female, girl, lady, makeup, lipstick. Spawrious location: desert, sand, dune, jungle, forest, tree(s), snow, ice, winter, beach, mountain(s), rock(s), dirt, grass, field. The \emph{strict} rule matches plurals only where listed; the \emph{lenient} rule (the pre-registered secondary rule) strips a plural -s/-es first.

\subsubsection{What the trouble sets contain}
Without group labels, the trouble sets are strongly enriched for shortcut-conflicting images (\cref{tab:md-sets}). The exception is Spawrious M2M-hard, where each class has two training locations at 50/50 and the validation pool is in-distribution. Its pool contains no test-like (unseen-location) images and almost no errors (0--5 per class).

\begin{table}[h!]
\centering
\caption{Fraction of shortcut-conflicting images in the trouble set $T_c$ (Variant L and U) and the history $H_c$, round 1, mean over seeds 0--2. Reporting only; never used by the method. CelebA: fraction male. Spawrious: fraction not in the class's most frequent training location.}
\label{tab:md-sets}
\small
\begin{tabular}{lllcccc}
\toprule
dataset & class & attribute & $T_c$ (L) & $T_c$ (U) & $H_c$ & misclassified in $T_c$ (L) \\
\midrule
Waterbirds-2S & landbird & bg & 13\% & 13\% & 0\% & 6 \\
 & landbird & wm & 26\% & 29\% & 2\% & 6 \\
 & waterbird & bg & 10\% & 9\% & 0\% & 6 \\
 & waterbird & wm & 15\% & 17\% & 5\% & 6 \\
Waterbirds & landbird & bg & 93\% & 79\% & 0\% & 70 \\
 & waterbird & bg & 64\% & 43\% & 1\% & 44 \\
UrbanCars & urban & bg / co & 75\% / 80\% & 39\% / 45\% & 0\% / 0\% & 143 \\
 & country & bg / co & 77\% / 73\% & 50\% / 45\% & 0\% / 0\% & 76 \\
CelebA & not blond & male & 8\% & 16\% & 82\% & 413 \\
 & blond & male & 27\% & 14\% & 1\% & 489 \\
Spawrious O2O & 4 classes & location & 26--44\% & 27--45\% & 0--2\% & 1--9 \\
Spawrious M2M & 4 classes & location & 49--55\% & 47--56\% & 45--57\% & 0--5 \\
\bottomrule
\end{tabular}
\end{table}

\subsubsection{Discovery}
\Cref{tab:md-rank} lists the ranked round-1 hypotheses of seed 0 on Waterbirds and UrbanCars with their test statistics. \Cref{tab:md-disc} summarizes every dataset and seed, and \cref{tab:md-loop} the rounds of the loop. The CLIP concept contrast is the strongest signal (e.g.\ sea/ocean/harbor/forest on Waterbirds with $|d|\approx1$--2). The 7B caption summarizer alone is noisy (e.g.\ pirate-themed settings, tufa formations), and it never names gender on CelebA because its prompt excludes the identity of the object.

\begin{table}[h!]
\centering
\caption{Ranked round-1 hypotheses, seed 0, Variant L (CLIP + captions), after the core filter (all 16 labelled ``context''). Each hypothesis reads ``Photos in group H/T are more likely to show $\langle$factor$\rangle$''; the factor is quoted verbatim; dir: the group in which it is more common (H $=$ history, T $=$ trouble). $d$: directional CLIP effect (permutation $p=5\cdot10^{-5}$ unless shown). op: the exact operation chosen by the mapper (``--'': none, so untestable). REMOVE: mean margin drop on history images; ADD: mean margin rise on trouble images; in parentheses the two nulls (brightness/blur). Hit: strict keyword rule. Trouble sets: Waterbirds 200/200 (65/49 misclassified), UrbanCars 200/200 (149/70).}
\label{tab:md-rank}
\footnotesize
\setlength{\tabcolsep}{3pt}
\begin{tabular}{lcp{3.8cm}ccccccc}
\toprule
class & rk & factor & dir & $d$ & op & REMOVE (nulls) & ADD (nulls) & Holm $p$ & hit \\
\midrule
\multicolumn{10}{l}{\emph{Waterbirds}} \\
landbird & 1 & waterfront scenes & T & 2.00 & bg & 9.89 (0.01/2.99) & 9.08 (0.04/$-$2.16) & $2.5\mathrm{e}{-34}$ & -- \\
 & 2 & a seaside setting & T & 1.99 & bg & '' & '' & '' & -- \\
 & 3 & a tern & T & 1.69 & -- & & & & -- \\
 & 4 & a forested environment & H & 1.68 & bg & '' & '' & '' & -- \\
 & 5 & vegetation & H & 1.66 & bg & '' & '' & '' & bg \\
 & 6 & a sailboat & T & 1.51 & -- & & & & -- \\
 & 7 & a branch & H & 1.50 & -- & & & & -- \\
 & 8 & bamboo & H & 1.15 & bg & '' & '' & '' & bg \\
waterbird & 1 & a sea background & H & 1.09 & bg & 7.41 (0.01/1.00) & 4.82 ($-$0.01/$-$1.70) & $9.5\mathrm{e}{-29}$ & bg \\
 & 2 & coastal scenes & H & 1.05 & bg & '' & '' & '' & -- \\
 & 3 & forested areas & T & 1.04 & bg & '' & '' & '' & -- \\
 & 4 & a forest background & T & 0.97 & bg & '' & '' & '' & bg \\
 & 5 & snowy, frost-covered forests & T & 0.77 & -- & & & & -- \\
 & 6 & seagulls in flight & H & 0.66 & -- & & & & -- \\
 & 7 & snowy landscapes & H & 0.40 & -- & & & & -- \\
 & 8 & digital alterations and superimpositions & H & 0.25 {\scriptsize($p$=.004)} & -- & & & & -- \\
\midrule
\multicolumn{10}{l}{\emph{UrbanCars}} \\
urban & 1 & rural landscapes & T & 2.33 & bg & 8.37 (0.03/3.33) & 6.75 ($-$0.04/1.52) & $4.6\mathrm{e}{-31}$ & bg \\
 & 2 & animals & T & 2.11 & co & 8.11 (0.06/3.33) & 7.47 ($-$0.06/1.52) & $4.3\mathrm{e}{-28}$ & (co) \\
 & 3 & natural settings & T & 1.83 & bg & as rank 1 & & & -- \\
 & 4 & forests & T & 1.41 & -- & & & & -- \\
 & 5 & buildings & H & 1.11 & -- & & & & bg \\
 & 6 & cloudy skies & T & 1.02 & -- & & & & -- \\
 & 7 & urban elements & H & 0.99 & bg & as rank 1 & & & bg \\
 & 8 & surreal elements & T & 0.95 & -- & & & & -- \\
country & 1 & a countryside setting & H & 2.05 & bg & 8.42 ($-$0.02/4.31) & 6.74 (0.01/$-$2.38) & $5.4\mathrm{e}{-27}$ & bg \\
 & 2 & animals present & H & 1.98 & co & 7.57 ($-$0.02/4.31) & 5.14 (0.02/$-$2.38) & $8.8\mathrm{e}{-25}$ & (co) \\
 & 3 & natural elements & H & 1.93 & bg & as rank 1 & & & -- \\
 & 4 & a garage & T & 1.64 & bg & as rank 1 & & & bg \\
 & 5 & a rural road & H & 1.60 & bg & as rank 1 & & & bg \\
 & 6 & architectural elements & T & 1.57 & -- & & & & -- \\
 & 7 & a forested area & H & 1.27 & -- & & & & -- \\
 & 8 & streetlights & T & 1.22 & -- & & & & -- \\
\bottomrule
\multicolumn{10}{l}{\scriptsize '': same operation and statistics as the row above. (co): co-object hit under the lenient rule only (``animals'' vs.\ listed ``animal'').}
\end{tabular}
\end{table}

\begin{table}[h!]
\centering
\caption{Round-1 discovery, Variant L, strict keyword rule. Candidates: context hypotheses after the filter (of 16; Spawrious of 32). Precision: fraction of hypotheses that hit a true shortcut. Rank: best within-class rank of a hitting hypothesis. Confirmed: by the exact-editor test (only where an exact editor exists). WB2S $=$ Waterbirds-2S.}
\label{tab:md-disc}
\small
\setlength{\tabcolsep}{4pt}
\resizebox{\textwidth}{!}{%
\begin{tabular}{lccccccc}
\toprule
dataset & seed & \#cand. & candidate recall & cand.\ precision & first rank & \#confirmed (precision) & confirmed ops \\
\midrule
WB2S & 0/1/2 & 16/16/16 & bg \checkmark, wm $\times$ (all seeds) & .44/.31/.25 & bg 1/1/2 & 8 (.75) / 9 (.44) / 8 (.50) & bg / bg,wm / bg,wm \\
Waterbirds & 0/1/2 & 16/16/16 & bg \checkmark & .25/.50/.38 & 1/2/1 & 9 (.44) / 11 (.73) / 10 (.60) & bg / bg / bg \\
UrbanCars & 0/1/2 & 16/16/16 & bg \checkmark, co $\times$ (all seeds) & .38/.38/.38 & bg 1/1/1 & 9 (.56) / 13 (.38) / 12 (.50) & bg,co (all) \\
CelebA & 0/1/2 & 12/15/12 & gender \checkmark & .33/.20/.42 & 2/2/2 & generative (\cref{app:e4-gen}) & \\
Spawrious O2O & 0/1/2 & 31/30/32 & location \checkmark & .42/.43/.41 & 1/1/1 & generative & \\
Spawrious M2M & 0/1/2 & 27/27/29 & location \checkmark & .11/.07/.03 & 2/5/8 & generative & \\
\bottomrule
\end{tabular}}
\end{table}

The watermark and co-object operations are confirmed although the strict rule credits no confirmed hypothesis with them: the confirmed watermark and co-object hypotheses are phrased in the plural (``watermarks'', ``watermarks or logos'', ``animals'', ``animals like cows, horses, and goats''), which the strict word-level rule does not match. Many background hypotheses also miss both rules (``forested'', ``coastal'', ``seaside'', ``waterfront'', ``wilderness''). Over all rounds (\cref{tab:md-loop}) the dataset-level result on the exact-editor datasets is, in every seed: Waterbirds hit under both rules; Waterbirds-2S and UrbanCars only \emph{partial} (background) under the strict rule and hit under the lenient rule. Precision of the confirmed hypotheses (strict/lenient, mean over seeds): Waterbirds-2S $0.50/0.64$, Waterbirds $0.61/0.61$, UrbanCars $0.53/0.67$. Combining these with the generative test of seed 0 on CelebA and Spawrious (\cref{app:e4-gen}; Spawrious counts as hit only if both O2O and M2M are hit), seed 0 hits \textbf{2 of 5} datasets under the strict rule and \textbf{4 of 5} under the lenient rule. The pre-registered hypothesis H1 (true shortcut confirmed on $\ge4/5$ datasets with precision $\ge0.5$) is therefore \emph{not supported} under the strict rule and supported under the lenient rule for seed 0. The generative test was run on seed 0 only, so seeds 1 and 2 have exact-editor results only (strict 1/3, lenient 3/3 of those datasets).

\begin{table}[h!]
\centering
\caption{The MD-\peel{} loop with the exact editor (Variant L). Per round: the operations confirmed (\#confirmed/\#context hypotheses) and the WGA after the fix of that round. In rounds $\ge2$, an already-banked operation that is re-confirmed is residual reliance with a much smaller effect (e.g.\ Waterbirds landbird background REMOVE drop $9.9\to1.1$--$1.2$ logits); no new operation means stop.}
\label{tab:md-loop}
\small
\begin{tabular}{llll}
\toprule
dataset & seed & round 1 & rounds 2, 3 \\
\midrule
Waterbirds-2S & 0 & bg (8/16), WGA 93.5 & r2: bg, \textbf{wm} (3/15), WGA 98.1; r3: none (0/15) \\
 & 1 & bg, wm (9/16), WGA 98.1 & r2: none \\
 & 2 & bg, wm (8/16), WGA 97.9 & r2: none \\
 & 3 & bg, wm (6/16), WGA 98.1 & r2: bg (1/15) \\
 & 4 & bg, wm (7/16), WGA 97.9 & r2: bg, wm (2/9) \\
Waterbirds & 0/1/2 & bg (9/11/10 of 16), WGA 93.8/93.5/93.1 & r2: bg (6/7/4 of 16) \\
UrbanCars & 0 & bg, co (9/16), WGA 92.8 & r2: bg (4/16) \\
 & 1 & bg, co (13/16), WGA 95.2 & r2: bg (4/16) \\
 & 2 & bg, co (12/16), WGA 94.4 & r2: none \\
\bottomrule
\end{tabular}
\end{table}

\paragraph{Variant U (no labels).} Using low confidence and disagreement with the ERM of another seed instead of class labels confirms \emph{exactly the same operations} as Variant L on all 9 exact-editor runs (3 datasets $\times$ 3 seeds). Datasets whose true shortcuts are all confirmed in round 1 (of 3; strict/lenient): L $1/2$, $1/3$, $1/3$ and U $1/2$, $2/3$, $2/3$ for seeds 0/1/2. At candidate level over the 6 benchmarks (strict/lenient): L $4/5$, $4/6$, $4/6$ and U $4/5$, $5/6$, $4/5$; the only loss is Spawrious M2M-hard seed 2. H4 (U recall $\ge$ L recall $-1$ dataset) is supported.

\subsubsection{Ablations and the negative control}
\paragraph{Without the test step.} H2 predicted that trusting the contrast list directly would lower precision by $\ge0.2$. \Cref{tab:md-h2} compares the precision of all context candidates (a) with that of the confirmed ones (c). The mean gain is $+0.184$ (strict) and $+0.201$ (lenient) over 9 runs, at the threshold. But the margin test rejected \textbf{0 of 89} testable round-1 hypotheses: the whole gain comes from the mapper declaring hypotheses without an exact operation untestable (sailboats, harbors, seagulls, branches, \dots), not from the statistical test. We therefore count H2 as not supported as a statement about the test. Adding the placebo null below does not change this: every round-1 test is still confirmed, and the test's rejections come in later rounds.

\begin{table}[h!]
\centering
\caption{Precision with and without the test step (round 1, Variant L; strict/lenient).}
\label{tab:md-h2}
\small
\resizebox{\textwidth}{!}{%
\begin{tabular}{lcccccccc}
\toprule
dataset & seed & \#context & \#testable & \#confirmed & \#rejected by test & (a) untested & (c) confirmed & (c)$-$(a) \\
\midrule
Waterbirds-2S & 0 & 16 & 8 & 8 & 0 & .44/.44 & .75/.75 & .31/.31 \\
 & 1 & 16 & 9 & 9 & 0 & .31/.38 & .44/.56 & .13/.18 \\
 & 2 & 16 & 8 & 8 & 0 & .25/.31 & .50/.62 & .25/.31 \\
Waterbirds & 0 & 16 & 9 & 9 & 0 & .25/.31 & .44/.44 & .19/.13 \\
 & 1 & 16 & 11 & 11 & 0 & .50/.50 & .73/.73 & .23/.23 \\
 & 2 & 16 & 10 & 10 & 0 & .38/.38 & .60/.60 & .22/.22 \\
UrbanCars & 0 & 16 & 9 & 9 & 0 & .38/.56 & .56/.78 & .18/.22 \\
 & 1 & 16 & 13 & 13 & 0 & .38/.50 & .38/.54 & .01/.04 \\
 & 2 & 16 & 12 & 12 & 0 & .38/.50 & .50/.67 & .12/.17 \\
\bottomrule
\end{tabular}}
\end{table}

\paragraph{Without the core filter.} On the exact-editor datasets the filter has no effect (every hypothesis was labelled ``context''; identical confirmed operations). On CelebA and Spawrious it removes 0--5 of 16--32 hypotheses, and on CelebA it mislabels ``Blonde hair is more common in group T photos'' as context.

\paragraph{Evidence source.} CLIP-only contrast finds the shortcut at candidate level wherever the full arm does, except Spawrious M2M-hard (0/3 seeds). Captions-only misses gender on CelebA (0/3 seeds) and is otherwise comparable.

\paragraph{Post-hoc placebo negative control.} The brightness/blur nulls are tiny, while history images are selected for extreme high margin and trouble images for extreme low margin, so any large edit tends to lower or raise them. We therefore ran, after the main results, a same-operation placebo: the factor is re-sampled \emph{within its own value} (another background of the same type, the watermark moved but kept, another co-object of the same type). It is ``confirmed'' by the specified rule in \textbf{10 of 30} placebo tests (\cref{tab:md-placebo}), so the test as specified is not specific. Nevertheless, all 26 distinct real round-1 (class, operation, value) edits beat their placebo by a wide margin (REMOVE drop $2.7$--$10.7$ vs.\ $-0.0$--$3.9$ logits; ADD rise $0.8$--$9.5$ vs.\ $0.1$--$2.8$; Holm $p<4\cdot10^{-22}$), so the confirmations in \cref{tab:md-loop} are not placebo artefacts. A same-operation placebo, or the editor-matched nulls of \cref{app:e4-gen}, should therefore be the null, and we re-ran the whole test with one.

\begin{table}[h!]
\centering
\caption{Placebo control (post hoc). Same images, statistic and decision rule as the main test.}
\label{tab:md-placebo}
\small
\resizebox{\textwidth}{!}{%
\begin{tabular}{lccccccc}
\toprule
dataset & seed & placebo tests & placebo ``confirmed'' & placebo REMOVE/ADD & real tests & real $>$ placebo & real REMOVE/ADD \\
\midrule
Waterbirds-2S & 0 & 4 & 2 & 0.60/0.55 & 2 & 2 & 5.55/1.44 \\
 & 1 & 4 & 1 & 0.71/0.65 & 3 & 3 & 4.93/1.42 \\
 & 2 & 4 & 2 & 0.81/0.56 & 3 & 3 & 5.03/1.38 \\
Waterbirds & 0 & 2 & 1 & 1.87/1.10 & 2 & 2 & 8.65/6.95 \\
 & 1 & 2 & 1 & 1.60/1.05 & 2 & 2 & 8.64/7.02 \\
 & 2 & 2 & 1 & 2.32/1.75 & 2 & 2 & 9.14/7.56 \\
UrbanCars & 0 & 4 & 1 & 1.63/0.71 & 4 & 4 & 8.12/6.53 \\
 & 1 & 4 & 1 & 1.55/0.67 & 4 & 4 & 8.24/6.41 \\
 & 2 & 4 & 0 & 1.50/0.78 & 4 & 4 & 8.13/6.62 \\
\midrule
total & & 30 & 10 & & 26 & 26 & \\
\bottomrule
\end{tabular}}
\end{table}

\paragraph{The test with a placebo null.} We re-ran the exact-editor test of every round and seed. These are 26 files, including Waterbirds-2S seeds 3--4 and all round-2 and round-3 tests on the fixed models. The rule is the specified one with the same-operation placebo added to the nulls: a direction passes only if it beats the placebo and both pixel nulls. Images and real edits are those of the main test (\cref{tab:md-placebo-null}).
\begin{itemize}[leftmargin=*,itemsep=1pt,topsep=2pt]
\item \emph{Calibration.} We replace the real edit by a second, independent placebo draw. The specified rule then confirms 14 of 64 tests, and 23\% of directions reach unadjusted $p<0.05$. The placebo rule confirms none of them, and no direction reaches $p<0.05$, so it is conservative.
\item \emph{Round 1.} All 34 distinct real (class, operation, value) tests are still confirmed.
\item \emph{Covered operations on the fixed model.} In rounds 2 and 3 the model has been trained with these operations randomized. The specified rule still confirms 9 of the 28 re-tests, with margin changes of $0.5$--$1.6$ logits, the size of a placebo. The placebo rule confirms none of them. The one operation that first appears in a later round, the watermark in round 2 of Waterbirds-2S seed 0, is confirmed under both rules.
\item \emph{Final banks.} They are unchanged in all 11 runs, so the fixes below are unaffected.
\end{itemize}
Adding a watermark to an image without one has no within-value placebo; the placebo is a no-op there. In every run that confirms the watermark, a direction that \emph{removes} an existing watermark (placebo: moving it) is significant on its own ($p_h\le0.003$). The placebo null was chosen after the results were known. These numbers are therefore a post-hoc check of the specified procedure, not a pre-registered test.

\begin{table}[h!]
\centering
\caption{Exact-editor test under the specified pixel nulls and with the same-operation placebo added (post hoc), over all 26 test files (Waterbirds-2S, Waterbirds, UrbanCars; every seed and round). Calibration: a second placebo draw in place of the real edit. Covered: operations already in the bank, re-tested on the model fixed with them.}
\label{tab:md-placebo-null}
\small
\begin{tabular}{lcc}
\toprule
 & pixel nulls (as specified) & pixel nulls $+$ same-op placebo \\
\midrule
calibration: tests confirmed & 14/64 & 0/64 \\
calibration: directions with $p<0.05$ & 30/128 (23\%) & 0/128 (0\%) \\
round-1 tests confirmed & 34/34 & 34/34 \\
covered re-tests confirmed (rounds 2--3) & 9/28 & 0/28 \\
new later-round operation confirmed & 1/1 & 1/1 \\
final banks unchanged & -- & 11/11 \\
\bottomrule
\end{tabular}
\end{table}

\subsubsection{Fix with the exact (oracle) editor}
With exact edits, MD-\peel{} matches hand-list \peel{} on all three datasets (\cref{tab:md-fix-wb2s}; Waterbirds and UrbanCars in \cref{tab:e4-wb,tab:e4-uc}): $98.0\pm0.1$ vs.\ $98.2\pm0.1$ on Waterbirds-2S (5 seeds), $93.5\pm0.3$ vs.\ $94.0\pm0.4$ on Waterbirds, $94.1\pm1.0$ vs.\ $94.1\pm0.4$ on UrbanCars, with every UrbanCars gap within 2 points. It is above every $^\dagger$ baseline on all three. The KL ablation reproduces \peel{} within seed noise ($98.3/94.2/93.9$). H3 (oracle part: within 2 WGA points of hand-list \peel) is supported.

\begin{table}[h!]
\centering
\caption{Waterbirds-2S fix (ResNet-50, 20 epochs; flip groups over all 5{,}794 test birds). The 3-seed rows and the $^\dagger$ rows use seeds 0--2; the 5-seed rows correspond to \cref{tab:main} and E3. GPU-hours (GPU-h) per seed; MD-\peel{} includes discovery.}
\label{tab:md-fix-wb2s}
\small
\begin{tabular}{lccccccc}
\toprule
method & seeds & IID & $\varnothing$ & $\{$bg$\}$ & $\{$wm$\}$ & $\{$bg,wm$\}$ $=$ WGA & GPU-h \\
\midrule
ERM & 3 & 98.9\apm{0.1} & 99.6 & 91.7 & 97.3 & 65.9\apm{1.1} & 0.02 \\
Random library edits & 3 & 99.2\apm{0.1} & 99.6 & 96.9 & 97.1 & 76.9\apm{0.9} & 0.04 \\
\peel{} hand list (KL) & 3 & 98.5\apm{0.1} & 98.3 & 98.7 & 98.4 & 98.3\apm{0.1} & 0.04 \\
Oracle coverage $\{$bg,wm$\}$, aug-only & 3 & 98.2\apm{0.1} & 98.0 & 98.5 & 98.2 & 98.0\apm{0.1} & 0.04 \\
MD-\peel{} (exact, aug-only) & 3 & 98.2\apm{0.1} & 98.0 & 98.5 & 98.1 & 98.0\apm{0.1} & 0.3 \\
MD-\peel{} (exact, + KL) & 3 & 98.5\apm{0.0} & 98.3 & 98.7 & 98.4 & 98.3\apm{0.1} & 0.03 \\
\midrule
ERM & 5 & 98.9\apm{0.1} & 99.6 & 91.9 & 97.3 & 66.5\apm{1.2} & 0.02 \\
\peel{} hand list (KL) & 5 & 98.5\apm{0.1} & 98.3 & 98.6 & 98.4 & 98.2\apm{0.1} & 0.04 \\
MD-\peel{} (exact, aug-only) & 5 & 98.2\apm{0.1} & 98.0 & 98.5 & 98.1 & 98.0\apm{0.1} & 0.3 \\
\midrule
JTT$^\dagger$ & 3 & 96.8\apm{0.1} & 97.7 & 96.7 & 86.6 & 82.2\apm{3.2} & 0.02 \\
DFR$^\dagger$ & 3 & 96.6\apm{0.1} & 96.7 & 95.0 & 93.7 & 89.7\apm{0.4} & 0.01 \\
GroupDRO$^\dagger$ & 3 & 98.3\apm{0.3} & 98.6 & 93.4 & 96.6 & 85.4\apm{1.6} & 0.02 \\
\bottomrule
\end{tabular}
\end{table}

\paragraph{Cost (exact editor).} Per seed, discovery (selection, captioning, CLIP, LLM; both variants and all four evidence arms) takes $0.27/0.24/0.23$ GPU-h on Waterbirds-2S/Waterbirds/UrbanCars and $0.13$--$0.14$ on CelebA and Spawrious; an exact-editor test takes $<0.01$; the fix is one retraining per round ($0.05/0.17/0.27$). The total is $0.34/0.41/0.51$ GPU-h per seed over $2.3/2.0/2.0$ rounds on average, against $0.04$--$0.27$ for hand-list \peel.

\begin{figure}[h!]
\centering
\includegraphics[height=0.8\textheight]{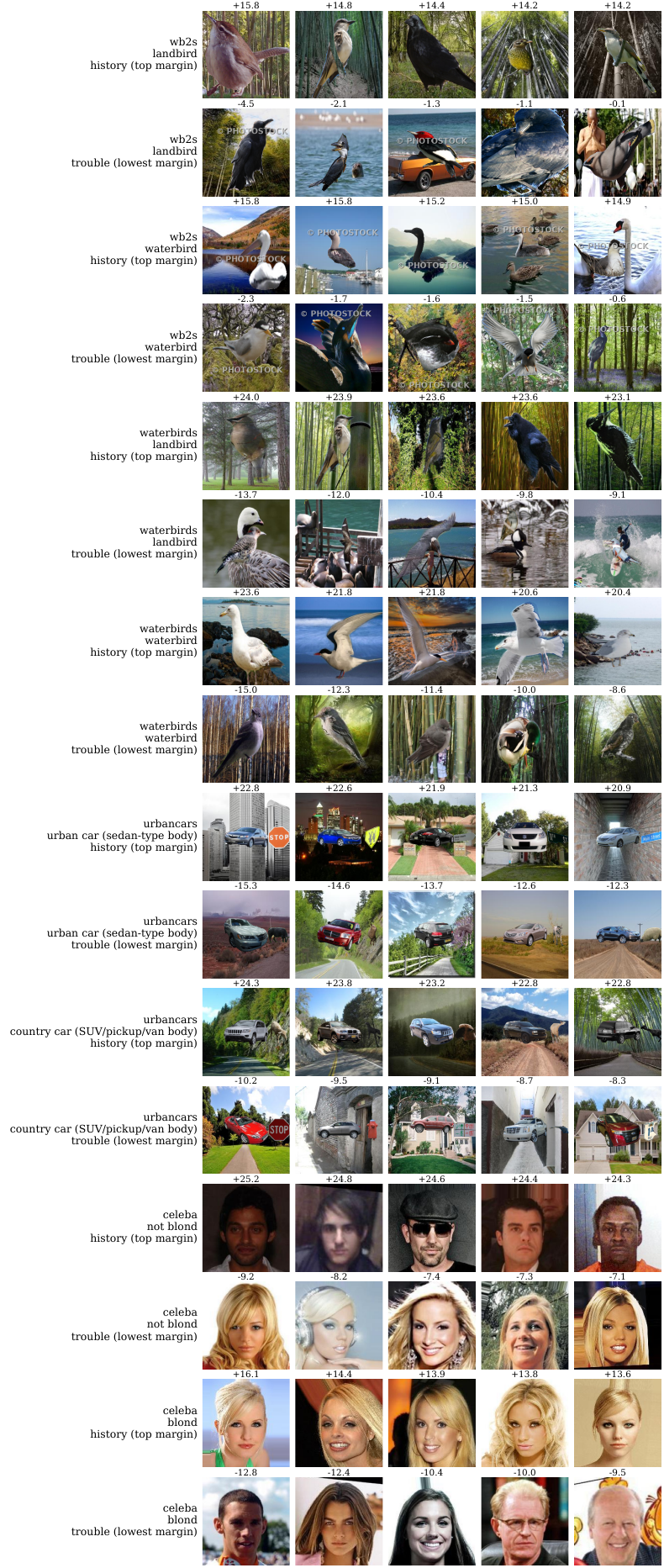}
\caption{MD-\peel{} history (highest correct margin) and trouble (lowest margin) images per class, seed 0, with $f$'s margin above each image. No group labels were used to select them; the trouble sets are dominated by shortcut-conflicting images (Waterbirds landbirds on water, UrbanCars cars in the other context, CelebA blond men).}
\label{fig:md-examples}
\end{figure}
\FloatBarrier

\subsubsection{Generative editor: test, audit and fix without exact edits}\label{app:e4-gen}

\paragraph{Models.} InstructPix2Pix, Stable Diffusion (SD) 1.5 inpainting (half precision, fp16), CLIPSeg (rd64-refined) masks, the validator Qwen2.5-VL-7B (Q1: the main object and its class are unchanged; Q2: the edit was applied), and the audit judges InternVL2.5-8B and CLIP ViT-L/14 on an object crop. To measure edits on UrbanCars we trained ground-truth background and co-object detectors (95.7\% and 98.3\% on test renders). Background and watermark detectors for the bird datasets and the object masks come from E3.

\paragraph{Editor pilot.} Three editors were compared on 200 random validation images per dataset with hand-written canonical shortcut flips (background swaps, watermark toggles, co-object swaps, gender, Spawrious location). These instructions were used only to choose the editor, never for discovery. The pre-registered rule picks the highest mean \emph{yield}, i.e.\ the fraction of edits that are accepted, reach the target attribute according to the ground-truth detectors, and leave the object intact (pixel rule on the object mask), over Waterbirds-2S, Waterbirds and UrbanCars (\cref{tab:gen-pilot}). Inpainting outside the CLIPSeg object mask wins (mean yield $38.8$, vs.\ $28.3$ for InstructPix2Pix with object paste-back and $4.8$ for raw InstructPix2Pix). Raw InstructPix2Pix re-renders the object in 75--81\% of edits, and the VLM validator still accepts many of them ($\delta_{\rm nat}$ among accepted edits $38.5/63.7/81.4\%$). This is the natural-validator failure of \cref{thm:validator}, the same failure as CSI's CLIP validator. With object protection the ground-truth false-accept rate is 0 (Clopper--Pearson upper bound 1.7--2.8\%). The residual problem is edits that do not happen: the validator says ``applied'' far more often than the detectors confirm (Waterbirds inpainting: Q2 ``yes'' $92.5\%$, background reached in $53\%$ of accepted edits). Editing ability depends on the shortcut. Background swaps work (target reached in 48--79\% with inpainting), watermark toggling fails with all three editors (target reached $\le9\%$), and co-object swaps are weak (18--27\%). CelebA gender edits are accepted only 43--44\% of the time; Spawrious location edits 73--86\%.

\begin{table}[h!]
\centering
\caption{Generative editor pilot (200 validation images per dataset; \%). ip2p: InstructPix2Pix; protect: with object paste-back; inpaint: SD inpainting outside the CLIPSeg object mask. Target reached: by the ground-truth detectors, among all / accepted edits. $\delta_{\rm nat}$: object altered among accepted edits (Clopper--Pearson 95\% upper bound). Yield: accepted, target reached and object intact. Second judge: pre-registered definition. UB: upper bound.}
\label{tab:gen-pilot}
\small
\setlength{\tabcolsep}{4pt}
\resizebox{\textwidth}{!}{%
\begin{tabular}{llcccccc}
\toprule
dataset & editor & accept & target reached & object altered & $\delta_{\rm nat}$ (UB) & yield & 2nd judge bad (UB) \\
\midrule
Waterbirds-2S & ip2p & 32.5 & 29.5 / 41.5 & 75.0 & 38.5 (49.4) & 8.0 & 9.2 (17.4) \\
 & protect & 56.0 & 31.5 / 51.8 & 0.0 & 0.0 (2.6) & 29.0 & 4.5 (9.2) \\
 & inpaint & 60.0 & 28.5 / 45.8 & 0.0 & 0.0 (2.5) & 27.5 & 5.0 (9.6) \\
Waterbirds & ip2p & 40.0 & 42.0 / 30.0 & 79.0 & 63.7 (72.7) & 6.0 & 13.8 (21.7) \\
 & protect & 84.0 & 32.0 / 35.1 & 0.0 & 0.0 (1.8) & 29.5 & 8.9 (13.4) \\
 & inpaint & 88.0 & 50.5 / 53.4 & 0.0 & 0.0 (1.7) & 47.0 & 10.8 (15.4) \\
UrbanCars & ip2p & 21.5 & 29.5 / 41.9 & 81.0 & 81.4 (90.4) & 0.5 & 16.3 (28.4) \\
 & protect & 60.0 & 30.0 / 44.2 & 0.5 & 0.0 (2.5) & 26.5 & 10.8 (16.7) \\
 & inpaint & 53.5 & 49.0 / 78.5 & 0.0 & 0.0 (2.8) & 42.0 & 4.7 (9.6) \\
CelebA & ip2p / protect / inpaint & 43.0 / 43.5 / 44.0 & -- & -- & -- & -- & 5.8 / 2.3 / 6.8 \\
Spawrious O2O & ip2p / protect / inpaint & 75.5 / 82.0 / 85.5 & -- & -- & -- & -- & 5.3 / 4.9 / 4.1 \\
Spawrious M2M & ip2p / protect / inpaint & 72.5 / 74.5 / 75.0 & -- & -- & -- & -- & 3.4 / 2.0 / 0.7 \\
\bottomrule
\end{tabular}}
\end{table}

\paragraph{Decisions made after the pilot, before any test result.}
\begin{itemize}[leftmargin=*,itemsep=1pt,topsep=2pt]
\item \emph{Hybrid editor.} Some hypotheses are whole-image factors (lighting, weather, ``digital manipulation'') that inpainting cannot express. An LLM (Qwen2.5-7B-Instruct, neutral few-shot examples, no class names) turns each hypothesis direction into either a \emph{region} edit (SD inpainting of the CLIPSeg mask of the named region, object excluded, with a positive prompt) or a \emph{global} edit (InstructPix2Pix with object paste-back). Images are edited at 512\,px with 20 diffusion steps.
\item \emph{Editor-matched nulls}, because a diffusion re-render moves margins by itself. For region edits the null is a resample of the same region with a prompt describing its current content; for global edits the nulls are ``make the image slightly brighter / slightly blurry'' through the same editor. Every hypothesis must also beat the two pixel nulls of \cref{app:e4-md}.
\item \emph{Audit judge.} The pre-registered ``CLIP crop class $\neq c$'' fires on 2--14\% of \emph{original} images (CLIP's own errors) and would push $\bar\delta$ over $0.15$ for correct hypotheses. The audit therefore flags an edit if the CLIP crop class flips \emph{because of the edit}, or if InternVL2.5 answers ``no'' to an identity question without class names (``is the main bird/car/dog / the person's hair colour unchanged?''). On the pilot, this judge flags 0--3.6\% of the ground-truth-intact accepted edits (CelebA 5.7--12.5\%), and 43--56\% of raw-InstructPix2Pix edits that re-render the object. The second judge's recall for identity changes can only be bounded, not measured.
\end{itemize}

\paragraph{Test, audit and fix as run.} The input is the round-1 hypotheses of MD-\peel's discovery of seed 0 (Variant L, after the core filter): 16 each for the bird datasets and UrbanCars, 12 for CelebA, 31 and 27 for Spawrious O2O and M2M. REMOVE is applied to the 100 highest-margin history images and ADD to the 100 lowest-margin trouble images of the class, with $f=$ the seed-0 ERM. Edits count if the validator accepts them (Q1 and Q2; nulls Q1 only). For each direction, $p$ is the largest one-sided Welch $p$ over its nulls; $p_h=\min(1,2\min(p_{\rm REMOVE},p_{\rm ADD}))$, Holm across the file, and the winning direction must have the expected sign. The audit judges the first 50 accepted edits of the winning direction and drops the hypothesis if $\bar\delta>0.15$. Fix: confirmed hypotheses are grouped into at most 3 factors without labels (same REMOVE region phrase $\to$ same factor; each global hypothesis is its own factor). For every training image and factor, one hypothesis of the group is drawn and ADD or REMOVE is applied with probability $1/2$, chained across factors. Each step is validated against the \emph{original} image, and a rejected step keeps the previous image. Training uses CE on the edited copy only, with the dataset's engine and schedule, last epoch, seeds 0--2 sharing one edited training set; the KL ablation ($\lambda_{\rm cons}=1$) uses seed 0. CelebA uses head-only retraining on the frozen ERM backbone with a random 20k-image training subset. The $\pm$ in \cref{tab:gen-fix} therefore measures retraining variance only, not the variance of discovery and editing. Discovery seeds 1 and 2 were later run end to end (\cref{tab:gen-e2e}), and \cref{tab:main} reports those three runs. Round 2 was \emph{not run}, for time rather than because nothing was left: a round-2 discovery on the fixed Waterbirds-2S model produced 16 background-type hypotheses and no watermark hypothesis, and was not tested.

\paragraph{What the generative test confirms.} \Cref{tab:gen-disc}. Background-type shortcuts are confirmed on every dataset that has them except Spawrious M2M-hard. Many kept hypotheses that miss the strict keywords are background-type too (``seaside setting'', ``waterfront scenes'', ``arid setting'', ``mountainous landscape'', ``streetlights'', ``architectural elements''), so strict precision ($0.25$--$0.56$) understates how much of what is kept is context. On CelebA, 5 hypotheses pass the test and 2 survive the audit: ``Group T photos are more likely to feature a beard'' (a gender hit) and a sports-stadium setting. ``Feature a man'' passes the test (Holm $p=0.003$) but the audit drops it (11/33 edits judged to change the hair colour or the person). Gender edits are hard for the editor (acceptance 13--25\%). On Spawrious M2M-hard, none of the 4 kept hypotheses hits the location keywords. Two are background-related (``lush, tropical background'', ``scattered background'') but use no listed keyword; the other two are the dog ``running''/``standing'', a pose factor that the core filter let through. M2M-hard therefore counts as a miss under the pre-registered scoring. No watermark or text hypothesis was proposed by discovery on seed 0, so there was nothing for the editor to test. The UrbanCars co-object hypothesis ``contain animals'' is confirmed in both classes (a strict miss, a lenient hit). Every UrbanCars REMOVE edit repaints the whole background, which contains the co-object, so the generative fix randomizes the co-object only incidentally.

\begin{table}[h!]
\centering
\caption{Round-1 generative test, seed 0, Variant L. Kept: confirmed by the test and passing the audit. Hits and precision: strict keyword rule (lenient in parentheses where it differs). Precision without the test: over all tested hypotheses. Oracle: the exact-editor test of the same file (\cref{app:e4-md}). Accept: validator acceptance of REMOVE / ADD edits.}
\label{tab:gen-disc}
\small
\setlength{\tabcolsep}{3.5pt}
\begin{tabular}{lcccccccc}
\toprule
dataset & tested & test & kept & hits among kept & precision & w/o test & oracle & accept R/A \\
\midrule
Waterbirds-2S & 16 & 11 & 9 & bg 4; wm 0 & 0.44 & 0.44 & 8/8 & 70 / 66 \\
Waterbirds & 16 & 8 & 8 & bg 2 & 0.25 & 0.25 & 9/9 & 65 / 66 \\
UrbanCars & 16 & 10 & 10 & bg 5 (6); co 0 (2) & 0.50 & 0.38 & 9/9 & 81 / 85 \\
CelebA & 12 & 5 & 2 & gender 1 & 0.50 & 0.33 & -- & 25 / 13 \\
Spawrious O2O & 31 & 9 & 9 & location 5 & 0.56 & 0.42 & -- & 80 / 72 \\
Spawrious M2M & 27 & 4 & 4 & location 0 & 0.00 & 0.11 & -- & 64 / 61 \\
\bottomrule
\end{tabular}
\end{table}

\paragraph{False accepts and edit efficacy.} \Cref{tab:gen-delta}. Among accepted test edits the ground-truth object was altered in 0/2{,}174 (Waterbirds-2S), 1/2{,}094 (Waterbirds) and 0/2{,}654 (UrbanCars) edits: $\delta_{\rm nat}\approx0$ once the object is protected. Among \emph{rejected} edits it is 1.4--4.9\%, so the validator does filter the few damaging edits. Among the training edits of the fix, the object was altered in 0.12--0.28\% of edited images. The problem is efficacy. By the ground-truth detectors, only 13.9--42.6\% of accepted REMOVE/ADD test edits change the true background, which is a lower bound because not every hypothesis is a background flip (e.g.\ ``add a sailboat''). The editor-matched resample nulls also change the detected background in up to 23\% of edits.

\begin{table}[h!]
\centering
\caption{Generative test edits (all REMOVE/ADD edits of round 1). $\delta_{\rm nat}$: ground-truth object altered among accepted edits (Clopper--Pearson 95\% upper bound, UB). bg changed: ground-truth background attribute changed among accepted REMOVE / ADD edits, and among accepted resample-null edits (REMOVE / ADD side). UrbanCars co-object: changed by 73.3 / 69.3\% of REMOVE / ADD edits and 67.5 / 68.0\% of resample nulls.}
\label{tab:gen-delta}
\small
\setlength{\tabcolsep}{4pt}
\begin{tabular}{lcccccc}
\toprule
dataset & edits & accepted & $\delta_{\rm nat}$ (UB) & altered among rejected & bg changed R / A & resample null R / A \\
\midrule
Waterbirds-2S & 3200 & 67.9 & 0.0 (0.1) & 1.4 & 14.5 / 13.9 & 1.6 / 16.7 \\
Waterbirds & 3200 & 65.4 & 0.0 (0.2) & 4.9 & 17.0 / 25.4 & 6.0 / 15.2 \\
UrbanCars & 3200 & 82.9 & 0.0 (0.1) & 1.5 & 42.6 / 27.7 & 21.3 / 23.3 \\
CelebA & 2400 & 18.7 & -- & -- & -- & -- \\
Spawrious O2O & 6200 & 76.0 & -- & -- & -- & -- \\
Spawrious M2M & 5400 & 62.4 & -- & -- & -- & -- \\
\bottomrule
\end{tabular}
\end{table}

\paragraph{Fix with generated edits.} \Cref{tab:gen-fix}. Where the confirmed factors are backgrounds that the editor can repaint, the generative fix improves a lot on ERM: Waterbirds-2S $+13$ WGA, Waterbirds $+6$, UrbanCars $+45$, Spawrious O2O-hard $+10$ accuracy ($+18$ WGA) and M2M-hard $+14$ accuracy ($+15$ WGA), with IID accuracy within about 1 point on the bird and car datasets. On Spawrious, where no edit library exists, its average accuracy (the official metric) reaches or exceeds the best group-label baselines (O2O $89.6\pm0.9$ vs.\ GroupDRO$^\dagger$ $89.3$, within noise; M2M $75.3\pm0.4$ vs.\ DFR$^\dagger$ $72.2$). On worst-group accuracy it is best on M2M ($49.1$ vs.\ DFR$^\dagger$ $44.1$) but below GroupDRO$^\dagger$ on O2O ($73.5$ vs.\ $80.1$). On M2M this holds although no confirmed hypothesis hits the location keywords. On the mask-based datasets it stays 10--30 WGA points below the same discovery with the exact editor and below hand-list \peel: $79.9$ vs.\ $98.0$ (Waterbirds-2S), $80.4$ vs.\ $93.5$ (Waterbirds), $65.1$ vs.\ $94.1$ (UrbanCars). The gap is edit efficacy, not edit damage: the watermark is never edited (no watermark hypothesis, and no editor can toggle it), background flips succeed in only 14--43\% of accepted test edits, and UrbanCars keeps a BG gap of $-12.1$ while its CoObj gap closes ($-2.8$; background inpainting removes or replaces the co-object). CelebA is a negative result: the kept factors are ``beard'' (a gender proxy) and a sports-stadium background, validator acceptance is low (20\% and 52\% of training images), and the head-only fix \emph{lowers} WGA to $35.0$, below ERM ($44.8$) and below head-only retraining on the same subset without edits ($43.5$). A likely cause, not tested separately, is that beard edits act on the face region that also carries the hair cue. The KL ablation (seed 0 only, indicative) changes WGA by $+1.9$ (Waterbirds-2S), $-0.8$ (Waterbirds), $+9.3$ (UrbanCars), $+3.3$ (CelebA), $+0.1$ (O2O) and $+3.4$ (M2M); with $\delta_{\rm nat}\approx0$ the consistency term does not hurt here. Coverage of the training set: 84--96\% of training images receive at least one accepted edit on the bird datasets, UrbanCars and Spawrious, and 64\% on CelebA.

\paragraph{End to end over three discovery seeds.} To measure the variance of the whole pipeline, we re-ran it after the main runs for discovery seeds 1 and 2 on every dataset. Each run uses the ERM of its seed for discovery and the test, its own audit, an edited training set built from its own kept hypotheses, and one classifier trained with that seed. Round 2 was again not run. \Cref{tab:gen-e2e} compares these three end-to-end runs (seeds 0--2) with the seed-0 protocol. On Waterbirds-2S, Waterbirds and both Spawrious splits the end-to-end mean is within 3 points of the seed-0 protocol, with a spread of the same order. On UrbanCars, discovery variance dominates: worst-group accuracy ranges from $64.8$ (seed 0) to $76.8$. On CelebA, seeds 1 and 2 keep no hypothesis, because the audit drops every test-confirmed one. The loop then stops and returns ERM, and the end-to-end mean ($42.8$) stays below ERM ($44.8$). The kept hypotheses hit a true shortcut by keyword for every discovery seed on Waterbirds-2S (background only, never the watermark), Waterbirds, UrbanCars and O2O, for two of three on M2M, and for one of three on CelebA. On Spawrious the comparison with the group-label baselines is unchanged on M2M (worst-group $46.5$ vs.\ DFR$^\dagger$ $44.1$; average $75.2$ vs.\ $72.2$). On O2O, average accuracy ($89.1$) now ties GroupDRO$^\dagger$ ($89.3$) from slightly below, and worst-group accuracy stays below it ($73.2$ vs.\ $80.1$).

\begin{table}[h!]
\centering
\caption{Generative MD-\peel{} end to end. Seed-0 protocol: one discovery run, one edited training set, training seeds 0--2 (\cref{tab:gen-fix}). End to end: discovery seeds 0--2, each with its own test, audit, edited training set and classifier (mean$\pm$std and per seed). WGA on Spawrious is over class$\times$location groups; avg $=$ Spawrious average accuracy. $^\circ$: empty bank, so the loop returns ERM. Hits: discovery seeds whose kept hypotheses hit a true shortcut (strict keywords).}
\label{tab:gen-e2e}
\small
\setlength{\tabcolsep}{3.5pt}
\resizebox{\textwidth}{!}{%
\begin{tabular}{lcccccc}
\toprule
& \multicolumn{3}{c}{WGA} & \multicolumn{2}{c}{avg (Spawrious)} & \\
\cmidrule(lr){2-4}\cmidrule(lr){5-6}
dataset & seed-0 protocol & end to end & per discovery seed & seed-0 protocol & end to end & hits \\
\midrule
Waterbirds-2S & 79.9\apm{0.4} & 78.9\apm{1.6} & 80.3/76.8/79.7 & -- & -- & 3/3 \\
Waterbirds & 80.4\apm{0.9} & 80.0\apm{0.5} & 79.4/79.8/80.7 & -- & -- & 3/3 \\
UrbanCars & 65.1\apm{1.6} & 72.0\apm{5.2} & 64.8/74.4/76.8 & -- & -- & 3/3 \\
CelebA (head-only) & 35.0\apm{1.4} & 42.8\apm{5.7} & 35.0/45.0$^\circ$/48.3$^\circ$ & -- & -- & 1/3 \\
Spawrious O2O & 73.5\apm{3.8} & 73.2\apm{3.5} & 71.6/70.0/78.2 & 89.6\apm{0.9} & 89.1\apm{0.8} & 3/3 \\
Spawrious M2M & 49.1\apm{3.7} & 46.5\apm{1.6} & 48.8/45.1/45.6 & 75.3\apm{0.4} & 75.2\apm{1.5} & 2/3 \\
\bottomrule
\end{tabular}}
\end{table}

\paragraph{H3, generative part: not met.} The generative fix beats the best annotation-free baseline on Waterbirds-2S and Spawrious (average accuracy O2O $89.6$ vs.\ $79.6$, M2M $75.3$ vs.\ $61.7$; both also $\ge$ the best $^\dagger$ baseline on this metric, but on O2O not on worst-group accuracy). That is 2 of 5 datasets with Spawrious counted once (3 of 6 benchmarks if O2O and M2M are counted separately), against the required $\ge3/5$. It loses to ShortcutProbe on Waterbirds, UrbanCars and CelebA. The oracle-editor half of H3 is supported (\cref{app:e4-md}).

\begin{table}[h!]
\centering
\caption{Fix with the generative editor, seed-0 protocol (round 1 only; one discovery run; seeds 0--2 share one edited training set, so $\pm$ is retraining variance; end-to-end runs in \cref{tab:gen-e2e}; KL ablation: seed 0). Headline: WGA (Spawrious: average test accuracy). GPU-h: editing (once per dataset) + training (per seed).}
\label{tab:gen-fix}
\small
\setlength{\tabcolsep}{3.5pt}
\resizebox{\textwidth}{!}{%
\begin{tabular}{lccccccc}
\toprule
dataset & ERM & MD-\peel{} gen. & gen.\ + KL & MD-\peel{} exact & best annot.-free & best $^\dagger$ & GPU-h \\
\midrule
Waterbirds-2S & 66.5\apm{1.3} & 79.9\apm{0.4} & 81.8 & 98.0\apm{0.1} & 66.5 (ERM) & 89.6 (DFR) & 2.51 + 0.05 \\
Waterbirds & 74.1\apm{1.4} & 80.4\apm{0.9} & 79.6 & 93.5\apm{0.3} & 85.6 (SP) & 92.2 (DFR) & 3.70 + 0.18 \\
UrbanCars & 20.0\apm{1.3} & 65.1\apm{1.6} & 74.4 & 94.1\apm{1.0} & 88.5 (SP) & 86.9 (DFR) & 2.32 + 0.30 \\
CelebA (head-only) & 44.8\apm{3.0} & 35.0\apm{1.4} & 38.3 & n/a & 82.4 (SP) & 89.9 (DFR) & 6.54 + 0.09 \\
Spawrious O2O & 79.6\apm{0.6} & \textbf{89.6}\apm{0.9} & 89.7 & n/a & 79.6 (ERM/SP) & 89.3 (GroupDRO) & 14.05 + 0.17 \\
Spawrious M2M & 61.3\apm{0.5} & \textbf{75.3}\apm{0.4} & 78.7 & n/a & 61.7 (EIIL) & 72.2 (DFR) & 7.77 + 0.16 \\
\bottomrule
\multicolumn{8}{l}{\scriptsize SP: ShortcutProbe. Waterbirds-2S per group (gen.): $\varnothing$ 99.3, $\{$bg$\}$ 94.0, $\{$wm$\}$ 96.2. UrbanCars gaps (gen.): BG $-12.1$, CoObj $-2.8$, BG+CoObj $-22.6$.} \\
\multicolumn{8}{l}{\scriptsize CelebA control (head-only, same 20k subset, no edits): 43.5\apm{2.3}. Spawrious WGA (class$\times$location), gen.: O2O 73.5\apm{3.8} (ERM 55.1), M2M 49.1\apm{3.7} (ERM 34.1).}
\end{tabular}}
\end{table}

\paragraph{Cost.} Editing the training set is the dominant cost: 2.2 (UrbanCars), 2.5 (Waterbirds-2S), 3.6 (Waterbirds), 6.7 (CelebA), 7.7 (Spawrious M2M) and 14.1 (Spawrious O2O) GPU-h per dataset, shared by all seeds, plus 0.05--0.3 GPU-h of training per run and about 1--2 GPU-h per dataset for the test step. This is 10--50$\times$ the cost of \peel; the generative part used about 55 GPU-h in total (estimated from job logs).
\FloatBarrier

\subsection{CSI: the learned-adversary baseline}\label{app:e4-csi}

CSI (``certified shortcut immunization'') is an adversarial alternative to \peel{} that replaces the edit library by a learned adversary, the human audit by a CLIP validator, and the significance test on edit types by a bank of adversaries with a novelty reward. We implemented it as specified and ran it from the same ERM checkpoints as \peel.

\subsubsection{Implementation}
\begin{itemize}[leftmargin=*,itemsep=1pt,topsep=2pt]
\item \emph{Adversary $A_\phi(x,z,\tilde x)$}, $z\sim\mathcal N(0,I_{16})$, $\tilde x$ a donor image (a permutation of the batch). \emph{Compositional family} (default): $x'=(1-m)\odot x+m\odot c$ with a soft mask $m=\sigma(M_\phi(x,z))$ predicted at $\le1/4$ resolution, and content $c=g\cdot\mathrm{clip}(\tilde x+C_\phi(x,\tilde x,z))+(1-g)\cdot\mathrm{colorT}(x)$ (a residual on the donor, or a learned per-image $3\times3$ color transform of $x$). \emph{Residual family} (ablation): $x'=\mathrm{clip}(x+0.25\tanh(U_\phi(x,z)))$. Architecture: a small U-Net (base width 32, 3 levels, feature-wise linear modulation (FiLM) on $z$), initialized near the identity (mask logit bias $-4$).
\item \emph{Validator.} Frozen OpenAI CLIP ViT-B/16 zero-shot probability of the true class with a prompt ensemble; an edit is accepted if $P(y\mid x')\ge\tau=0.5$ and CLIP classifies the original $x$ correctly (otherwise the example is not edited). Prompts: 18 CIFAR templates (E2); a ``species'' prompt set chosen by worst-class IID validation accuracy (E3, Waterbirds); hair-color phrases (CelebA); breed names (Spawrious); car body types (UrbanCars).
\item \emph{Realism.} $d_{\rm real}(x')=$ mean cosine distance of the CLIP embedding of $x'$ to its $k=10$ nearest neighbors among CLIP embeddings of real training images (own source excluded); $\rho=$ the 95th percentile of the same statistic on held-out real images. Plus an edit-size penalty $\eta\cdot\mathrm{mean}(m)$.
\item \emph{Adversary objective}: $\mathrm{KL}(p_\theta(x)\|p_\theta(x'))+\alpha\,\mathrm{novelty}+\beta\,\mathrm{coverage}-\gamma\,\mathrm{drift}-\eta\,\mathrm{size}$, with novelty $=-\max_{b\in B}\mathrm{mean}_i\cos(\Delta r_{\phi,i},\Delta r_{b,i})$ on representation changes, coverage $=$ the $z$-diversity of $r_\theta(A(x,z))$, drift $=\mathrm{relu}(\tau-V)+\mathrm{relu}(d_{\rm real}-\rho)$; $\alpha=1,\beta=0.1,\gamma=5,\eta=0.1$, Adam $10^{-4}$.
\item \emph{Bank.} In round $r\le R_{\max}=6$ a fresh adversary is trained against the frozen classifier ($S_{\rm adv}=2000$ steps on E2, 1000 at 224\,px). It enters the bank $B$ if the flip rate of its accepted held-out edits exceeds that of a null adversary (identity plus mild color jitter) under a one-sided two-proportion $z$-test ($\alpha=0.05$); otherwise the loop stops. Then the classifier is immunized for $E_r$ epochs while the adversary keeps training (1:1 alternation).
\item \emph{Classifier loss} on validator-accepted edits: $L_{\rm CSI}=\mathrm{CE}(f(x),y)+\lambda_1\max_{T\in B}\mathrm{CE}(f(Tx),y)+\lambda_2\,\mathrm{mean}_T\|n(r(x))-n(r(Tx))\|^2+\lambda_3\,\mathrm{mean}_T\mathrm{CE}(f(Tx),y)+\lambda_4\,\mathrm{mean}_{T\in B}\mathrm{KL}(f(x)\|f(Tx))$ with $\lambda=(1,1,0.5,1)$; final model $=$ last round.
\item \emph{Necessary deviation.} The plain KL gives the adversary no gradient against the saturated ERM classifiers (median top-1 vs.\ top-2 logit gap $\approx10$ on E2; after 200 steps KL $=0.0000$ and flip rate $0.4\%$). KL is therefore computed on temperature-scaled logits, with $T=$ the median logit gap of the current $f$ on the held-out set (label-free). Other deviations: $E_r=20$ (E2) / 10 (E3, Waterbirds, UrbanCars) epochs per round with warm start; micro-batching and activation checkpointing on 224\,px data.
\item \emph{CSI-repaired} (three a-priori, label-free fixes of the failure modes below): (a) transplant/re-style only (content scale 0, no synthesized pixels); (b) effect-gated novelty, reward $=\mathrm{KL}\cdot(1+\alpha\,\mathrm{novelty})$, so a negligible edit cannot collect novelty; (c) a stricter validator with an extra abstain class (8 generic prompts such as ``an abstract pattern'', ``random noise'') and $\tau=0.8$ for both eligibility of $x$ and acceptance of $x'$.
\item \emph{Oracle validator}: acceptance by the ground-truth core audit instead of CLIP (CLIP remains the differentiable drift surrogate). Ground-truth audit: E2, an edited interior whose affine-color regression on the donor core beats that on the own core, or $R^2_{\rm own}<0.25$; E3, more than $25\%$ of bird-mask pixels changed by $>0.15$, or the donor content covers $>10\%$ of the donor bird. $\delta_{\rm nat}$ is the fraction of a bank member's accepted held-out edits flagged by this audit.
\end{itemize}
On the benchmarks, the adversary learning rate was chosen per dataset on seed 0 from $\{10^{-4},10^{-3}\}$ by worst-\emph{class} validation accuracy (class labels only): $10^{-3}$ on Waterbirds and Spawrious, $10^{-4}$ on CelebA and UrbanCars.

\subsubsection{Three tiers on E2 and E3}
\Cref{tab:csi-e2,tab:csi-e3} compare CSI as written, CSI-repaired and CSI with an oracle validator against \peel{} from the same ERM checkpoints. No tier improves on ERM beyond seed noise. On E3, the change in WGA paired with each run's own ERM start is $+1.2\pm3.4$ (as written), $+1.1\pm2.0$ (repaired) and $+1.9\pm2.6$ (oracle validator), all within noise; no ``same epochs without edits'' control was run.

\begin{table}[h!]
\centering
\caption{E2 (four shortcuts on CIFAR-10; ResNet-18). WGA over 16 flip groups; core-follow $=$ accuracy w.r.t.\ the core label on the core-swap test. Bank: mean members per run (clean $=$ $\delta_{\rm nat}\le0.15$, total over runs). $\delta_{\rm nat}$ pooled over members.}
\label{tab:csi-e2}
\small
\setlength{\tabcolsep}{4pt}
\resizebox{\textwidth}{!}{%
\begin{tabular}{lcccccccc}
\toprule
method & $n$ & IID & WGA & mean flip-group & core-follow & GPU-h & bank (clean) & $\delta_{\rm nat}$ \\
\midrule
ERM (start point) & 5 & 99.2\apm{0.4} & 0.0\apm{0.0} & 51.1 & 0.0 & 0.03 & -- & -- \\
\peel{} (exact library, audit) & 5 & 73.4\apm{2.7} & \textbf{71.3}\apm{4.0} & 72.9 & 71.6 & 0.05 & 4.0 (4) & 0 \\
Oracle coverage, all four & 5 & 73.0\apm{2.5} & 70.9\apm{4.0} & 72.4 & 71.2 & 0.05 & -- & -- \\
CSI as written & 5 & 98.3\apm{1.6} & 0.0\apm{0.0} & 51.1 & 0.0 & 0.81 & 1.6 (3 of 8) & 56.6\% \\
CSI-repaired & 5 & 99.4\apm{0.2} & 0.0\apm{0.0} & 51.3 & 0.0 & 0.29 & 0.2 (1 of 1) & 0.9\% \\
CSI, oracle validator (repaired adversary) & 5 & 99.3\apm{0.2} & 0.0\apm{0.0} & 51.4 & 0.0 & 0.35 & 0.4 (2 of 2) & 0.0\% \\
CSI, oracle validator, spec adversary & 3 & 99.1\apm{0.6} & 0.0\apm{0.0} & 51.0 & 0.0 & 0.23 & 0 & -- \\
CSI, oracle validator, transplant-only & 3 & 81.0\apm{31.1} & 14.7\apm{25.4} & 50.9 & 18.7 & 0.93 & 2.0 (6 of 6) & 0.0\% \\
\midrule
DFR$^\dagger$ / GroupDRO$^\dagger$ / JTT$^\dagger$ & 5 & 63.6/60.1/98.6 & 43.6/34.4/0.2 & & 49.5/40.4/0.2 & & & \\
\bottomrule
\end{tabular}}
\end{table}

\begin{table}[h!]
\centering
\caption{E3 (Waterbirds-2S; ResNet-50). Columns as in \cref{tab:csi-e2}; $\{$bg,wm$\}=$ WGA.}
\label{tab:csi-e3}
\small
\setlength{\tabcolsep}{3.5pt}
\resizebox{\textwidth}{!}{%
\begin{tabular}{lccccccccc}
\toprule
method & $n$ & IID & $\{$bg$\}$ & $\{$wm$\}$ & WGA & core-follow & GPU-h & bank (clean) & $\delta_{\rm nat}$ \\
\midrule
ERM (start point) & 5 & 98.9\apm{0.1} & 91.9 & 97.3 & 66.5\apm{1.3} & 58.8 & 0.02 & -- & -- \\
\peel{} & 5 & 98.5\apm{0.1} & 98.6 & 98.4 & \textbf{98.2}\apm{0.1} & 97.0 & 0.04 & 2.0 (2) & 0 \\
CSI as written & 5 & 99.0\apm{0.1} & 92.6 & 97.6 & 67.7\apm{4.0} & 62.3 & 0.62 & 1.0 (1 of 5) & 47.1\% \\
CSI-repaired & 5 & 98.9\apm{0.1} & 92.1 & 97.5 & 67.6\apm{2.7} & 61.5 & 0.37 & 0.4 (2 of 2) & 3.6\% \\
CSI, oracle validator (repaired adversary) & 5 & 98.9\apm{0.1} & 92.4 & 97.4 & 68.4\apm{3.5} & 62.2 & 0.37 & 0.4 (2 of 2) & 0.0\% \\
CSI, oracle validator, spec adversary & 3 & 99.0\apm{0.2} & 92.7 & 97.5 & 69.1\apm{4.0} & 63.9 & 0.48 & 0.7 (2) & 0.0\% \\
CSI, oracle validator, transplant-only & 3 & 99.0\apm{0.1} & 93.3 & 97.2 & 66.9\apm{4.6} & 63.9 & 0.61 & 1.0 (3) & 0.0\% \\
Random library edits (E3) & 5 & 99.3 & 97.1 & 97.1 & 76.3\apm{1.2} & 45.8 & 0.04 & -- & -- \\
\midrule
JTT$^\dagger$ / DFR$^\dagger$ / GroupDRO$^\dagger$ & 5 & 96.9/96.6/98.2 & & & 83.1/89.6/85.6 & 82.8/92.9/86.0 & & & \\
\bottomrule
\end{tabular}}
\end{table}

\subsubsection{Failure modes}
\paragraph{Validator gaming.} On E2 the round-1 adversary does not find a shortcut; it erases the image. For all three adversary learning rates tried ($10^{-4}$, $3\cdot10^{-4}$, $10^{-3}$) the accepted edits replace 70--99\% of the image by a nearly input-independent pattern (saturated stripes and a shared green texture; \cref{fig:csi-garbage}). The cat or dog is gone (correlation of the edited interior with the own core $0.01$--$0.21$, with the donor $0.01$--$0.03$). Yet the binary CLIP validator gives $P(y\mid x')\approx0.53$--$0.60\ge\tau$ to 74--79\% of them, and the realism distance ($0.066$--$0.068$) sits just below $\rho=0.0696$: both soft constraints are saturated at their boundary. Over all bank-test candidates (\cref{tab:csi-val}), CLIP accepts \textbf{79.4\%} (E2) and \textbf{92.2\%} (E3) of the edits that the ground-truth audit flags as core-altering. In \peel's terms, $\delta\approx0.8$--$0.9$, far above the $0.15$ the audit enforces. Because the pattern overwrites patch, border, tint and grating, a ``which shortcuts changed'' audit would credit such a member with all four shortcuts. After immunization on these edits, WGA is still $0.0$. The CLIP validator also relies on the shortcuts itself: its zero-shot WGA is $73.6$ on Waterbirds-2S, $71.0$ on Waterbirds, $80.6$ on CelebA, $70.4$ on UrbanCars, and the worst group is $78.3/78.5$ on Spawrious O2O/M2M.

\begin{table}[h!]
\centering
\caption{The CLIP validator against the ground-truth audit, pooled over all bank-test candidates.}
\label{tab:csi-val}
\small
\begin{tabular}{lccc}
\toprule
validator & $P(\text{accept}\mid\text{core altered})$ & $P(\text{reject}\mid\text{intact})$ & $\delta_{\rm nat}$ of bank members \\
\midrule
E2, CLIP as written ($\tau=0.5$) & \textbf{79.4\%} & 14.8\% & 56.6\% \\
E3, CLIP as written & \textbf{92.2\%} & 3.5\% & 47.1\% \\
E2, CLIP repaired (abstain, $\tau=0.8$) & 55.6\% (few altered edits) & 36.3\% & 0.9\% \\
E3, CLIP repaired & 86.7\% (few altered edits) & 8.7\% & 3.6\% \\
\bottomrule
\end{tabular}
\end{table}

\paragraph{Novelty harvesting.} After the first bank member, the scale-invariant novelty reward is collected by negligible edits, so the next bank test fails and the loop stops with 1--3 members. With $\alpha=0$ the loop continues further (round-2/3 candidates flip 1.8--48\%, 4.0 members on average), confirming that novelty, not exhaustion, stopped it; WGA stays $0.0$. The effect-gated novelty of CSI-repaired removes this route, but the surviving members are weak.

\paragraph{Empty banks with a perfect validator.} With the oracle validator and the specified adversary on E2, at most 1 of 1{,}000 held-out edits is accepted per seed and the bank stays empty. On the benchmarks (seed 0), CSI with the oracle validator finds no significant intervention on Waterbirds or UrbanCars (bank empty, WGA $73.7$ and $18.4$, i.e.\ ERM), although an exact background swap flips $13\%$ (Waterbirds) and $23\%$ (UrbanCars) of ERM predictions. With the transplant-only or repaired adversary, the oracle-validator bank is small (0.4--2.0 members per run) and its clean members make tiny edits (E3: area $0.03$--$0.06$, flip rate $0.6$--$1.0\%$). No member ever performed the natural interventions in \peel's library (swap the whole background, toggle the watermark, re-render a whole border, tint or grating). On E3 the size penalty makes a background swap ($\approx75\%$ of the pixels) cost about as much as it gains; a post-hoc run with $\eta=0$ (3 seeds) still found no background swap (WGA $68.0\pm2.7$). Open-ended discovery, not only the validator, is the obstacle.

\begin{figure}[h!]
\centering
\includegraphics[width=0.85\textwidth]{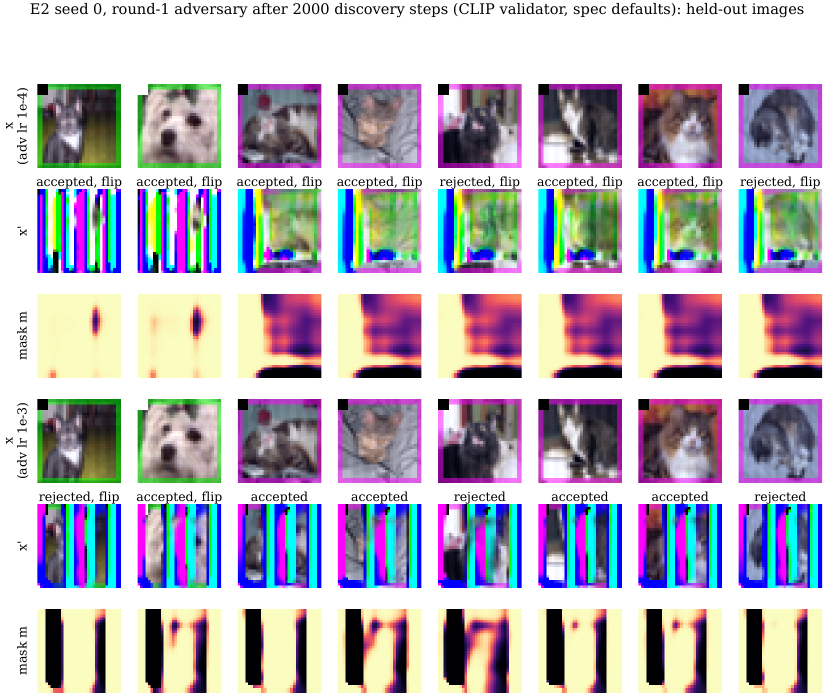}
\caption{CSI on E2, seed 0, round-1 adversary after 2000 discovery steps (CLIP validator, specified defaults), on held-out images, for adversary learning rates $10^{-4}$ (top) and $10^{-3}$ (bottom): original $x$, edit $x'$ with the validator decision and whether the prediction flipped, and the mask $m$. The accepted edits overwrite most of the image with a near input-independent pattern.}
\label{fig:csi-garbage}
\end{figure}

\subsubsection{Ablations}
\Cref{tab:csi-abl} lists the ablations on E2 (3 seeds unless noted). None moves WGA off $0.0$. The residual $L_\infty$ adversary is the only one that finds clean shortcut edits: the tint $S_2$ (27--87\% of its flips) and grating $S_4$ (18--91\%), the two low-amplitude global shortcuts that fit inside $\varepsilon=0.25$ (the tint salience $\hat\lambda_{S_2}$ drops from 2.9--5.8 to 0.3--0.6). It still stops after one member. The coverage curve (bank capped at $R_{\max}\in\{0,1,2,3,6\}$) is flat: WGA $0.0$ at every cap on E2 as written, and $66.5/67.7/67.7/67.7/67.7$ on E3. Oracle corruption (oracle validator with transplant-only adversary; rejected edits accepted with probability $\varepsilon$) reaches the bank ($\delta_{\rm nat}=0.0/2.3/4.7/11.3/18.9\%$ for $\varepsilon=0/0.05/0.1/0.2/0.3$), but WGA is already $\approx0$ at $\varepsilon=0$ ($14.7\pm25.4$ from one diverged seed, then $0.1/0.0/0.0/0.0$), so there is no robustness to degrade.

\begin{table}[h!]
\centering
\caption{CSI ablations on E2 (3 seeds unless noted) and E3 where run. $L_{\rm CF\text{-}inv}$: the counterfactual (CF) representation-invariance term of $L_{\rm CSI}$ (weight $\lambda_2$); $L_{\rm suff}$: its sufficiency term, the cross-entropy on edits (weight $\lambda_3$).}
\label{tab:csi-abl}
\small
\begin{tabular}{lcccccc}
\toprule
variant & E2 IID & E2 WGA & E2 bank (clean) & E2 $\delta_{\rm nat}$ & E2 GPU-h & E3 WGA ($n$) \\
\midrule
as written (5 seeds) & 98.3\apm{1.6} & 0.0 & 1.6 (3/8) & 56.6\% & 0.81 & 67.7\apm{4.0} (5) \\
single adversary, no bank & 99.5\apm{0.1} & 0.0 & 1.0 (0/3) & 84.7\% & 0.68 & -- \\
no novelty ($\alpha=0$) & 98.9\apm{0.9} & 0.0 & 4.0 (3/12) & 65.7\% & 1.64 & -- \\
no $L_{\rm suff}$ ($\lambda_3=0$) & 99.2\apm{0.2} & 0.0 & 1.7 (0/5) & 89.0\% & 0.82 & -- \\
no $L_{\rm CF\text{-}inv}$ ($\lambda_2=0$) & 98.9\apm{1.2} & 0.0 & 2.0 (2/6) & 55.9\% & 0.98 & -- \\
residual $L_\infty$ adversary & 97.4\apm{2.1} & 0.0 & 1.0 (3/3) & 4.3\% & 0.55 & 67.8\apm{4.2} (3) \\
transplant-only & 97.7\apm{2.7} & 0.2\apm{0.3} & 2.3 (4/7) & 26.6\% & 1.13 & 67.1\apm{3.3} (3) \\
oracle validator (spec adversary) & 99.1\apm{0.6} & 0.0 & 0 & -- & 0.23 & 69.1\apm{4.0} (3) \\
\bottomrule
\end{tabular}
\end{table}

\subsubsection{CSI on the standard benchmarks}
CSI as written does not beat the strongest annotation-free baseline anywhere (\cref{tab:main,tab:csi-bench}). The bank never grows beyond one member. The round-1 adversary is significant on Waterbirds, CelebA and UrbanCars, but its accepted edits mostly alter the object ($\delta_{\rm nat}=0.40$--$1.00$ on Waterbirds; $1.00$ on UrbanCars, where the edit is a near-global color overlay that covers the car). Immunizing against these edits leaves the true shortcut untouched (exact-edit flip rates: Waterbirds background $13.1\to13.9\%$; UrbanCars background/co-object $22.9/18.0\to21.1/15.5\%$). On Spawrious the adversary never beats the null after 1{,}000 discovery steps ($\le0.1\%$ flips, 0--1 of $\approx950$ accepted edits). A 3$\times$ longer discovery on Waterbirds (seed 0) changed nothing (WGA $73.1$). CSI-repaired helps at most a little (Waterbirds $76.9\pm2.3$, UrbanCars $28.3\pm4.9$) and remains 9--60 points below \peel, ShortcutProbe and DFR$^\dagger$.

\begin{table}[h!]
\centering
\caption{CSI on the benchmarks (selected configuration, 3 seeds). Bank sizes and $\delta_{\rm nat}$ per seed ($\delta_{\rm nat}$ needs a ground-truth mask: Waterbirds and UrbanCars only). CLIP zero-shot: the validator's own WGA (Spawrious: worst group). Oracle validator: seed 0. GPU-h per seed on top of the ERM checkpoint.}
\label{tab:csi-bench}
\small
\begin{tabular}{lcccccc}
\toprule
dataset & adv.\ lr & bank & $\delta_{\rm nat}$ & CLIP zero-shot & oracle validator: bank / WGA & GPU-h \\
\midrule
Waterbirds & $10^{-3}$ & 1/1/1 & 0.40/0.93/1.00 & 71.0 & 0 / 73.7 & 0.61 \\
CelebA & $10^{-4}$ & 1/1/1 & -- & 80.6 & -- & 1.11--1.12 \\
UrbanCars & $10^{-4}$ & 1/1/1 & 1.00/1.00/1.00 & 70.4 & 0 / 18.4 & 1.10--1.11 \\
Spawrious O2O & $10^{-3}$ & 0/0/0 & -- & 78.3 & -- & 0.20--0.21 \\
Spawrious M2M & $10^{-3}$ & 0/0/0 & -- & 78.5 & -- & 0.20--0.21 \\
\bottomrule
\end{tabular}
\end{table}

\paragraph{Cost.} CSI as written costs $0.81\pm0.38$ GPU-h per run on E2 against $0.05$ for \peel{} (over-stated, since those \peel{} jobs shared a GPU), and $0.62$ on E3 against $0.04$, i.e.\ about $16\times$ and 8--16$\times$ \peel{} (other tiers in \cref{tab:csi-e2,tab:csi-e3}). The dominant cost is the CLIP ViT-B/16 forward and backward pass inside every adversary step (E2: 0.41\,s/step at batch 256; E3: 0.7\,s/step at batch 64). The runs are cheap only because they stop early; a run reaching $R_{\max}=6$ would cost $\approx2$\,h (E2) or $\approx3$\,h (E3). On the benchmarks CSI takes 0.6\,h (Waterbirds), 1.1\,h (CelebA, UrbanCars) and 0.2\,h (Spawrious) per seed, against 0.17\,h (Waterbirds) and 0.27\,h (UrbanCars) for \peel{} including its retraining from ImageNet. The 83 finished CSI runs on E2/E3 used 53.5 GPU-h.
\FloatBarrier

\subsection{Transfer: eight shortcuts, four covered}\label{app:e4-transfer}

\paragraph{Setup.} The core, splits, model and schedule are those of E2 (CIFAR-10 cat vs.\ dog, ResNet-18 from scratch, 40 epochs, last epoch). There are eight shortcuts, each with $\rho_j=0.95$ and independent given $y$. Families A--D are the four E2 shortcuts with E2's amplitudes (corner patch, tint $0.06$, border $\alpha=0.5$, grating $0.02$). Families E--H are four new, spatially local, exactly rendered shortcuts inside the border ring, disjoint from each other and from $S_1$ (together 74 pixels, $7.2\%$ of the image): a second $3\times3$ corner patch (yellow-ish vs.\ blue-ish), a $3\times3$ central dot (``+'' vs.\ ``$\times$''), a $1\times20$ stripe (cyan-ish vs.\ red-ish, contrast 0.5), and a $6\times6$ additive checkerboard texture (1-px vs.\ 2-px, amplitude 0.10). The E--H amplitudes were chosen by a screen of 6 settings on ERM validation saliences; with weaker settings some families are not used at all. In the three seeds checked (0--2), ERM uses $S_1,S_2,S_3,S_5,S_6,S_7$ in all of them, $S_8$ in seeds 1--2, and the grating $S_4$ clearly only in seed 2. Its core reliance is $0.0$ and its WGA (all eight flipped) is $0.0$. Evaluation uses all $2^8=256$ flip groups on the full 2{,}000-image test set. ``A--D-only'' and ``E--H-only'' are the 15 groups that flip a non-empty subset of one half with the other half aligned. Five seeds per arm.

\paragraph{Arms.} Oracle exact coverage of A--D ($q=1$, $\lambda_{\rm cons}=1$); \peel{} with the library restricted to A--D edits (plus nulls and core swap, simulated validator $\delta=0.1$, audit $m=50$), with its logit-margin test (mean $|\Delta$margin$|$ and Welch $t$-test) or, for comparison, with the flip-rate test of our first runs (same loop); \peel{} with the full A--H library (both tests); CSI with an unrestricted adversary, and CSI-R whose adversary is masked so that it can never change the E--H pixels (verified: E--H change $0.000$ on every accepted edit); the repaired variants of both; and references with oracle coverage of E--H only and of all eight.

\paragraph{Results.} \Cref{tab:transfer} and \cref{fig:transfer}. Covering only A--D \emph{lowers} accuracy on the groups where only the never-covered E--H families are flipped: oracle A--D coverage takes the E--H-only mean from $89.7$ to $47.3$ (paired $-42.4\pm1.6$), and 11 of 15 such groups drop significantly (one-sided paired $t$-test, 5 seeds) while none rises. The other 4 are single-flip groups, which the remaining aligned shortcuts out-vote. After coverage only $S_5$--$S_8$ carry salience ($\hat\lambda\approx15$--$17$ vs.\ $\approx0$ for $S_1$--$S_4$). This is \cref{cor:wam}(b), and it rules out transfer to uncovered families. No CSI arm gains on E--H: CSI-R changes the E--H-only mean by $-6.3\pm13.1$ and CSI-repaired-R by $+0.2\pm0.3$. Unrestricted CSI moves reliance between families rather than to the core (seed 1: $\{5678\}$ $8.0\to81.4$ but $\{1234\}$ $91.5\to23.8$), and its round-1 members are mostly donor transplants ($\delta_{\rm nat}$ 63--100\% of accepted edits, 32--52\% label-corrupting where logged).

\paragraph{The flip-rate test fails with redundant shortcuts.} With eight redundant $\rho=0.95$ shortcuts and saturated ERM logits, randomizing a single family almost never flips an ERM prediction (round-1 $\widehat\Gamma\le1.3\%$ for every family and seed). The flip-rate variant therefore keeps nothing in \textbf{3 of 5} seeds, with both the A--D library (seeds 1--3) and the full library (seeds 1--3), and the model stays ERM. With the full library, seed 0 finds all eight (WGA 70.6) and seed 4 finds only $S_1$. This is the early-stopping failure noted after \cref{prop:greedy}. \peel's margin test finds all eight families within two to three rounds in \emph{every} seed and reaches $71.2\pm6.2$ WGA, against $68.6\pm3.4$ for oracle coverage of all eight. With the A--D library it finds $S_1$--$S_3$ in every seed ($S_4$ in 2 seeds) and reproduces the oracle's Whac-a-Mole drop ($-41.8\pm0.8$). This is a weakness of the flip-rate statistic under redundancy, not of coverage, and it is why \peel{} and MD-\peel{} test on margins (\cref{prop:greedy}).

\begin{table}[h!]
\centering
\caption{Transfer on 8-shortcut CIFAR (5 seeds). WGA over all 256 groups. A--D-only / E--H-only: mean over the 15 groups flipping only that half. $\Delta$E--H: paired change of the E--H-only mean vs.\ ERM. Groups: number of the 15 E--H-only groups that decrease by $>0.5$ points (of which significant), and that increase.}
\label{tab:transfer}
\small
\setlength{\tabcolsep}{3.5pt}
\begin{tabular}{lcccccccc}
\toprule
arm & IID & WGA & A--D-only & E--H-only & $\{5678\}$ & $\Delta$E--H & down (sig.) / up \\
\midrule
ERM & 99.9\apm{0.1} & 0.0\apm{0.0} & 98.0\apm{1.1} & 89.7\apm{2.4} & 17.0\apm{12.4} & -- & -- \\
Oracle coverage A--D & 99.2\apm{0.1} & 0.0\apm{0.0} & 100.0\apm{0.0} & 47.3\apm{1.0} & 0.0\apm{0.0} & $-$42.4\apm{1.6} & 11 (11) / 0 \\
flip-rate variant, A--D library & 99.9\apm{0.2} & 0.0\apm{0.0} & 98.7\apm{1.6} & 80.5\apm{14.9} & 12.1\apm{15.9} & $-$9.1\apm{13.1} & 11 (0) / 0 \\
\peel{} (margin test), A--D library & 99.1\apm{0.1} & 0.0\apm{0.0} & 100.0\apm{0.0} & 47.9\apm{2.2} & 0.0\apm{0.0} & $-$41.8\apm{0.8} & 11 (11) / 0 \\
flip-rate variant, full library & 94.5\apm{12.2} & 14.1\apm{31.6} & 93.1\apm{11.8} & 83.7\apm{10.0} & 26.8\apm{29.8} & $-$6.0\apm{8.2} & 13 (0) / 2 \\
\peel{} (margin test), full library & 74.7\apm{4.2} & \textbf{71.2}\apm{6.2} & 73.7\apm{4.7} & 74.7\apm{4.2} & 74.7\apm{4.3} & $-$15.0\apm{3.7} & 13 (12) / 2 \\
CSI, unrestricted & 99.5\apm{0.5} & 0.0\apm{0.0} & 85.4\apm{14.6} & 92.1\apm{5.7} & 53.2\apm{36.1} & $+$2.4\apm{6.8} & 3 (0) / 4 \\
CSI-R (cannot touch E--H) & 99.3\apm{1.1} & 0.0\apm{0.0} & 95.8\apm{7.8} & 83.3\apm{11.2} & 20.1\apm{34.4} & $-$6.3\apm{13.1} & 10 (0) / 1 \\
CSI-repaired, unrestricted & 99.9\apm{0.0} & 0.0\apm{0.0} & 98.0\apm{1.2} & 89.1\apm{3.7} & 18.6\apm{11.0} & $-$0.5\apm{3.7} & 2 (0) / 4 \\
CSI-repaired-R & 99.9\apm{0.1} & 0.0\apm{0.0} & 98.4\apm{0.9} & 89.8\apm{2.5} & 14.7\apm{10.7} & $+$0.2\apm{0.3} & 2 (0) / 2 \\
\midrule
Oracle coverage E--H (ref.) & 98.7\apm{0.8} & 0.0\apm{0.0} & 48.6\apm{0.8} & 100.0\apm{0.0} & 100.0\apm{0.0} & $+$10.3\apm{2.4} & 0 (0) / 7 \\
Oracle coverage all 8 (ref.) & 72.6\apm{2.4} & 68.6\apm{3.4} & 71.7\apm{2.7} & 72.5\apm{1.7} & 72.5\apm{1.3} & $-$17.1\apm{1.2} & 13 (12) / 2 \\
\bottomrule
\end{tabular}
\end{table}

\begin{figure}[h!]
\centering
\includegraphics[width=0.8\textwidth]{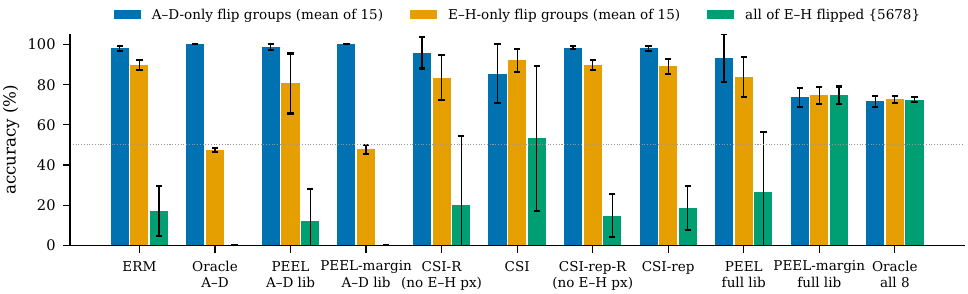}
\caption{Transfer on 8-shortcut CIFAR: A--D-only mean, E--H-only mean and $\{5678\}$ accuracy per arm (mean$\pm$std over 5 seeds).}
\label{fig:transfer}
\end{figure}
\FloatBarrier

\subsection{Scaling: six backbones on Waterbirds-2S}\label{app:e4-scale}

\paragraph{Setup.} The E3 data, training loop, \peel{} and evaluation are reused unchanged, with an architecture-aware model factory. Backbones (pretrained, new 2-way head): ResNet-18 and ResNet-50 (ImageNet-1k); ViT-S/16 and ViT-B/16 (ImageNet-21k then 1k); CLIP ViT-B/16 and ViT-L/14 image towers (OpenAI). ResNets use the E3 protocol (SGD $10^{-3}$, 20 epochs); all ViTs, including CLIP, use E3's CLIP protocol (AdamW $10^{-5}$, weight decay $0.1$, 10 epochs), untuned. \peel{} and CSI start from the ERM of the same backbone and seed (CSI: E3 settings, $S_{\rm adv}=1000$, 10 immunization epochs per round, adversary learning rate $10^{-4}$). GPU time is the measured training wall-clock including the ERM run each method starts from. Training floating-point operations (FLOPs) are $3\times$ forward FLOPs $\times$ images processed for ERM and \peel; for CSI they count classifier and CLIP-validator passes only (the adversary is excluded, so it is a lower bound). 3 seeds per cell.

\paragraph{Results.} \Cref{tab:scale} and \cref{fig:scale}. ERM robustness grows with scale and pretraining ($54.7\to85.4$ WGA), but no backbone removes the shortcuts. \peel{} reaches 97.1--99.4 at every scale, with bank $\{$bg, wm$\}$ in all 18 runs; its gain over ERM shrinks ($+42.5\to+13.9$) only because ERM improves. CSI gives no reliable gain at any scale (paired $\Delta$WGA vs.\ ERM $-0.8$ to $+5.8$; repaired $0.0$ to $+2.6$), and its gain does not grow with scale. On the two CLIP backbones the adversary fails the null test in 5 of 6 runs (repaired: 6 of 6), so the model is ERM. Where a member is found on a ResNet it is a partial background/bird transplant ($\delta_{\rm nat}$ 40--94\%, 11--27\% label-corrupting); the watermark is never targeted. CSI is the most expensive method at every scale.

\begin{table}[h!]
\centering
\caption{Waterbirds-2S backbone sweep (3 seeds; WGA $=\{$bg,wm$\}$ group). GPU-min and training PFLOPs for ERM / \peel{} / CSI.}
\label{tab:scale}
\small
\setlength{\tabcolsep}{3.5pt}
\begin{tabular}{lcccccccc}
\toprule
backbone & GFLOPs/img & ERM & \peel & CSI & CSI-repaired & GPU-min & PFLOPs \\
\midrule
ResNet-18 & 3.6 & 54.7\apm{2.3} & 97.1\apm{0.1} & 53.9\apm{4.0} & 56.0\apm{3.0} & 0.4 / 1.3 / 32.9 & 1.04 / 3.13 / 25.67 \\
ResNet-50 & 8.2 & 65.9\apm{1.4} & 98.3\apm{0.1} & 67.5\apm{4.9} & 68.3\apm{3.1} & 1.0 / 3.4 / 37.5 & 2.35 / 7.06 / 34.26 \\
ViT-S/16 & 9.2 & 69.0\apm{9.6} & 98.5\apm{0.3} & 74.8\apm{4.4} & 69.1\apm{9.7} & 0.4 / 1.5 / 34.6 & 1.32 / 3.97 / 34.87 \\
ViT-B/16 & 35.1 & 81.9\apm{5.5} & 99.4\apm{0.1} & 82.4\apm{6.3} & 84.4\apm{2.9} & 1.0 / 3.4 / 22.8 & 5.06 / 15.17 / 45.26 \\
CLIP ViT-B/16 & 35.1 & 81.4\apm{1.0} & 98.8\apm{0.1} & 80.6\apm{1.7} & 81.4\apm{1.0} & 1.0 / 3.4 / 22.8 & 5.06 / 15.17 / 45.26 \\
CLIP ViT-L/14 & 162.0 & 85.4\apm{2.8} & 99.3\apm{0.1} & 85.4\apm{2.8} & 85.4\apm{2.8} & 4.5 / 15.5 / 26.7 & 23.33 / 69.99 / 103.59 \\
\bottomrule
\end{tabular}
\end{table}

\paragraph{CLIP zero-shot reference.} Without training, CLIP is itself shortcut-reliant on Waterbirds-2S: WGA $73.6$ (ViT-B/16) and $83.5$ (ViT-L/14) with the ``species'' prompts, $46.3/74.3$ with plain class prompts, and $32.8/34.5$ with habitat-worded prompts. CLIP ViT-B/16 zero-shot is CSI's validator, so the validator shares the background bias (its $\{$bg$\}$-group accuracy is $78.5$). Fine-tuned CLIP towers: ERM $81.4/85.4$, \peel{} $98.8/99.3$, CSI $80.6/85.4$. Fine-tuning of large vision--language models (e.g.\ on SpuriVerse) was out of scope.

\begin{figure}[h!]
\centering
\includegraphics[width=0.9\textwidth]{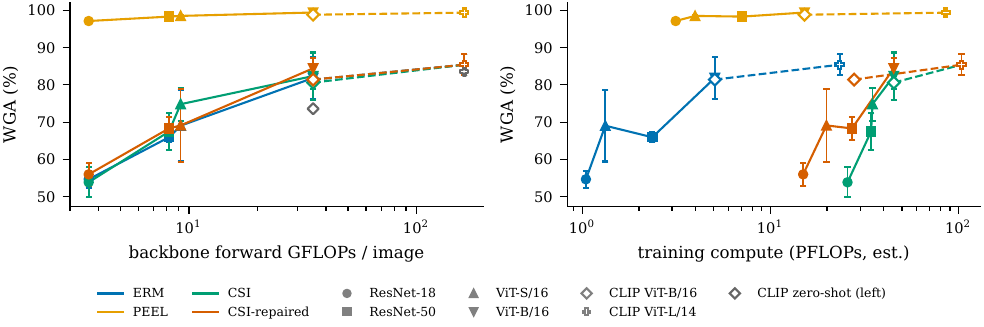}
\caption{Waterbirds-2S WGA vs.\ backbone forward GFLOPs per image (left) and vs.\ estimated training compute (right) for ERM, \peel{} and CSI. Solid: supervised ImageNet backbones; dashed: CLIP towers; grey (left): CLIP zero-shot. ViT-B/16 and CLIP ViT-B/16 have identical FLOPs.}
\label{fig:scale}
\end{figure}
\FloatBarrier

\subsection{Text: MNLI $\to$ HANS and the Shortcut Suite}\label{app:e4-text}

\paragraph{Setup.} Training uses MNLI train minus a stratified 3{,}000-pair held-out set (389{,}702 pairs); the held-out set is used for bank tests, type tests and trouble sets and is never trained on. Evaluation: the MNLI matched and mismatched validation sets (MNLI-m, MNLI-mm); HANS (30{,}000 pairs; 3 heuristics $\times$ \{entailment, non-entailment\}); the Shortcut Suite (SS; 7 sets of 3{,}000: an MNLI-m subset, negation, position, Bible style, lexical overlap, subsequence, constituent; ``SS worst'' $=$ minimum over the six shortcut sets); and the counterfactual (CF) flip rate on validator-accepted rewrites of 3{,}000 MNLI-m pairs, macro-averaged over instructions. Classifier: BERT-base-uncased, AdamW $2\cdot10^{-5}$, 3 epochs, last checkpoint, seeds 0--2. A 3-class prediction is mapped to entailment/non-entailment for HANS-derived sets.

\paragraph{Editor, validator and instructions.} Qwen2.5-7B-Instruct (vLLM) is told the gold relation and asked for a label-preserving rewrite under one of 12 instructions: increase/decrease lexical overlap, add/remove negation, make the hypothesis verbose/short, change names, change demographic terms, add a distractor, reorder, change style, change formatting; plus a null edit (light premise paraphrase). Two samples are drawn per pair and instruction at temperature 0.9 (1.30M rewrites for a 50{,}000-pair edit pool). The validator is a frozen DeBERTa-v3-large NLI model: an edit is accepted if $P_V(\text{gold}\mid x')\ge0.5$, $V$ is correct on the original, the text changed, and a GPT-2 realism check passes (mean token negative log-likelihood [NLL] $\le$ the 95th percentile on real MNLI text). A second judge (RoBERTa-large NLI) and a zero-shot LLM judge are used for audits only. The instructions do change what they claim (overlap$\uparrow$: $+0.32$ overlap; negation$\uparrow$: $+0.88$ negation words; verbose: $+18$ hypothesis words), but the rewriter almost never produces HANS-style full-overlap non-entailment pairs ($\le2.1\%$ of accepted rewrites). The riskiest types by excess second-judge disagreement are ``verbose hypothesis'' ($+6.6$ points) and ``add negation'' ($+6.2$); all others are $\le\approx2$.

\paragraph{Methods.} \emph{CF-aug}: \peel's Learn step on all 12 instructions uniformly, with no test or audit. \emph{\peel-style}: per round, a margin test of each instruction type against the null (\cref{prop:greedy}; mean absolute change of the logit margin of the predicted class, one-sided Welch test, Holm; our first runs used the flip rate), an audit of 50 accepted edits per significant type (bad if both judges reject; Clopper--Pearson $\bar\delta\le0.15$), and Learn with CE $+$ CE on edits $+$ KL. \emph{CSI for text}: the adversary is a learned policy over the 12 instructions (REINFORCE), with the reward of \cref{app:e4-csi} on accepted rewrites and a bank of up to 6 members. \emph{MD-\peel}: below. References: product of experts with a known-bias model on hand-crafted HANS features (PoE known-bias$^\ddagger$; uses prior knowledge of the target shortcut), PoE with a weak-learner bias model, and JTT with a fixed $\lambda_{\rm up}=6$. Default budgets differ (CF-aug 4k steps, \peel-style 4--8k, CSI 10--12k), so compute-matched 12k-step controls are included.

\paragraph{Results.} \Cref{tab:text-bert}. Every edit-based method lowers the CF flip rate at unchanged IID accuracy. On HANS, at matched compute (12k steps), uniform CF-aug is as good as or better than every selection method: $64.7\pm1.3$ (augmentation-only objective) and $63.3\pm0.4$ (CE+KL objective), against $62.8$ (MD-\peel), $62.4$ (\peel-style) and $62.3$ (CSI), and ERM $59.0$. CSI $-$ CF-aug-12k (CE+KL) is $-0.94\pm0.63$ on HANS ($p=.09$), and CSI $-$ \peel-style-12k is $-0.04\pm1.73$ (not significant). A single CSI adversary without bank or novelty, given the same 12k steps, reaches the same HANS ($62.3\pm2.4$). CSI's bank recovers the classic artifact families (round 1 in every seed: verbose hypothesis; later: style, decrease overlap, add negation). The first member \emph{lowers} HANS ($59.0\to56.6$), a Whac-a-Mole step, and HANS recovers only as later members are added ($\to62.3$ with 6). In every seed, CSI's adversary picks first the instruction type with the highest second-judge false-accept excess (verbose hypothesis) and later returns to the second highest (add negation): maximizing the prediction change under an imperfect validator selects label-changing rewrites. The known-bias PoE$^\ddagger$ is far ahead on HANS ($69.8$; worst cell $38.7$ vs.\ $8.3$) at $-1.3$ points IID. The same picture holds for DeBERTa-v3-base and for a Qwen2.5-0.5B classifier (\cref{tab:text-other}): CSI is not distinguishable from \peel-style or compute-matched CF-aug. On Qwen2.5-0.5B, CSI raises SS worst by $+4.2\pm1.1$ over ERM (constituent set), as do \peel-style and CF-aug.

\begin{table}[h!]
\centering
\caption{BERT-base on MNLI (3 seeds). HANS non-ent: the non-entailment half (the worst-group analogue); worst cell: minimum over the 6 HANS cells. CF flip: lower is better. GPU-h: the method's own training (2--3 runs sharing an A100; edit generation, $\approx4.5$ A100-h, not included). $^\ddagger$: uses prior knowledge of the target shortcut.}
\label{tab:text-bert}
\small
\setlength{\tabcolsep}{3pt}
\begin{tabular}{lccccccccc}
\toprule
method & MNLI-m & MNLI-mm & HANS & non-ent & worst cell & SS worst & SS mean & CF flip & GPU-h \\
\midrule
ERM & 84.6\apm{0.2} & 84.8\apm{0.1} & 59.0\apm{1.2} & 20.0\apm{2.4} & 5.8\apm{1.9} & 53.1\apm{1.3} & 63.4\apm{0.6} & 10.9\apm{0.2} & 0.85 \\
PoE known-bias$^\ddagger$ & 83.3\apm{0.2} & 83.8\apm{0.1} & 69.8\apm{2.2} & 58.5\apm{3.2} & 38.7\apm{4.1} & 59.2\apm{0.2} & 68.9\apm{1.0} & 11.8\apm{0.1} & 1.48 \\
PoE self-debias & 82.5\apm{0.3} & 82.2\apm{0.4} & 59.8\apm{0.8} & 29.3\apm{1.3} & 19.2\apm{1.2} & 55.7\apm{0.8} & 63.1\apm{0.7} & 12.6\apm{0.2} & 1.45 \\
JTT ($\lambda_{\rm up}=6$) & 79.7\apm{0.4} & 79.9\apm{0.7} & 58.3\apm{0.3} & 33.8\apm{7.8} & 24.4\apm{6.4} & 54.7\apm{0.6} & 60.7\apm{0.3} & 15.2\apm{0.5} & 0.90 \\
\midrule
CF-aug, 12 instr.\ (4k) & 84.6\apm{0.0} & 84.9\apm{0.2} & 59.4\apm{0.5} & 20.1\apm{0.4} & 5.7\apm{2.0} & 52.9\apm{1.1} & 63.9\apm{0.2} & 8.8\apm{0.2} & 0.35 \\
CF-aug, 12k, CE+KL & 84.5\apm{0.3} & 84.6\apm{0.1} & 63.3\apm{0.4} & 28.5\apm{0.5} & 8.7\apm{3.1} & 54.2\apm{1.6} & 65.4\apm{0.1} & 8.6\apm{0.2} & 0.87 \\
CF-aug, 12k, aug-only & 84.5\apm{0.1} & 84.7\apm{0.0} & \textbf{64.7}\apm{1.3} & 31.7\apm{2.5} & 9.5\apm{0.6} & 54.6\apm{0.4} & 66.4\apm{0.7} & 8.8\apm{0.1} & 0.83 \\
\peel-style (4k) & 84.8\apm{0.1} & 84.9\apm{0.1} & 57.2\apm{1.2} & 15.5\apm{2.3} & 3.9\apm{0.2} & 52.1\apm{0.3} & 62.8\apm{0.5} & 8.6\apm{0.1} & 0.41 \\
\peel-style, 12k & 84.5\apm{0.1} & 84.8\apm{0.1} & 62.4\apm{1.8} & 27.3\apm{3.1} & 9.7\apm{1.0} & 55.0\apm{0.6} & 65.1\apm{0.9} & 8.7\apm{0.0} & 0.89 \\
CSI & 84.8\apm{0.2} & 84.9\apm{0.1} & 62.3\apm{0.4} & 27.7\apm{1.1} & 8.3\apm{1.9} & 53.8\apm{0.7} & 65.1\apm{0.2} & 8.5\apm{0.0} & 1.66 \\
MD-\peel{} (12k, aug-only) & 84.5\apm{0.2} & 84.6\apm{0.2} & 62.8\apm{0.7} & 26.9\apm{1.6} & 7.1\apm{2.7} & 53.7\apm{1.4} & 65.2\apm{0.6} & 10.8\apm{0.4} & 0.66 \\
MD-\peel{} (12k, CE+KL) & 84.3\apm{0.1} & 84.7\apm{0.1} & 62.9\apm{1.0} & 27.4\apm{2.2} & 7.4\apm{2.9} & 53.9\apm{1.7} & 65.5\apm{0.4} & 10.6\apm{0.4} & 0.49 \\
\bottomrule
\end{tabular}
\end{table}

\begin{table}[h!]
\centering
\caption{Second backbones (3 seeds). DeBERTa-v3-base uses the BERT settings. Qwen2.5-0.5B (base, linear head on the last token) is trained for 1 epoch of MNLI; CSI uses a reduced budget ($R_{\max}=3$) and CF-aug a matched 3k steps. Zero-shot prompting of Qwen2.5-7B-Instruct is one run. $^\S$\peel-style was re-run with the margin statistic one seed per GPU; the other rows ran three seeds per GPU, which inflates their wall-clock GPU-h.}
\label{tab:text-other}
\small
\setlength{\tabcolsep}{3.5pt}
\resizebox{\textwidth}{!}{%
\begin{tabular}{lccccccccc}
\toprule
method & MNLI-m & MNLI-mm & HANS & non-ent & worst cell & SS worst & SS mean & CF flip & GPU-h \\
\midrule
\multicolumn{10}{l}{\emph{DeBERTa-v3-base}} \\
ERM & 90.2\apm{0.2} & 90.5\apm{0.1} & 77.3\apm{0.3} & 54.8\apm{0.7} & 30.1\apm{2.6} & 63.6\apm{0.5} & 76.6\apm{0.2} & 5.5\apm{0.1} & 1.94 \\
CF-aug (4k) & 90.3\apm{0.1} & 90.5\apm{0.2} & 78.9\apm{0.4} & 58.0\apm{0.8} & 31.2\apm{2.0} & 65.3\apm{0.6} & 77.5\apm{0.2} & 4.0\apm{0.1} & 0.33 \\
CF-aug, 12k & 90.2\apm{0.1} & 90.5\apm{0.1} & 79.3\apm{0.6} & 58.8\apm{1.1} & 32.4\apm{0.5} & 65.4\apm{1.8} & 77.7\apm{0.8} & 3.9\apm{0.1} & 0.96 \\
\peel-style & 90.3\apm{0.0} & 90.5\apm{0.1} & 78.6\apm{1.5} & 57.3\apm{2.9} & 31.3\apm{1.2} & 65.6\apm{0.6} & 77.2\apm{0.5} & 4.1\apm{0.1} & 0.29$^\S$ \\
CSI & 90.2\apm{0.0} & 90.5\apm{0.0} & 79.7\apm{1.1} & 59.7\apm{2.2} & 33.6\apm{2.7} & 65.7\apm{0.4} & 77.8\apm{0.3} & 3.9\apm{0.1} & 1.99 \\
\midrule
\multicolumn{10}{l}{\emph{Qwen2.5-0.5B}} \\
ERM & 87.7\apm{0.1} & 87.7\apm{0.1} & 67.7\apm{1.3} & 36.4\apm{2.5} & 6.4\apm{0.8} & 53.1\apm{0.2} & 69.3\apm{0.8} & 7.7\apm{0.2} & 1.52 \\
CF-aug, 3k & 87.1\apm{0.2} & 87.0\apm{0.2} & 68.7\apm{0.8} & 38.7\apm{1.6} & 15.2\apm{2.6} & 57.9\apm{1.0} & 70.0\apm{0.5} & 6.5\apm{0.1} & 1.20 \\
\peel-style & 87.3\apm{0.0} & 87.2\apm{0.2} & 69.5\apm{0.5} & 39.7\apm{0.9} & 14.3\apm{2.1} & 57.2\apm{1.0} & 70.3\apm{0.2} & 6.6\apm{0.2} & 0.69$^\S$ \\
CSI & 87.5\apm{0.4} & 87.2\apm{0.2} & 69.3\apm{0.6} & 40.2\apm{1.8} & 16.0\apm{3.3} & 57.3\apm{1.2} & 70.1\apm{0.2} & 7.0\apm{0.2} & 2.48 \\
Zero-shot Qwen2.5-7B-Instruct & 73.8 & 76.1 & 82.2 & 88.9 & 69.6 & 49.4 & 69.7 & -- & 0.12 \\
\bottomrule
\end{tabular}}
\end{table}

\paragraph{MD-\peel{} on text.} The loop follows \cref{app:e4-md}, with the text as its own description. Trouble set (Variant L): per gold class, the misclassified and lowest-margin correct held-out pairs, 200 per class (123--180 are errors); Variant U uses predicted classes and disagreement with another ERM seed. History: the 500 highest-margin correct training pairs per class. Qwen2.5-7B-Instruct reads 50 history and 50 trouble pairs (without being told which group holds the failures) and lists differences in wording, length, structure, style or topic. Word and bigram presence statistics (Fisher's exact test, Holm) run in parallel, and the LLM merges both into at most 8 hypotheses per class, followed by the core filter. Test: REMOVE on 100 history and ADD on 100 trouble pairs, nulls $=$ light premise and light hypothesis paraphrase, statistic $=$ change in $f$'s logit margin, Holm. Audit: 50 accepted edits judged by the second judge (RoBERTa-large NLI), dropped if $\bar\delta>0.15$. Fix: one label-preserving rewrite (ADD or REMOVE with probability 1/2) of each of $\approx48.9$k pool pairs per confirmed hypothesis, then 12k steps from the ERM checkpoint with the same budget as the compute-matched controls. This deviates from retraining from scratch on every MNLI pair, which would need 390k rewrites per hypothesis.

\paragraph{Discovery confirms style, not overlap or negation.} \Cref{tab:text-md}. In every seed and variant, the confirmed hypotheses are style or complexity factors, plus the known neutral artifact ``because'' (seed 1, round 2). \textbf{No seed confirms a lexical-overlap or negation hypothesis} (0/3; 0/5 with two supplementary seeds). The automatic contrast does see the negation artifact: contradiction history hypotheses contain negation words far more often than trouble ones (e.g.\ seed 2: ``no'' 46\% vs.\ 4\%), and neutral history hypotheses contain ``because'' (35\% vs.\ 0\%). But the LLM merge mostly rewrites this evidence into style descriptions. Where a negation hypothesis survives, it is labelled ``core'' by the filter (seed 0), misses Holm (seed 2, $p=0.008$ against a threshold of $0.05/13$), or, in a post-hoc n-gram-only variant where ``the hypothesis contains `no'\,'' is significant in 3/3 seeds, fails the audit (the second judge rejects 8--12\% of the negation edits). Overlap was never proposed by any variant: neither the LLM reading pairs nor single-sentence n-gram statistics represent the premise--hypothesis relation. Post-hoc diagnostics weaken the confirmations further. A magnitude-matched null (a full label-aware paraphrase) explains about half of them, since any heavy label-aware rewrite makes a hard pair easier for $f$. The surviving REMOVE effects are not class-specific (making any confident pair more complex lowers $f$'s confidence in whatever its label is).

\paragraph{Fix.} MD-\peel{} reaches HANS $62.8\pm0.7$ vs.\ ERM $59.0\pm1.2$ ($+3.75\pm1.63$, $p=.03$), below the uniform CF-aug control with the same objective and budget ($64.7\pm1.3$; $-1.95\pm1.93$, $p=.15$, a tie by the pre-registered rule). Its deficit is concentrated in the constituent non-entailment cell ($23.6$ vs.\ $36.6$). Its CF flip rate does not improve ($10.8$ vs.\ ERM $10.9$; CF-aug $8.8$), because its edits do not cover the 12 instruction families the CF test uses. Over 5 seeds (2 supplementary, post hoc, with a truncated loop), MD-\peel{} $-$ uniform CF-aug $=-2.38\pm1.64$ HANS (paired $p=.04$). The pre-registered hypothesis H5 (discovery proposes overlap- or negation-type hypotheses, and the fix gives HANS $\ge$ uniform CF-aug at matched compute) is \textbf{not supported}: discovery fails, and the fix comparison is a tie, i.e.\ discovery was not needed. The full 3-seed MD-\peel{} loop cost $\approx6.7$ A100-h (discovery 1.03, fix rewrites 3.33, training 1.41, KL ablation 0.93).

\begin{table}[h!]
\centering
\caption{MD-\peel{} discovery on MNLI (BERT). Proposed: hypotheses (context after the filter). Confirmed: significant after Holm and passing the audit. Hypotheses quoted verbatim. No row has an overlap or negation hit.}
\label{tab:text-md}
\small
\begin{tabular}{llccp{8.2cm}}
\toprule
round & var./seed & proposed & sig.\ / conf. & confirmed hypotheses \\
\midrule
1 & L / 0 & 20 (18) & 2 / 1 & con: ``Shorter and fragmented sentences'' (neu ``Narrative style and hypothetical scenarios'' failed the audit, 8/50) \\
1 & L / 1 & 24 (21) & 6 / 6 & ent: ``Complex or indirect phrasing''; neu: ``Complex sentence structure and indirect statements.'', ``Narrative or descriptive style.''; con: ``Use of descriptive and emotive language'', ``Length and complexity of sentences'', ``Specificity and concreteness of scenarios'' \\
1 & L / 2 & 13 (13) & 1 / 1 & con: ``Sentence complexity and structure'' \\
1 & U / 0 & 24 (18) & 1 / 0 & -- (``Complex sentence structures'' failed the audit, 4/50) \\
1 & U / 1 & 24 (22) & 1 / 1 & con: ``Emotive language and expressions'' \\
1 & U / 2 & 22 (21) & 2 / 1 & con: ``Sentence structure complexity'' \\
\midrule
2 & L / 0 & 14 (12) & 1 / 1 & con: shorter vs.\ longer, more complex examples with anecdotes \\
2 & L / 1 & 23 (20) & 1 / 1 & neu: ``Frequency of `the' and `because' in the hypothesis'' \\
2 & L / 2 & 24 (20) & 1 / 0 & -- (``Length and detail of examples'' failed the audit, 4/50); seed 2 stops \\
3 & L / 0, 1 & 22, 19 & -- / 0 & none; the loop stops \\
\bottomrule
\end{tabular}
\end{table}

\begin{figure}[h!]
\centering
\includegraphics[width=0.85\textwidth]{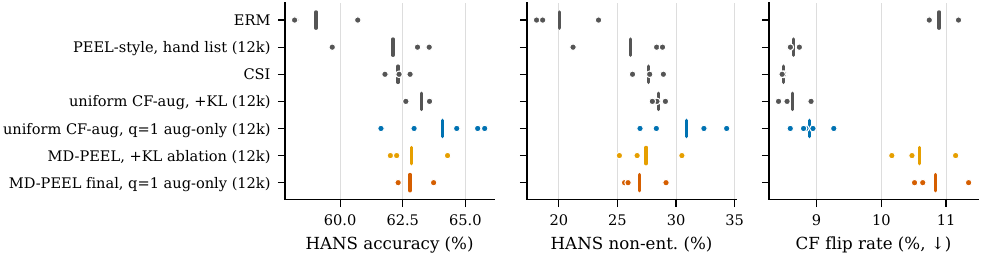}
\caption{MNLI$\to$HANS (BERT): per-seed HANS, HANS non-entailment and CF flip rate by method.}
\label{fig:text-hans}
\end{figure}
\FloatBarrier
}{}
\section{Experimental Details}\label{app:experiments}

This appendix documents the protocol of every experiment in \cref{sec:exp} and reports the full results. All numbers are copied from the result summaries generated by our analysis scripts. Unless stated otherwise, accuracies are in \%, entries are mean$\pm$sample standard deviation over seeds, and flip groups are written by the set of flipped shortcuts (e.g.\ $134$ is $G_{\{1,3,4\}}$). Labels are coded $y\in\{0,1\}$ in the code and $\{\pm1\}$ in the paper.

\subsection{Common protocol}
\label{app:protocol}

\paragraph{Coverage objective.} For a covered set $K$, each training example is replaced, with probability $q$, by an edited copy in which every attribute $a_j$, $j\in K$, is \emph{jointly} resampled uniformly and independently of the label (the label is kept). The loss is cross-entropy on the (possibly edited) batch plus $\mu\,\mathrm{KL}\big(p_\theta(\cdot\mid x)\,\|\,p_\theta(\cdot\mid Tx)\big)$ with $\mu=1$, no stop-gradient, averaged over edited examples. The main setting is $q=1$, which matches $\Ptil_{K,\delta}$: every example is edited every time it is seen, cross-entropy is computed only on edited examples, and originals enter only through the KL term. The pre-registered setting $q=0.5$ was run as a full replicate in E2 and E3 and as a mixture variant in E1 (without the KL term); it is reported separately below. The ``aug-only'' ablation sets $\mu=0$.

\paragraph{False accepts.} With injected false-accept rate $\delta$, each edited example is additionally, with probability $\delta$, given a label-changing edit (its core is replaced by a core of the other class) while keeping the original label. This realizes the $Z_c$ flip of $\Ptil_{K,\delta}$.

\paragraph{Salience.} The measured salience of shortcut $j$ for a model $f$ is $\hat\lambda_j=$ mean over validation inputs of $|z(x;a_j{=}y)-z(x;a_j{\neq}y)|$, where $z=$ logit$_1-$logit$_0$ and all other shortcuts are aligned. The core salience $\hat\lambda_{\rm core}$ is defined analogously with core swap. Salience is on the raw logit scale, which varies across runs, so only orderings should be compared across runs.

\paragraph{\peel{} hyper-parameters.} The edit library consists of one randomizing edit per shortcut, null (nuisance) edits that affect neither label nor shortcuts, and core swap, a label-changing edit that plays the false accept. The simulated validator always accepts valid edits and accepts core-swap proposals with probability $\delta_{\rm val}=0.1$, so every accepted core swap is bad. \emph{Evaluate}: $\widehat\Delta_t$ is the mean absolute change of the current model's logit margin ($z_1-z_0$; for more than two classes, the margin of the originally predicted class) over $n_{\rm eval}=1000$ validator-accepted validation edits of type $t$ (about $100$ for core swap). Type $t$ is significant if its per-example changes exceed those of the null edit with the largest mean by a one-sided Welch $t$-test, with Holm correction over types at $\alpha=0.05$. This is the statistic of \cref{prop:greedy}. Our first runs used the flip rate (a two-proportion $z$-test against the largest null flip rate) instead. We re-ran every \peel{} experiment with the margin statistic, and all reported \peel{} numbers use it. Where the flip-rate runs differ, we say so. \emph{Audit}: $m=50$ accepted edits of each significant type are checked against ground truth (known in simulation), giving a one-sided $95\%$ Clopper--Pearson upper bound $\bar\delta_t$. Types with $\bar\delta_t>\delta_{\max}=0.15$ are dropped. With $0/50$ bad edits, $\bar\delta_t=0.058$; with $50/50$, $\bar\delta_t=1.00$. \emph{Learn}: surviving types are added to the bank and the model is retrained from the initial weights with the coverage objective ($q=1$). The loop stops when no new type is kept, with at most $5$ rounds. Variants: \emph{greedy} keeps only the top-$\widehat\Delta$ surviving type per round; \emph{head} freezes the ERM backbone and retrains only the last layer; \emph{no-audit} skips the audit, so accepted core swaps enter the bank; \emph{aug-only} uses $\mu=0$ (E2).

\paragraph{Certificate.} For the final model and each bank type $t$ we compute \eqref{eq:cert} per type,
$\mathrm{bound}_t=\widehat{\err}_{\rm val}+\varepsilon_{n_{\rm v}}^{\alpha/3}+\big(\widehat\Gamma^{\rm acc}_t+\varepsilon_{n_t}^{\alpha/3\tau}\big)/(1-\bar\delta_t)$ with $\alpha=0.05$ and $\tau$ the number of bank types, and compare it with the error measured on the IID test set edited by type $t$. \Cref{thm:cert} requires all three samples to be independent of $f$ and of the bank. Discovery selects the bank and, through it, $f$, and the discovery audit decides which types are kept. The certificate therefore never reuses them. Its $n_{\rm v}$ points and $n_t$ edits come from a split that neither training nor discovery touched (E2: the certificate half of the validation set; E3: a fixed random half of the test birds, with the bound checked on the other half). Its audit is drawn afresh on that split, and $\bar\delta_t$ is the Clopper--Pearson bound at level $\alpha/(3\tau)$ ($0/50$ bad gives $\bar\delta_t=0.091$ for $\tau=2$ and $0.104$ for $\tau=4$), not the $95\%$ bound used to filter types during discovery. By \cref{thm:cert}, the maximum over $t$ bounds every mixture in $\mathcal Q$. It does not bound compositions of edits (see E3).

\paragraph{Model selection.} No group labels are used for training or model selection by any method without a $\dagger$. We use the last epoch of a fixed schedule; no hyper-parameter search was performed for our methods.

\paragraph{$\dagger$ baselines.} JTT$^\dagger$ \citep{liu2021just}, GroupDRO$^\dagger$ \citep{sagawa2020distributionally} and DFR$^\dagger$ \citep{kirichenko2023last} use group labels for model selection (JTT, GroupDRO) or for the retraining set (DFR), as in their papers. Their settings are given per experiment.

\paragraph{Compute and software.} E1 runs on a central processing unit (CPU) (about 25 minutes with at most 16 processes). E2 and E3 ran on an academic high-performance computing (HPC) cluster on NVIDIA A100 (PCIe, 40\,GB) GPUs, one GPU per job, with several small runs packed per job. The training wall-clock times recorded in the run files sum to about $30.0$\,h for E2 (several runs often shared one GPU concurrently, so this overstates GPU-hours) and about $15.7$\,h for E3 (including watermark tuning; evaluation excluded). Typical per-run times: a 40-epoch ResNet-18 run in E2 takes a median of $2.7$ minutes, and a 20-epoch ResNet-50 run in E3 a median of $2.3$ minutes ($5.5$ minutes for 60 epochs). Software: Python with PyTorch 2.10 (CUDA 12.8), torchvision 0.25 and timm 1.0.28; mixed precision (bf16 autocast) was used for E2 and E3.

\subsection{E1: stylized model}
\label{app:e1}

\paragraph{Setup.} Figures write $(\star)$ for the decision rule of \cref{thm:partial}: $G_F$ is correct iff $\Mmarg_K(F)>0$. Data follow \eqref{eq:model}. The classifier is linear without intercept (the model is sign-symmetric, so the population intercept is $0$). Because the logistic loss of $(x,y)$ depends only on the sign pattern $yx$, a training set is summarized exactly by counts over the $2^{k+1}$ patterns, and the finite-$n$ fits are exact maximum-likelihood estimates on $n$ i.i.d.\ samples (Newton's method). A core-intact flip group is a single point in feature space, so group accuracies are computed exactly from the fitted weights. With a noisy test core, $\mathrm{acc}_F=p\,\1[\lambda(p)+A_F>0]+(1-p)\,\1[-\lambda(p)+A_F>0]$ with $A_F$ the shortcut part of the margin; group accuracy then takes only the values $\{0,1-p,p,1\}$. Ties count as $0.5$ (none occurred). ``Theory setting'' means $q=1$; we also ran $q=0.5$ (the E2/E3 pre-registered protocol without the consistency term).

\begin{figure}[h!]
\centering
\includegraphics[width=\textwidth]{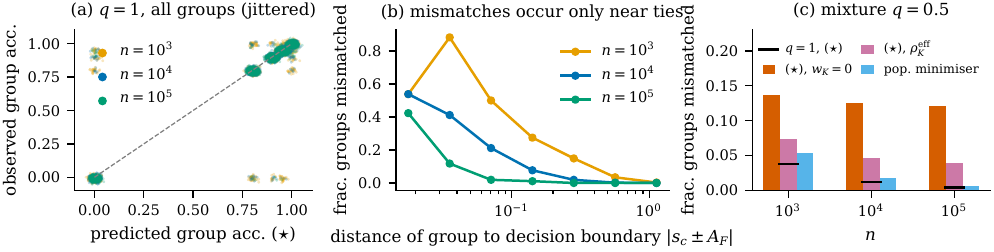}
\caption{E1, decision rule. (a) Predicted vs.\ observed group accuracy at $q=1$ (jittered). (b) Mismatch rate vs.\ distance of the group to the decision boundary: mismatches occur only near ties and shrink with $n$. (c) Mixture $q=0.5$: mismatch rate of the rule with $w_K=0$, with the effective $\rho^{\rm eff}_K=(1-q)\rho+q/2$, and of the exact population minimizer; the black bar is the $q=1$ rate.}
\label{fig:e1-star}
\end{figure}

\paragraph{Decision rule (\cref{thm:partial}).} $k\in\{3,5\}$, $p\in\{0.8,0.9,0.95\}$, $\rho_j\sim U[0.6,0.98]$, 10 seeds (each with its own $\rho$ draw and data), all 8 covers for $k=3$ and the 6 salience-prefix covers for $k=5$, all $2^k$ flip groups, $\delta=0$; 7{,}680 group evaluations per $(q,n)$. \Cref{tab:e1-star} and \cref{fig:e1-star} report the results. At $q=1$ the fitted weights match $\lambda(\cdot)$ (median max-abs deviation $0.021$ at $n=10^5$, worst $0.13$ when $\rho=0.98$), and at $n=10^5$ all $30$ of $7{,}680$ mismatches have margin below $0.10$. The maximum absolute error stays at $0.95$ for every $n$ because accuracy jumps from $p$ to $0$ at the boundary; a near-tie group on the wrong side produces this error. At $q=0.5$ mismatches do not vanish with $n$: the covered weights are not $0$ but $\approx\lambda((1-q)\rho_j+q/2)$ (e.g.\ $0.85$ for $\rho=0.9$). Applying the decision rule with this effective $\rho$ lowers the mismatch rate to $4.0\%$, and the exact population minimizer of the mixture lowers it to $0.6\%$ at $n=10^5$ (\cref{fig:e1-star}(c)). The remainder is non-product structure: the mixture correlates the covered coordinates.

\begin{table}[h!]
\centering
\caption{E1: decision rule of \cref{thm:partial} vs.\ exact finite-sample fits (7{,}680 group evaluations per row). ``Margin'' is the distance of the group's points to the theoretical boundary in log-odds units.}
\label{tab:e1-star}
\small
\begin{tabular}{llcccc}
\toprule
$q$ & $n$ & frac.\ mismatched (mean$\pm$sd) & mean abs.\ err. & max abs.\ err. & max abs.\ err.\ (margin $\ge0.2$) \\
\midrule
1 & $10^3$ & $0.038\pm0.012$ & 0.019 & 0.95 & 0.95 \\
1 & $10^4$ & $0.012\pm0.007$ & 0.006 & 0.95 & 0.95 \\
1 & $10^5$ & $\mathbf{0.0039\pm0.0043}$ & 0.002 & 0.95 & \textbf{0.00} \\
\midrule
0.5 & $10^3$ & $0.137\pm0.015$ & 0.069 & 0.95 & 0.95 \\
0.5 & $10^4$ & $0.125\pm0.017$ & 0.063 & 0.95 & 0.95 \\
0.5 & $10^5$ & $0.122\pm0.018$ & 0.061 & 0.95 & 0.95 \\
\bottomrule
\end{tabular}
\end{table}

\paragraph{Whac-a-Mole (\cref{cor:wam}).} \emph{Illustration} (\cref{fig:e1}(a)): $q=1$, $n=10^5$, 10 seeds, $p=0.8$, $\rho=(0.95,0.88,0.60)$, so $\lambda(p)=1.386$ and $\lambda(\rho)=(2.944,1.992,0.405)$. For $F=\{2\}$ and $j=1$ the condition of \cref{cor:wam}(b) holds ($1.79\le1.99<4.73$). Group $\{2\}$ is correct under $K=\varnothing$, wrong in $10/10$ seeds under $K=\{1\}$, and correct again under $K=\{1,2\}$. WGA rises from $0$ ($K=\varnothing,\{1\}$) to $1$ ($K=\{1,2\}$), and observed accuracy equals the prediction on all 24 $(K,F)$ bars in all seeds. With $q=0.5$ the illustration fails: the $S_1$ weight is not removed, so group $\{2\}$ stays correct. \emph{Aggregate}: 200 random $k=3$ instances, $n=10^5$, all $(K,j\notin K,F)$ triples (19{,}200 per $q$), in \cref{tab:e1-wam}. The set of triples where the rule predicts a decrease coincides exactly with the interval condition of \cref{cor:wam}(b); all mismatches at $q=1$ are near ties.

\begin{table}[h!]
\centering
\caption{E1: Whac-a-Mole over 200 random $k=3$ instances ($n=10^5$, 19{,}200 triples per $q$).}
\label{tab:e1-wam}
\small
\begin{tabular}{lcc}
\toprule
 & $q=1$ & $q=0.5$ \\
\midrule
sign of $\Delta$acc ($-/0/+$) matches the rule & \textbf{99.4\%} & 88.1\% \\
\cref{cor:wam}(b) predicted-decrease triples (813): observed decrease & \textbf{98.4\%} & 34.6\% \\
$j\notin F$, \cref{cor:wam}(b) condition not met (8{,}787): observed unchanged & 99.8\% (never increased) & 98.9\% \\
$j\in F$ (9{,}600): observed decrease & 0\% & 0\% \\
\bottomrule
\end{tabular}
\end{table}

\paragraph{False-accept threshold (\cref{thm:validator}(i)).} $k=5$, $p=0.95$, $\lambda(\rho)=(2.5,1.5,1.0,0.6,0.4)$, coverage $=$ salience prefix of length $m$, $\delta\in\{0,0.01,\dots,0.40\}$, $n=10^6$, 10 seeds, WGA over all 32 core-intact flip groups (\cref{tab:e1-delta}). At $q=1$ the fitted core weight matches $\lambda(p_\delta)$ to within $0.023$ (and to $10^{-15}$ for the population fit). At $q=0.5$ the threshold does not carry over by rescaling $\delta\to q\delta$: covered weights stay nonzero, and false accepts occur only in edited samples, where $z_K$ is uniform, so they are correlated with the covering. The population core weight then deviates from $\lambda(p_\delta)$ by up to $0.24$.

\begin{table}[h!]
\centering
\caption{E1: predicted threshold $\delta^\star$ vs.\ the largest $\delta$ on the grid (step $0.01$) with WGA $=1$.}
\label{tab:e1-delta}
\small
\begin{tabular}{ccccc}
\toprule
$m$ & $\delta^\star(m)$ & observed, $q=1$ ($n=10^6$) & $q=0.5$: naive $\delta^\star/q$ & $q=0.5$: observed (pop.\ min.) \\
\midrule
0 & $<0$ & never robust & $<0$ & never \\
1 & $<0$ & never robust & $<0$ & never \\
2 & 0.077 & 0.07 & 0.154 & \textbf{never} \\
3 & 0.243 & 0.24 & 0.487 & 0.06 \\
4 & 0.390 & 0.38 (pop.\ 0.39) & 0.781 & 0.17 \\
5 & 0.500 & $>0.40$ (grid end) & 1.0 & 0.25 \\
\bottomrule
\end{tabular}
\end{table}

\begin{figure}[h!]
\centering
\includegraphics[width=\textwidth]{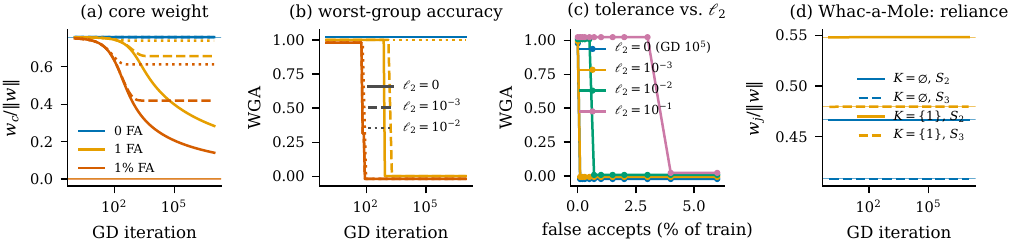}
\caption{E1, separable regime. (a) Normalized core weight along gradient descent with 0, 1 and $1\%$ false accepts (dashed/dotted: $\ell_2=10^{-3}/10^{-2}$). (b) WGA collapses as soon as a false accept is fitted. (c) Number of false accepts tolerated grows with $\ell_2$ regularization. (d) Configuration B: covering $S_1$ raises the normalized reliance on $S_2$ and $S_3$.}
\label{fig:e1-sep}
\end{figure}

\paragraph{Separable regime (\cref{thm:partial}(ii), \cref{thm:validator}(ii)).} Configuration A: $p=1$, $\rho_U=1$, $a=1$, $b=(0.9,0.7,0.5)$, $K=\{1\}$, $q=1$, $n=1000$, 3 seeds, full-batch gradient descent with step size 1 for $10^7$ steps. Since $a^2=1>b_2^2+b_3^2=0.74$, every group is correct with a perfect validator. Results are in \cref{tab:e1-sep} and \cref{fig:e1-sep}. Hard-margin support vector machine (SVM) directions match the theory: $(0.758,0,0.531,0.379)=(a,0,b_U)/\|\cdot\|$ without false accepts and $(0.000,0.000,0.814,0.581)=(0,0,b_U)/\|b_U\|$ with 1 or 10 false accepts. Gradient descent approaches these directions slowly (cosine $0.954$ for one false accept and $0.990$ for $1\%$ after $10^7$ steps), and regularized gradient descent converges to the exact $\ell_2$-regularized minimizer (agreement $\le10^{-12}$). Worst-group accuracy breaks after about $10^3$ steps, long before the direction converges, because with $w_K\approx0$ the worst group's point is the negation of the false accept's (\cref{sec:theory}). Tolerance vs.\ $\ell_2$ (5 seeds, $n=1000$): $0$ false accepts tolerated for $\ell_2\in\{0\ (\text{GD }10^5\text{ steps}),10^{-4},10^{-3}\}$, at most $5$ ($0.5\%$) for $\ell_2=10^{-2}$, and at most $30$ ($3\%$) for $\ell_2=10^{-1}$. Configuration B (Whac-a-Mole in the max-margin regime, $b=(0.9,0.8,0.7)$, no false accepts): covering $S_1$ raises the normalized reliance on $S_2$ from $0.467$ to $0.548$ and on $S_3$ from $0.408$ to $0.480$, exactly $b_j/\|(a,b_U)\|$ (gradient descent matches with cosine $\ge0.99999$). Group $\{2,3\}$ flips from correct ($1+0.81>1.13$) to wrong ($1<1.13$) at every checkpoint.

\begin{table}[h!]
\centering
\caption{E1, separable regime (configuration A). Normalized core weight $w_c/\|w\|$ along gradient descent, its max-margin limit, and WGA. ``FA'' = false accepts among $n=1000$.}
\label{tab:e1-sep}
\small
\begin{tabular}{cccccc}
\toprule
FA & $\ell_2$ & $w_c/\|w\|$ at $10^3$ / $10^5$ / $10^7$ steps & max-margin $w_c/\|w\|$ & WGA at $10^7$ & first step with WGA $<1$ \\
\midrule
0 & 0 & 0.758 / 0.758 / 0.758 & 0.758 & 1 & -- \\
1 (0.1\%) & 0 & 0.665 / 0.398 / 0.283 & \textbf{0.000} & 0 & \textbf{873} \\
10 (1\%) & 0 & 0.392 / 0.206 / 0.140 & \textbf{0.000} & 0 & 67 \\
1 & $10^{-3}$ & 0.657 (converged) & -- & 0 & 1501 \\
1 & $10^{-2}$ & 0.740 (converged) & -- & \textbf{1} & -- \\
10 & $10^{-2}$ & 0.612 (converged) & -- & 0 & 76 \\
\bottomrule
\end{tabular}
\end{table}

\paragraph{Greedy discovery (\cref{prop:greedy}).} 100 random instances each for $k=3$ and $k=5$, $n=10^5$ training and $10^4$ validation samples, $q=1$. Each round refits the model and covers the uncovered edit with the largest mean $|\Delta\text{logit}|$ (\cref{tab:e1-greedy}). Every order mismatch is a near-tie in salience: the median $\min|\lambda_i-\lambda_j|$ is $0.006$--$0.007$ among mismatched instances, against $0.24$ ($k=3$) and $0.067$ ($k=5$) overall. Rule (ii) of \cref{prop:greedy} never stopped early: before full coverage, the smallest mean $|\Delta\text{logit}|$ of an uncovered shortcut was $0.37$ (median $0.68$), against $0$ for null edits. Ranking by population flip rate (the statistic of our first \peel{} runs) also never contradicted the salience order: $0$ strict violations of $\lambda_i>\lambda_j\Rightarrow\Gamma_i\ge\Gamma_j$ in 18{,}000 random instances ($k\in\{2,3,4\}$, $\rho\in[0.51,0.999]$, $p\in[0.52,0.999]$). In finite samples, the round-1 flip-rate pick agreed with the $|\Delta\text{logit}|$ pick in only $90$--$91\%$ of instances, because flip rates are often nearly tied. This is one more reason \peel{} ranks by $|\Delta\text{logit}|$.

\begin{table}[h!]
\centering
\caption{E1: greedy discovery (100 instances per $k$).}
\label{tab:e1-greedy}
\small
\begin{tabular}{lcc}
\toprule
 & $k=3$ & $k=5$ \\
\midrule
full discovery order $=$ salience order & 98\% & 94\% \\
first $m^\star$ picks $=$ top-$m^\star$ salient & 99\% & 96\% \\
first robust round $=m^\star$ (\cref{cor:suff}) & 99\% & 97\% \\
rule (ii) runs until all $k$ are covered & 100\% & 100\% \\
flip-rate rule (alternative) fires at round $m^\star$ & 99\% & 97\% \\
\bottomrule
\end{tabular}
\end{table}

\begin{figure}[h!]
\centering
\includegraphics[width=0.55\textwidth]{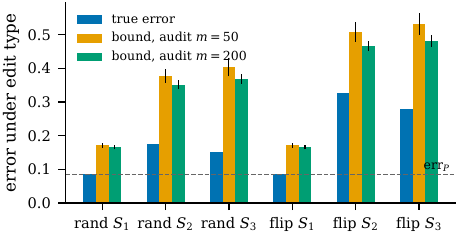}
\caption{E1: per-type certificate (mean$\pm$sd over 200 replications) vs.\ true error, for audit sizes $m=50$ and $m=200$. The dashed line is $\err_P$.}
\label{fig:e1-cert}
\end{figure}

\paragraph{Certificate (\cref{thm:cert}).} Model: maximum-likelihood fit with $K=\{1\}$, $p=0.9$, $\rho=(0.9,0.8,0.75)$, $\err_P=0.085$. $\tau=6$ valid edit types: rand$_j$ (randomize $S_j$) and flip$_j$ (set $S_j$ misaligned). Per-type false-accept rates $\delta_t\in\{0.02,0.05,0.10\}$ (a false accept also flips the core). $n_{\rm v}=2000$, $n_t=2000$ accepted pairs, audit $m_t\in\{50,200\}$, $\alpha=0.05$, 200 independent Monte-Carlo replications with fresh validation, edit and audit samples; true errors are exact (\cref{tab:e1-cert}, \cref{fig:e1-cert}). Most of the slack comes from the Clopper--Pearson bound at level $\alpha/(3\tau)$ with $m=50$ ($\bar\delta\approx0.15$--$0.27$ for true $\delta$ of $0.02$--$0.10$) and from false accepts inflating $\widehat\Gamma^{\rm acc}$. Two further findings motivated the final statement of \cref{thm:cert}. (i) Compositions are not covered: for this model, flip$_2$ and flip$_3$ have $\Gamma=0.240$ and $0.195$, so summing per-type terms gives $0.52$, while the composition (flip group $\{2,3\}$) has true error $1.00$. (ii) A validator that preferentially rejects valid edits that change $f$ breaks the bound: rejecting $90\%$ of prediction-changing valid edits gives a violation rate of $98\%$ ($m=200$), while $50\%$ rejection gives no violations. This is why \cref{ass:accept} and the discussion after \cref{thm:cert} are needed.

\begin{table}[h!]
\centering
\caption{E1: certificate over 200 replications. Left: uniform bound over all mixtures. Right: per-type bounds ($m=50$, mean).}
\label{tab:e1-cert}
\small
\begin{tabular}{lcc}
\toprule
 & $m=50$ & $m=200$ \\
\midrule
bound (uniform over mixtures), mean$\pm$sd & $0.542\pm0.033$ & $0.484\pm0.015$ \\
true worst-mixture error (flip $S_2$) & 0.325 & 0.325 \\
violation rate, worst mixture & \textbf{0/200} & \textbf{0/200} \\
violation rate, 100 random Dirichlet mixtures & 0 & 0 \\
mean true error of random mixtures & 0.172 & 0.172 \\
\bottomrule
\end{tabular}
\hspace{1em}
\begin{tabular}{lcc}
\toprule
edit type & true error & bound \\
\midrule
rand $S_1$ & 0.085 & 0.171 \\
rand $S_2$ & 0.175 & 0.377 \\
rand $S_3$ & 0.150 & 0.401 \\
flip $S_1$ & 0.085 & 0.171 \\
flip $S_2$ & 0.325 & 0.507 \\
flip $S_3$ & 0.280 & 0.534 \\
\bottomrule
\end{tabular}
\end{table}

\FloatBarrier
\subsection{E2: four shortcuts on CIFAR-10}
\label{app:e2}

\paragraph{Data and shortcuts.} Core task: CIFAR-10 cat ($y=0$) vs.\ dog ($y=1$), $4{,}000$ training and $1{,}000$ validation images per class (from the CIFAR-10 training set) and the $1{,}000$ test images per class. The data and attribute seed is fixed (1234) across runs; the run seed controls initialization, data order, augmentation and edit sampling. Four binary shortcuts are rendered exactly on the $32\times32$ images, each with $\rho_j=0.95$, independent given $y$. Render order: grating $\to$ tint $\to$ clip $\to$ border ($\alpha$-blend) $\to$ patch (overwrite) $\to$ clip.
\begin{itemize}[leftmargin=*,itemsep=0pt,topsep=2pt]
\item $S_1$, corner patch: $4\times4$ top-left, white ($a=1$) or black ($a=0$); it overwrites the border's corner pixels.
\item $S_2$, tint: $+0.06$ added to the red ($a=1$) or blue ($a=0$) channel.
\item $S_3$, border: 2 pixels, green ($a=1$) or magenta ($a=0$), $\alpha=0.5$.
\item $S_4$, grating: amplitude $0.02$, phase-shifted sinusoid with period 4 pixels, horizontal ($a=1$) or vertical ($a=0$) stripes.
\end{itemize}
Random crop (4-pixel zero padding) and horizontal flip are applied to the core only; shortcuts are rendered afterwards, so attributes stay exact. Every flip group is evaluated on the full test set (2{,}000 images) by re-rendering.

\paragraph{Deviation: shortcut amplitudes.} The pre-registered amplitudes (tint $0.12$, grating $0.04$) gave an unstable salience ordering across ERM seeds, with the logit scale itself varying by about $5\times$ (e.g.\ $\hat\lambda=15.4/13.4/15.9/2.2$ in seed 0, $19.3/16.9/18.0/18.1$ in seed 2, $37/50/52/48$ in seed 3). We screened 8 amplitude settings with ERM and adopted tint $0.06$ and grating $0.02$ (patch and border unchanged), which gave the consistent ordering $S_1\gtrsim S_3>S_4>S_2$ in $3/3$ screening seeds. Over the 5 main ERM seeds the saliences are $\hat\lambda=7.52/3.56/7.37/3.56$. Single-flip groups mostly stay $\ge93\%$ under ERM because three aligned shortcuts outvote one, so shortcut use shows in pair groups: in seeds 0, 2, 3, 4 every pair group drops below IID accuracy. $S_4$ is fragile: in ERM seed 1 it is essentially unused ($\hat\lambda_4=0.16$). ERM core reliance is $0.0\%$ in every seed.

\paragraph{Training.} CIFAR ResNet-18 from scratch, SGD (learning rate $0.1$, momentum $0.9$, weight decay $5\cdot10^{-4}$), cosine schedule, 40 epochs, batch 256, bf16 autocast, last epoch. False accepts replace the core of an edited example by a random training image of the other class (with the same crop/flip augmentation). Null edits for \peel: Gaussian noise ($\sigma=0.02$, after rendering), a 2-pixel reflect-padded translation of the core (shortcuts re-rendered), and contrast jitter $c\in[0.8,1.2]$ on the core before rendering. Interpolating regime: no weight decay, 150 epochs, no crop/flip (edits are still resampled every epoch). Core reliance is the accuracy w.r.t.\ the \emph{core} label on core-swapped test images whose shortcuts all agree with the original label (``aligned''), or are uniformly random (``random'').

\paragraph{Coverage sweep.} \Cref{tab:e2-cov} lists all 16 covers at $q=1$ (5 seeds) together with WGA and IID accuracy at $q=0.5$ (3 seeds). At $q=1$, covered shortcuts have mean measured salience $0.104$ (median $0.084$, max $0.38$) against $5.27$ for uncovered ones ($5.50$ under ERM). At $q=0.5$ the covered median is $0.40$ (mean $0.47$), and at full coverage the residual $\hat\lambda\approx0.3$ per shortcut sums to about $1.2$ logits, giving WGA $40.3\pm11.7$ against $70.9\pm4.0$ at $q=1$. WGA is at the floor for 13 of 16 covers because the core is weaker than every single shortcut. By \cref{cor:suff}, only full coverage can then make all groups correct, which is what we observe: WGA by $|K|$ is $0.0/0.1/0.7/6.3/70.9$. The only partial cover with non-trivial WGA is $\{1,3,4\}$, which leaves the weakest shortcut $S_2$ uncovered ($22.1\pm2.3$).

\begin{table}[h!]
\centering
\caption{E2 coverage sweep. Left block: $q=1$ (5 seeds); ``worst'' is the modal worst flip group, ``core'' is core reliance (aligned shortcuts), $\hat\lambda_j$ are mean saliences. Right block: $q=0.5$ (3 seeds). One $q=0.5$ seed with $K=\{4\}$ has a $10\times$ logit scale.}
\label{tab:e2-cov}
\footnotesize
\setlength{\tabcolsep}{3.5pt}
\begin{tabular}{lcccccccccc|cc}
\toprule
& \multicolumn{10}{c|}{$q=1$} & \multicolumn{2}{c}{$q=0.5$} \\
$K$ & WGA & worst & mean grp & IID & core & $\hat\lambda_1$ & $\hat\lambda_2$ & $\hat\lambda_3$ & $\hat\lambda_4$ & & WGA & IID \\
\midrule
$\varnothing$ & $0.0\pm0.0$ & 134 & $51.1\pm0.2$ & $99.2\pm0.4$ & $0.0\pm0.0$ & 7.52 & 3.56 & 7.37 & 3.56 & & $0.0\pm0.0$ & $99.1\pm0.6$ \\
1 & $0.3\pm0.3$ & 234 & $51.6\pm0.5$ & $97.6\pm1.1$ & $0.4\pm0.4$ & 0.14 & 2.56 & 5.65 & 3.66 & & $0.2\pm0.1$ & $97.8\pm0.8$ \\
2 & $0.0\pm0.0$ & 134 & $50.3\pm0.3$ & $99.5\pm0.1$ & $0.0\pm0.0$ & 6.91 & 0.19 & 7.04 & 5.74 & & $0.0\pm0.0$ & $98.7\pm1.1$ \\
3 & $0.0\pm0.0$ & 124 & $52.3\pm0.5$ & $97.8\pm0.1$ & $0.0\pm0.0$ & 6.25 & 2.67 & 0.21 & 5.89 & & $0.0\pm0.0$ & $98.2\pm0.1$ \\
4 & $0.0\pm0.0$ & 123 & $51.3\pm0.2$ & $98.3\pm0.2$ & $0.0\pm0.0$ & 6.75 & 3.29 & 6.72 & 0.06 & & $0.0\pm0.0$ & $98.6\pm0.2$ \\
12 & $0.0\pm0.0$ & 34 & $52.8\pm1.2$ & $96.3\pm0.4$ & $0.0\pm0.0$ & 0.09 & 0.12 & 5.90 & 5.26 & & $0.0\pm0.0$ & $97.5\pm0.2$ \\
13 & $2.4\pm0.6$ & 24 & $53.9\pm1.1$ & $96.1\pm0.2$ & $2.1\pm0.6$ & 0.10 & 2.95 & 0.30 & 6.12 & & $0.9\pm0.1$ & $96.8\pm0.3$ \\
14 & $0.8\pm0.4$ & 23 & $51.7\pm0.8$ & $96.0\pm0.2$ & $0.9\pm0.5$ & 0.11 & 3.37 & 6.40 & 0.03 & & $0.5\pm0.3$ & $96.6\pm0.2$ \\
23 & $0.0\pm0.0$ & 14 & $54.4\pm1.0$ & $96.3\pm0.4$ & $0.0\pm0.0$ & 6.05 & 0.14 & 0.15 & 5.69 & & $0.0\pm0.0$ & $97.8\pm0.4$ \\
24 & $0.0\pm0.0$ & 13 & $54.1\pm0.9$ & $96.4\pm0.4$ & $0.0\pm0.0$ & 6.00 & 0.05 & 5.95 & 0.03 & & $0.0\pm0.0$ & $97.5\pm0.5$ \\
34 & $1.0\pm0.1$ & 12 & $52.2\pm0.4$ & $95.2\pm0.2$ & $0.9\pm0.2$ & 6.00 & 2.78 & 0.17 & 0.04 & & $0.4\pm0.1$ & $95.7\pm0.1$ \\
123 & $2.9\pm1.7$ & 24 & $51.4\pm0.8$ & $95.6\pm0.1$ & $3.1\pm1.9$ & 0.05 & 0.14 & 0.14 & 6.05 & & $0.5\pm0.3$ & $95.8\pm0.1$ \\
124 & $0.0\pm0.0$ & 3 & $50.0\pm0.0$ & $96.1\pm0.0$ & $0.0\pm0.0$ & 0.03 & 0.04 & 5.82 & 0.01 & & $0.0\pm0.0$ & $96.1\pm0.0$ \\
134 & $22.1\pm2.3$ & 2 & $55.8\pm1.7$ & $84.5\pm1.2$ & $22.2\pm2.0$ & 0.06 & 3.14 & 0.17 & 0.08 & & $14.1\pm1.2$ & $89.1\pm0.6$ \\
234 & $0.0\pm0.0$ & 1 & $50.0\pm0.0$ & $95.0\pm0.0$ & $0.0\pm0.0$ & 6.00 & 0.07 & 0.05 & 0.02 & & $0.0\pm0.0$ & $95.0\pm0.0$ \\
1234 & $70.9\pm4.0$ & 123 & $72.4\pm3.2$ & $73.0\pm2.5$ & $71.2\pm3.6$ & 0.05 & 0.19 & 0.17 & 0.11 & & $40.3\pm11.7$ & $85.7\pm1.6$ \\
\bottomrule
\end{tabular}
\end{table}

\begin{figure}[h!]
\centering
\begin{minipage}[t]{0.48\textwidth}\centering
\includegraphics[width=\textwidth]{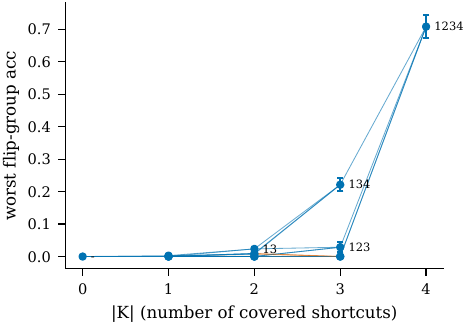}
\end{minipage}\hfill
\begin{minipage}[t]{0.48\textwidth}\centering
\includegraphics[width=\textwidth]{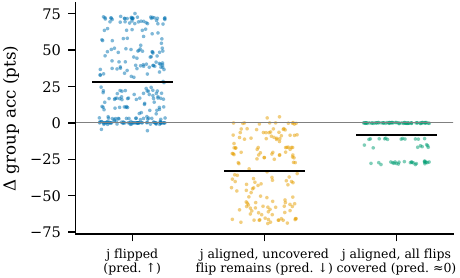}
\end{minipage}
\caption{E2 ($q=1$). Left: WGA over the coverage lattice (lines connect $K\subset K\cup\{j\}$). WGA stays near $0$ until the last shortcut is covered, except for $K=\{1,3,4\}$. Right: change in flip-group accuracy for all 512 (cover, shortcut, group) triples, by category; bars are means. The rightmost category (``no effect'' predicted) shows a mean loss of $8.5$ points.}
\label{fig:e2-lattice}
\end{figure}

\paragraph{\cref{cor:wam}(a).} Along the 32 cover-one-more pairs $K\subset K\cup\{j\}$ at $q=1$, WGA increases in $66\%$ of pairs and is non-decreasing within $0.5$ points in $94\%$, with mean change $+10.4$ points and minimum $-0.9$. Two decreases are significant over seeds (paired $t$-test, $p<0.05$): $K=\{1,4\}$ plus $S_2$ ($-0.77$) and $K=\{3,4\}$ plus $S_2$ ($-0.94$), both under 1 point and at the $0$--$3\%$ floor. The continuous worst-group statistics resolve the floor: the worst group's mean margin increases and its logistic loss decreases in all 32 pairs. At $q=0.5$, WGA is non-decreasing (within $0.5$ points) in all pairs and the worst-group margin increases in $97\%$.

\paragraph{\cref{cor:wam}(b).} \Cref{tab:e2-wam} groups the 512 triples $(K,j\notin K,F)$, with $\Delta=\mathrm{acc}_F(K\cup\{j\})-\mathrm{acc}_F(K)$. With significance defined as a two-sided paired $t$-test over seeds with $p<0.05$ and $|\Delta|>0.5$: at $q=1$, $72.5\%$ of triples ($371/512$) change significantly, and $97.6\%$ of these have the predicted sign (decrease iff $j$ is aligned in $F$). Of the 181 significant decreases, $95.0\%$ have $j$ aligned in $F$ ($70.2\%$ additionally have an uncovered flipped shortcut, i.e.\ category B). All 190 significant increases have $j$ flipped. At $q=0.5$, $52.7\%$ change significantly, with sign agreement $98.1\%$; $97.0\%$ of decreases have $j$ aligned. \emph{Per-example prediction}: from the $K$-model's per-example test margins, covering $j$ is predicted to subtract (if $j$ is aligned) or add back (if flipped) the example's own flip effect $D_j(x)=m_\varnothing(x)-m_{\{j\}}(x)$. The Pearson correlation of predicted and actual $\Delta$ is $0.85$ ($q=0.5$: $0.82$), and the sign agrees in $88.3\%$ of triples with $|\text{pred}|\ge1$ point ($94.8\%$ among significant changes). Of the 24 triples predicted to be unchanged ($|\text{pred}|<1$), $8.3\%$ change significantly, with mean $|\Delta|=0.17$ points.

\emph{Failure of the ``no-effect'' prediction.} Category C should be unchanged at the population level with an intact core. Instead it loses $8.5$ points on average. The loss is driven by covering the \emph{last} uncovered shortcut ($K\cup\{j\}=[4]$; 32 triples, mean $-24.5$ points), after which all groups fall from ${\approx}100\%$ to the core accuracy (${\approx}71\%$). The other 76 C-triples average $-1.8$ points (minimum $-11.3$). With a noisy core, hard examples are rescued by aligned shortcuts, and removing them costs accuracy; the per-example rule predicts this, the coarse population rule does not.

\begin{table}[h!]
\centering
\caption{E2: \cref{cor:wam}(b) categories over 512 triples ($q=1$, 5 seeds), with the predicted direction. ``Sig.'' uses a one-sided $t$-test over seeds, $p<0.05$; ``wrong'' is the fraction with significant wrong sign. In category C the largest $|\Delta|$ is $28.6$.}
\label{tab:e2-wam}
\small
\begin{tabular}{lccccc}
\toprule
category & $n$ & mean $\Delta$ & median $\Delta$ & predicted sign & sig.\ predicted sign \\
\midrule
B: $j$ aligned, uncovered flip remains ($\downarrow$) & 148 & $-32.7$ & $-27.7$ & 98.0\% & 93.9\% (wrong: 0.7\%) \\
A: $j$ flipped in $F$ ($\uparrow$) & 256 & $28.1$ & $20.3$ & 88.7\% & 79.3\% \\
C: $j$ aligned, all flips covered (no change) & 108 & $-8.5$ & $-0.3$ & $|\Delta|<1$: 59.3\% & sig.\ $\neq0$: 68.5\% \\
\bottomrule
\end{tabular}
\\[2pt]
{\footnotesize $q=0.5$ (3 seeds): B mean $-29.9$ (predicted sign $98.0\%$, sig.\ $65.5\%$); A mean $+24.5$; C mean $-8.4$.}
\end{table}

\paragraph{Additivity.} For each model, the margin of each test image in group $F$ is predicted as $m_\varnothing(x)-\sum_{j\in F}D_j(x)$, and group accuracy as the fraction of positive predicted margins. Over all models and groups with $|F|\ge2$ (880 (model, group) points at $q=1$): $R^2=0.996$, mean absolute error $1.0$ points, maximum error $16.0$, and $86.5\%$ within 2 points. The coarser prediction from the mean signed salience $\hat\lambda_j$ gives $R^2=0.971$ (mean absolute error $2.9$, maximum $69.8$). At $q=0.5$: $R^2=0.987$ and $0.921$.

\paragraph{Validator error.} Covers: the top-2 cover $\{1,3\}$ (pre-registered), full coverage, and $\{1,3,4\}$ (added because $\{1,3\}$ is at the WGA floor and cannot show a regime difference). \Cref{tab:e2-delta} gives WGA at $q=1$; \cref{tab:e2-delta-q05} at $q=0.5$. Interpolating runs reach $100.0\%$ clean training accuracy at $\delta\le0.05$ for every $K$; only with full coverage does training accuracy fall at large $\delta$ ($96.5\%$ at $\delta=0.2$, $67.9\%$ at $\delta=0.3$), where false-accepted cores conflict with clean ones. At $q=1$, WGA and core reliance decrease with $\delta$ in every (cover, regime) cell, but the decline is gradual, without the sharp threshold of the linear theory. For $\{1,3,4\}$, interpolation loses 12 points of WGA already at $\delta=0.02$ ($27.4\to15.5$) and ends at $8.4$ ($-69\%$ relative), against $20.5\to17.1\to11.4$ ($-44\%$) with regularization. For $\{1,3\}$ both regimes are at the floor, and core reliance with random shortcuts drops to chance in both ($53.8\to50.2$ interpolating, $53.6\to49.7$ regularized). With full coverage, interpolation is \emph{not} worse ($61.1$ vs.\ $54.7$ at $\delta=0.3$), consistent with the non-separability remark of \cref{thm:validator}(ii). These are partial confirmations of \cref{thm:validator}.

\emph{$q=0.5$ reversal.} At $q=0.5$ tolerance is much lower: with full coverage and regularization, WGA falls $49.8\to21.0$ ($\delta=0.1$) $\to1.2$ ($\delta=0.3$) while IID accuracy rises to $98.0$, i.e.\ the model reverts to the covered shortcuts, which remain predictive in the unedited half. The interpolating model holds up to $\delta=0.1$ ($69.7$) and then collapses ($50.7$ at $0.2$, $6.6$ at $0.3$). Here the \emph{regularized} regime is the more fragile one, contrary to our expectation; we report it as is.

\begin{table}[h!]
\centering
\caption{E2: WGA vs.\ injected false-accept rate $\delta$, $q=1$. Regularized (reg.): main schedule. Interpolating (interp.): no weight decay, 150 epochs, no crop/flip. 5 seeds, except 3 for $K=\{1,3,4\}$.}
\label{tab:e2-delta}
\footnotesize
\begin{tabular}{ccccccc}
\toprule
& \multicolumn{2}{c}{$K=\{1,2,3,4\}$} & \multicolumn{2}{c}{$K=\{1,3,4\}$} & \multicolumn{2}{c}{$K=\{1,3\}$} \\
\cmidrule(lr){2-3}\cmidrule(lr){4-5}\cmidrule(lr){6-7}
$\delta$ & reg. & interp. & reg. & interp. & reg. & interp. \\
\midrule
0 & $73.3\pm5.4$ & $74.5\pm2.4$ & $20.5\pm2.4$ & $27.4\pm3.2$ & $2.3\pm0.7$ & $3.8\pm0.3$ \\
0.02 & $67.5\pm4.7$ & $73.8\pm2.3$ & $17.1\pm0.6$ & $15.5\pm1.8$ & $1.4\pm0.5$ & $1.4\pm0.4$ \\
0.05 & $69.3\pm6.6$ & $71.1\pm2.3$ & $17.3\pm0.7$ & $13.0\pm0.8$ & $1.2\pm0.4$ & $1.1\pm0.2$ \\
0.1 & $63.7\pm4.9$ & $71.7\pm3.1$ & $14.9\pm1.4$ & $10.3\pm1.5$ & $0.6\pm0.1$ & $0.7\pm0.2$ \\
0.2 & $61.3\pm3.1$ & $67.2\pm3.7$ & $12.7\pm0.8$ & $8.5\pm1.4$ & $0.2\pm0.1$ & $0.4\pm0.2$ \\
0.3 & $54.7\pm1.3$ & $61.1\pm1.6$ & $11.4\pm1.2$ & $8.4\pm0.5$ & $0.2\pm0.1$ & $0.2\pm0.1$ \\
\bottomrule
\end{tabular}
\end{table}

\begin{table}[h!]
\centering
\caption{E2: WGA / IID accuracy vs.\ $\delta$ at $q=0.5$ (3 seeds).}
\label{tab:e2-delta-q05}
\footnotesize
\begin{tabular}{ccccc}
\toprule
& \multicolumn{2}{c}{$K=\{1,2,3,4\}$} & \multicolumn{2}{c}{$K=\{1,3\}$} \\
\cmidrule(lr){2-3}\cmidrule(lr){4-5}
$\delta$ & reg. & interp. & reg. & interp. \\
\midrule
0 & $49.8\pm8.7$ / $85.2\pm0.6$ & $71.6\pm0.4$ / $73.6\pm1.1$ & $0.9\pm0.1$ / $96.9\pm0.2$ & $4.2\pm0.6$ / $95.1\pm0.4$ \\
0.02 & $39.2\pm4.4$ / $85.6\pm0.5$ & $72.4\pm2.2$ / $76.1\pm1.0$ & $0.5\pm0.1$ / $96.6\pm0.4$ & $1.9\pm0.4$ / $95.6\pm0.3$ \\
0.05 & $25.9\pm7.8$ / $89.2\pm1.9$ & $68.6\pm1.9$ / $74.5\pm1.0$ & $0.5\pm0.1$ / $96.7\pm0.1$ & $1.6\pm0.2$ / $95.5\pm0.1$ \\
0.1 & $21.0\pm5.2$ / $90.9\pm1.2$ & $69.7\pm2.9$ / $77.9\pm2.7$ & $0.3\pm0.1$ / $96.5\pm0.2$ & $0.9\pm0.3$ / $95.8\pm0.2$ \\
0.2 & $9.2\pm3.3$ / $95.7\pm0.4$ & $50.7\pm9.1$ / $83.8\pm2.5$ & $0.2\pm0.1$ / $96.5\pm0.1$ & $0.5\pm0.1$ / $96.2\pm0.2$ \\
0.3 & $1.2\pm0.7$ / $98.0\pm0.2$ & $6.6\pm1.5$ / $97.2\pm0.9$ & $0.2\pm0.1$ / $96.2\pm0.1$ & $0.2\pm0.1$ / $96.3\pm0.2$ \\
\bottomrule
\end{tabular}
\end{table}

\begin{figure}[h!]
\centering
\includegraphics[width=\textwidth]{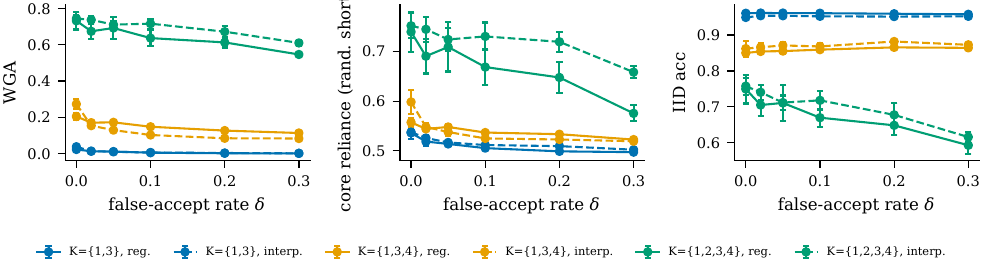}
\caption{E2 ($q=1$): WGA, core reliance (random shortcuts) and IID accuracy vs.\ false-accept rate, for three covers in the regularized (solid) and interpolating (dashed) regimes.}
\label{fig:e2-delta}
\end{figure}

\paragraph{\peel.} \peel{} starts from the 5 ERM models of the coverage sweep. The \emph{head} variant trains a linear head (initialized from ERM) with the coverage objective for 30 epochs (SGD, learning rate $0.01$) on the frozen ERM backbone. The \emph{no-audit} variant lets accepted core swaps into the bank, implemented as a core-swap fraction $1/|\text{bank}|$ of edits. \Cref{tab:e2-peel} summarizes the variants and \cref{tab:e2-peel-log} gives the per-round log of \peel{}-full. All four shortcuts are found in $5/5$ seeds by the full, greedy and aug-only variants. Core swap was significant in every round of \peel-full ($\widehat\Delta$ $0.6$--$5.9$), but the audit rejected it every time ($\bar\delta=1.00$). In seeds 0, 2, 3 and 4, round 1 keeps all four shortcuts and round 2 stops. The weakest shortcut, $S_2$, has $\widehat\Delta=1.39$--$1.59$ against a largest null of $0.74$--$1.24$. Under the flip-rate test of our first runs, its flip rate ($0.008$--$0.012$ vs.\ null $0.003$--$0.007$) was not significant in seeds 0, 2 and 3, and those runs needed an extra round. The final banks are the same under both statistics in all 25 runs. In seed 1, where ERM barely uses $S_4$ ($\widehat\Delta_4=0.08$), $S_4$ is kept in round 2 ($\widehat\Delta_4$: $0.08\to2.97$); this is the Whac-a-Mole effect seen by the discovery statistic. In the greedy variant the top-$\widehat\Delta$ candidate is the top-$\hat\lambda$ candidate of the \emph{current} model in 14 of 15 rounds. The exception (seed 2, round 1) is a near-tie ($\hat\lambda_1=7.13$, $\hat\lambda_3=6.91$). Retraining can reshuffle saliences. In seed 1, $S_4$, barely used by ERM, is covered before $S_2$. WGA stays near $0$ until the last shortcut is covered (seed 0: $0.002\to0.020\to0.222\to0.745$). \peel-full's final model has $K=[4]$ and the same seed as the oracle cover, so their agreement is expected once discovery is complete; the remaining differences are GPU non-determinism.

\begin{table}[h!]
\centering
\caption{E2: \peel{} variants (5 seeds). All variants end with the modal bank $\{S_1,S_2,S_3,S_4\}$ (plus core swap for no-audit). Rounds include the final stopping round. Spearman correlations are with ERM salience.}
\label{tab:e2-peel}
\small
\begin{tabular}{lcccccc}
\toprule
variant & rounds & WGA & IID & core rel. & Spearman (order) & Spearman (round-1 $\widehat\Delta$) \\
\midrule
\peel-full & 2.2 & $71.3\pm4.0$ & $73.4\pm2.7$ & $71.6\pm4.2$ & 0.96 & 0.96 \\
\peel-greedy & 5.0 & $73.0\pm4.5$ & $74.5\pm3.9$ & $74.3\pm4.1$ & 0.88 & 0.96 \\
\peel-aug-only & 2.2 & $72.1\pm5.7$ & $74.0\pm5.6$ & $73.9\pm5.7$ & 0.96 & 0.96 \\
\peel-head & 2.0 & $19.0\pm16.9$ & $67.4\pm12.2$ & $39.0\pm14.4$ & 0.96 & 0.96 \\
\peel{} w/o audit & 2.2 & $60.8\pm1.8$ & $63.3\pm2.2$ & $64.3\pm2.3$ & 0.96 & 0.96 \\
\midrule
oracle $K=[4]$, $\delta=0$ & -- & $70.9\pm4.0$ & $73.0\pm2.5$ & $71.2\pm3.6$ & -- & -- \\
\bottomrule
\end{tabular}
\end{table}

\begin{table}[h!]
\centering
\caption{E2: per-round log of \peel-full. $\widehat\Delta$ is the mean absolute logit-margin change on accepted validation edits; ``bank'' marks types already covered. Every core-swap audit gave $50/50$ bad ($\bar\delta=1.00$); every kept shortcut gave $\bar\delta=0.058$.}
\label{tab:e2-peel-log}
\footnotesize
\setlength{\tabcolsep}{4pt}
\begin{tabular}{cccccccccc}
\toprule
seed & round & $\widehat\Delta_{S_1}$ & $\widehat\Delta_{S_2}$ & $\widehat\Delta_{S_3}$ & $\widehat\Delta_{S_4}$ & $\widehat\Delta_{\rm swap}$ & max null & kept & WGA after \\
\midrule
0 & 1 & 3.71 & 1.59 & 3.40 & 2.67 & 2.61 & 0.93 & $S_1$,$S_3$,$S_4$,$S_2$ & 0.704 \\
0 & 2 & bank & bank & bank & bank & 1.66 & 0.21 & stop & -- \\
1 & 1 & 4.88 & 2.90 & 4.56 & 0.08 & 5.89 & 0.93 & $S_1$,$S_3$,$S_2$ & 0.008 \\
1 & 2 & bank & bank & bank & 2.97 & 0.57 & 0.26 & $S_4$ & 0.653 \\
1 & 3 & bank & bank & bank & bank & 1.34 & 0.24 & stop & -- \\
2 & 1 & 3.37 & 1.42 & 3.47 & 2.62 & 2.68 & 1.24 & $S_3$,$S_1$,$S_4$,$S_2$ & 0.748 \\
2 & 2 & bank & bank & bank & bank & 2.37 & 0.29 & stop & -- \\
3 & 1 & 3.23 & 1.39 & 3.53 & 2.51 & 1.96 & 1.18 & $S_3$,$S_1$,$S_4$,$S_2$ & 0.709 \\
3 & 2 & bank & bank & bank & bank & 1.57 & 0.27 & stop & -- \\
4 & 1 & 3.26 & 1.54 & 3.11 & 1.14 & 2.18 & 0.74 & $S_1$,$S_3$,$S_2$,$S_4$ & 0.750 \\
4 & 2 & bank & bank & bank & bank & 2.45 & 0.33 & stop & -- \\
\bottomrule
\end{tabular}
\end{table}

\begin{figure}[h!]
\centering
\includegraphics[width=0.7\textwidth]{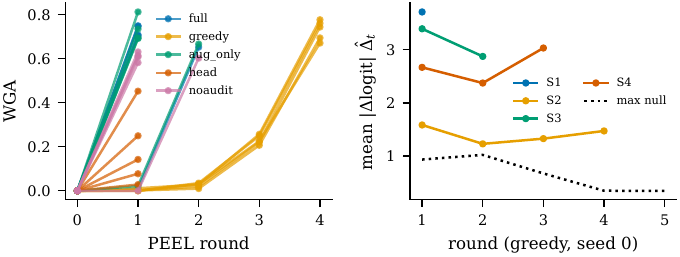}
\caption{E2 \peel. Left: WGA after each round for all variants and seeds. Right: $\widehat\Delta$ per round for the greedy variant, seed 0; the dotted line is the largest null-edit value.}
\label{fig:e2-peel}
\end{figure}

\paragraph{Baselines.} JTT$^\dagger$: identification for $T\in\{2,5\}$ epochs, upweighting $\lambda_{\rm up}\in\{20,50\}$, selected by group-labelled validation WGA. GroupDRO$^\dagger$: groups $(y,a_1,\dots,a_4)$; only groups present in training can be used, and with $\rho=0.95$ several minority groups (including the all-flipped one) are empty; $\eta\in\{0.01,0.1\}$ selected by validation WGA. DFR$^\dagger$: ERM features and an $\ell_1$-regularized logistic head trained on validation images re-rendered with uniformly random attributes (5 re-renders; an IID validation set has empty minority groups), $C\in\{0.01,0.1,1\}$ tuned on an image-disjoint validation half. Results are in \cref{tab:e2-base}. All group-label baselines fail because the worst flip groups have no or almost no training examples and the ERM representation carries no core information. GroupDRO$^\dagger$ is unstable across seeds (WGA from $0.8$ to $50.1$).

\begin{table}[h!]
\centering
\caption{E2: baselines (5 seeds). $\dagger$: uses group labels.}
\label{tab:e2-base}
\small
\begin{tabular}{lccc}
\toprule
method & WGA & IID & core rel.\ (aligned) \\
\midrule
ERM & $0.0\pm0.0$ & $99.2\pm0.4$ & $0.0\pm0.0$ \\
JTT$^\dagger$ & $0.2\pm0.2$ & $98.6\pm0.4$ & $0.2\pm0.2$ \\
GroupDRO$^\dagger$ & $34.4\pm19.8$ & $60.1\pm21.5$ & $40.4\pm22.9$ \\
DFR$^\dagger$ & $43.6\pm7.2$ & $63.6\pm5.6$ & $49.5\pm8.3$ \\
\peel-full (no group labels) & $71.3\pm4.0$ & $73.4\pm2.7$ & $71.6\pm4.2$ \\
oracle $K=[4]$, $\delta=0$ & $70.9\pm4.0$ & $73.0\pm2.5$ & $71.2\pm3.6$ \\
\bottomrule
\end{tabular}
\end{table}

\paragraph{Certificate.} Discovery uses one half of the validation set ($1{,}000$ images). That half selects the bank and hence the model, and its audit decides which types are kept. The certificate therefore uses only the other half: $n_{\rm v}=n_t=1000$, plus a fresh audit of $m=50$ accepted edits per type drawn on it ($0/50$ bad for every random-edit type). Here $\bar\delta_t$ is computed at level $\alpha/(3\tau)$: $0.104$ for $\tau=4$, $0.099$ for $\tau=3$ and $0.108$ for $\tau=5$. Measured error is on the IID test set edited by type $t$ (\cref{tab:e2-cert}). It is computed for the final models of the reported runs. It holds in all 99 (run, type) cases. The bound is dominated by the validation error (${\approx}28\%$; the core is hard). The remaining slack (${\approx}0.13$) comes mostly from the two Hoeffding terms (${\approx}0.10$). It certifies only the edit family (rand$_{S_1}$, \dots, rand$_{S_4}$), not flip groups.

\begin{table}[h!]
\centering
\caption{E2: certificate (\cref{thm:cert}) per (seed, bank type), as error rates, computed on the certificate half of the validation set.}
\label{tab:e2-cert}
\small
\begin{tabular}{lccccccc}
\toprule
variant & \#(seed, type) & holds & mean bound & mean meas. & mean gap & max bound & mean $\widehat{\err}_{\rm val}$ \\
\midrule
full & 20 & 100\% & 0.403 & 0.268 & 0.136 & 0.458 & 0.279 \\
greedy & 20 & 100\% & 0.386 & 0.254 & 0.132 & 0.454 & 0.264 \\
aug-only & 20 & 100\% & 0.390 & 0.260 & 0.130 & 0.475 & 0.261 \\
head & 19 & 100\% & 0.572 & 0.354 & 0.217 & 0.844 & 0.328 \\
no audit$^\ast$ & 20 & 100\% & 0.496 & 0.366 & 0.130 & 0.520 & 0.371 \\
\bottomrule
\end{tabular}
\\[2pt]
{\footnotesize $^\ast$Random-edit types only: for no-audit, the core-swap type has $\bar\delta=1$, so its bound is vacuous and it is excluded.}
\end{table}

\paragraph{Other deviations.} The corner patch overwrites the border in the top-left $4\times4$ corner (both remain exact). \peel-greedy and no-audit variants were added. Seeds: 5 (0--4) for the main coverage sweep, the $q=1$ $\delta$ sweep for $\{1,3\}$ and full coverage, \peel{} and baselines; 3 for $q=0.5$ replicates and $\{1,3,4\}$. Significance tests over 5 (or 3) seeds have low power. bf16 and cuDNN non-determinism make identical configurations differ slightly between runs.

\FloatBarrier
\subsection{E3: Waterbirds-2S}
\label{app:e3}

\paragraph{Dataset.} Birds: CUB-200-2011 \citep{wah2011caltech} (11{,}788 images) with the official segmentation masks. $y=1$ (waterbird) uses the species list of \citet{sagawa2020distributionally} (Albatross, Auklet, Cormorant, Frigatebird, Fulmar, Gull, Jaeger, Kittiwake, Pelican, Puffin, Tern, Gadwall, Grebe, Mallard, Merganser, Guillemot, Pacific Loon), giving 43 species and 2{,}483 waterbird images ($21.1\%$). \emph{Deviation}: we match species names at token boundaries (``Tern'' does not match ``Western\_*''). The original script uses substring matching, which also labels Western Meadowlark and Western Wood-Pewee as waterbirds; hence $21.1\%$ waterbirds instead of ${\approx}23\%$. The split is stratified by $y$ (seed 0): $4{,}795$ train, $1{,}199$ validation, $5{,}794$ test. Images and masks are resized to $256\times256$ and center-cropped to $224$, and masks are binarized at $0.5$. Backgrounds: Places365-Standard training images \citep{zhou2018places} (256\,px, center crop 224); land $=$ bamboo\_forest $+$ forest/broadleaf, water $=$ ocean $+$ lake/natural, with $5{,}000$ images per category ($10{,}000$ per type). Each category is split into disjoint pools: $50\%$ train, $15\%$ validation, $35\%$ test (train $5{,}000+5{,}000$, validation $1{,}500+1{,}500$, test $3{,}500+3{,}500$). Compositing ($x=m\cdot\text{bird}+(1-m)\cdot\text{bg}$, then watermark, then nuisance transforms) runs on the GPU. Birds are pasted at their original location and scale, as in Waterbirds, so thin halos of the original CUB background may remain at mask borders, a possible third, uncontrolled cue.

\paragraph{Shortcuts.} $S_{\rm bg}$: background type, $\rho_{\rm bg}=0.95$. $S_{\rm wm}$: the text ``\copyright{} PHOTOSTOCK'' (DejaVu Sans Bold, 20\,px, about $189\times21$\,px), composited in white at opacity $0.6$ with a 1-pixel dark offset shadow (strength $0.6$) at a uniformly random position; $\rho_{\rm wm}=0.90$, independent of $a_{\rm bg}$ given $y$. The training set is fixed (one composite per bird). The watermark was tuned with seed-0 ERM so that both shortcuts have clearly measurable salience (\cref{tab:e3-wm}). The initial faint watermark was used by ERM (it costs about 9 points on top of the background in group $\{$bg,wm$\}$) but only weakly on its own. Even after tuning, the watermark is the less salient shortcut ($\hat\lambda_{\rm wm}\approx0.6\,\hat\lambda_{\rm bg}$), and both are below the core ($\hat\lambda_{\rm core}\approx9$).

\begin{table}[h!]
\centering
\caption{E3: watermark tuning (seed-0 ERM, 20 epochs). The last row is used.}
\label{tab:e3-wm}
\small
\begin{tabular}{lccccccc}
\toprule
watermark & IID & $\varnothing$ & $\{$bg$\}$ & $\{$wm$\}$ & $\{$bg,wm$\}$ & $\hat\lambda_{\rm bg}$ & $\hat\lambda_{\rm wm}$ \\
\midrule
opacity 0.35, 18\,px, no shadow & 99.0 & 99.4 & 89.6 & 98.7 & 80.6 & 4.44 & 1.17 \\
opacity 0.5, 18\,px, shadow 0.5 & 99.0 & 99.6 & 91.7 & 97.9 & 72.8 & 4.25 & 2.20 \\
\textbf{opacity 0.6, 20\,px, shadow 0.6} & 98.9 & 99.6 & 91.9 & 97.4 & 67.3 & 4.28 & 2.57 \\
\bottomrule
\end{tabular}
\end{table}

\begin{figure}[h!]
\centering
\includegraphics[width=0.75\textwidth]{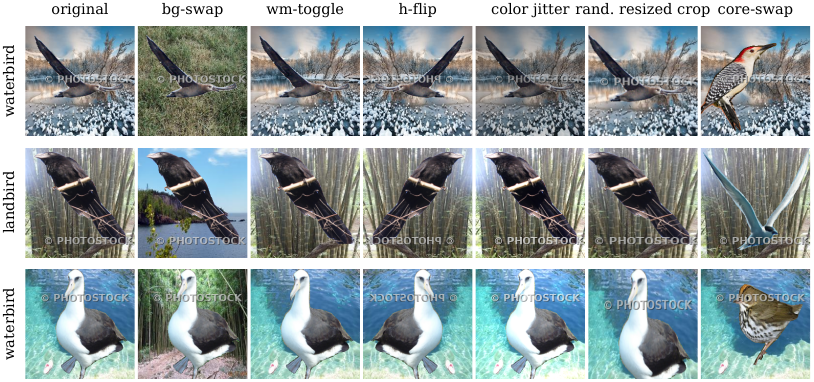}
\caption{E3: examples of the edit library on Waterbirds-2S composites.}
\label{fig:e3-examples}
\end{figure}

\paragraph{Evaluation sets and edits.} The IID validation set (training $\rho$) is used for \peel's $\widehat\Delta$ and for salience; the certificate uses held-out test birds (see \emph{Certificate}). A separate attribute-balanced validation composite ($a_{\rm bg},a_{\rm wm}$ uniform, independent of $y$) is used \emph{only} by the $\dagger$ baselines. Test: the 4 flip groups, each over all $5{,}794$ test birds with deterministic backgrounds and positions from the test pools; an IID test set; a core-swap test set (bird replaced by a random test bird of the other class, context aligned with the original label), on which ``core-follow'' is the fraction classified by the pasted bird; and the IID test set edited by each edit type (certificate targets). Edits (all exact): background swap ($a_{\rm bg}\sim U\{0,1\}$, fresh background from the split's pool), watermark toggle ($a_{\rm wm}\sim U\{0,1\}$, fresh position), null edits (horizontal flip; color jitter with brightness/contrast/saturation factors in $U[0.8,1.2]$; random resized crop with scale $U[0.8,1]$ and ratio $[3/4,4/3]$), and core swap (other-class bird pasted on the same background and watermark, label kept). Horizontal flip mirrors the watermark text, so it is not perfectly watermark-neutral. Examples are shown in \cref{fig:e3-examples}.

\begin{figure}[h!]
\centering
\includegraphics[width=0.55\textwidth]{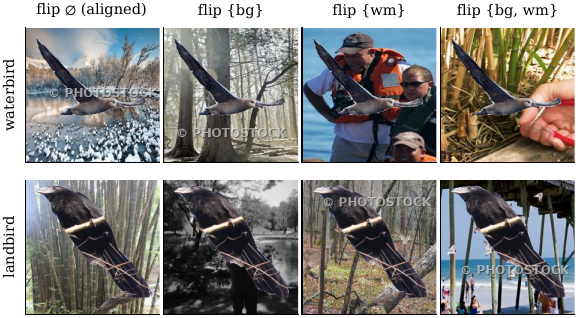}
\caption{E3: the four flip groups for one waterbird and one landbird. Every test bird appears in every group.}
\label{fig:e3-groups}
\end{figure}

\paragraph{Training.} ResNet-50 (torchvision IMAGENET1K\_V1 weights) with a new 2-way head, $224$\,px input, SGD (learning rate $10^{-3}$, momentum $0.9$, weight decay $10^{-4}$), batch 64, 20 epochs, cosine schedule, bf16 autocast, no standard augmentation (so null edits are genuine edits), last epoch. Interpolating regime: no weight decay, 60 epochs. \peel{} uses the library $\{$bg, wm, core swap$\}$ as candidates and $\{$h-flip, jitter, crop$\}$ as null edits, $\widehat\Delta$ on the first 1000 IID validation images, and the protocol of \cref{app:protocol}. \peel-head retrains only the last layer (initialized from the ERM head) on the frozen ERM backbone (batch-norm frozen), SGD with learning rate $10^{-2}$, 20 epochs, same objective. \emph{Random-edit} (equal-information baseline): every sample ($q=1$) gets one edit type drawn uniformly from the whole library $\{$bg, wm, h-flip, jitter, crop, core swap$\}$; core-swap proposals pass the validator with probability $0.1$ and rejected ones are redrawn among the 5 valid types (equal edit budget, no audit), so about $1.7\%$ of edited samples are label-corrupting. JTT$^\dagger$: 1-epoch ERM for identification, error set upweighted $\times50$, retraining for the same number of SGD steps as ERM, selection by worst-group accuracy on the attribute-balanced validation set (8 groups $y\times a_{\rm bg}\times a_{\rm wm}$). DFR$^\dagger$: ERM backbone features; $\ell_1$-regularized logistic regression on group-balanced subsamples (8 groups) of the attribute-balanced validation set, $C$ tuned on a half split by worst-group accuracy, final fit on all of it, weights averaged over 10 subsamples. GroupDRO$^\dagger$: 8 groups $(y,a_{\rm bg},a_{\rm wm})$, $\eta=0.01$, selection by validation worst-group accuracy. No hyper-parameter search was done for any method.

\paragraph{Main results.} \Cref{tab:e3-main} extends \cref{tab:main} (Waterbirds-2S column) with all flip groups and core-follow. Because the discovered bank equals $\{$bg,wm$\}$ and the Learn step is deterministic given the seed, \peel-full, \peel-greedy and the oracle cover yield the same final models. WGA is the mean over seeds of the per-seed minimum over groups, so it can lie below every group mean. Random-edit improves WGA by $9.8$ points over ERM but lowers core-follow ($45.8$ vs.\ $58.8$).

\begin{table}[h!]
\centering
\caption{E3: all methods (ResNet-50), mean$\pm$std over 5 seeds. $\dagger$: uses group labels. Core-follow: accuracy w.r.t.\ the pasted bird on the core-swap test set.}
\label{tab:e3-main}
\footnotesize
\setlength{\tabcolsep}{3pt}
\begin{tabular}{llccccccc}
\toprule
method & labels & IID & $\varnothing$ & $\{$bg$\}$ & $\{$wm$\}$ & $\{$bg,wm$\}$ & WGA & core-fol. \\
\midrule
ERM & -- & $98.9\pm0.1$ & $99.6\pm0.0$ & $91.9\pm0.3$ & $97.3\pm0.1$ & $66.5\pm1.3$ & $66.5\pm1.3$ & $58.8\pm2.4$ \\
Random-edit & -- & $99.3\pm0.1$ & $99.7\pm0.1$ & $97.1\pm0.3$ & $97.1\pm0.1$ & $76.3\pm1.2$ & $76.3\pm1.2$ & $45.8\pm6.1$ \\
\peel-full & -- & $98.5\pm0.1$ & $98.3\pm0.1$ & $98.6\pm0.1$ & $98.3\pm0.1$ & $98.3\pm0.1$ & $\mathbf{98.2\pm0.1}$ & $97.0\pm0.3$ \\
\peel-greedy & -- & $98.5\pm0.1$ & $98.3\pm0.1$ & $98.6\pm0.1$ & $98.3\pm0.1$ & $98.3\pm0.1$ & $\mathbf{98.2\pm0.1}$ & $97.0\pm0.3$ \\
\peel-head & -- & $97.3\pm0.1$ & $97.4\pm0.2$ & $96.5\pm0.1$ & $97.2\pm0.1$ & $95.7\pm0.3$ & $95.7\pm0.3$ & $91.5\pm0.3$ \\
\peel{} w/o audit & -- & $86.1\pm0.9$ & $85.6\pm0.5$ & $88.8\pm0.7$ & $87.4\pm0.7$ & $91.3\pm0.7$ & $85.6\pm0.5$ & $66.7\pm1.9$ \\
Oracle $K=\{$bg,wm$\}$ & (shortcuts) & $98.5\pm0.1$ & $98.3\pm0.1$ & $98.6\pm0.1$ & $98.4\pm0.1$ & $98.3\pm0.1$ & $98.2\pm0.1$ & $97.0\pm0.3$ \\
JTT$^\dagger$ & val & $96.9\pm0.4$ & $97.8\pm0.4$ & $97.0\pm0.7$ & $86.9\pm1.0$ & $83.1\pm3.8$ & $83.1\pm3.8$ & $82.8\pm1.6$ \\
DFR$^\dagger$ & val (bal.) & $96.6\pm0.1$ & $96.8\pm0.1$ & $95.0\pm0.2$ & $93.6\pm0.3$ & $89.6\pm0.6$ & $89.6\pm0.6$ & $92.9\pm0.4$ \\
GroupDRO$^\dagger$ & train+val & $98.2\pm0.4$ & $98.6\pm0.3$ & $93.1\pm0.5$ & $96.7\pm0.5$ & $85.6\pm1.4$ & $85.6\pm1.4$ & $86.0\pm2.1$ \\
\bottomrule
\end{tabular}
\end{table}

\begin{figure}[h!]
\centering
\includegraphics[width=0.55\textwidth]{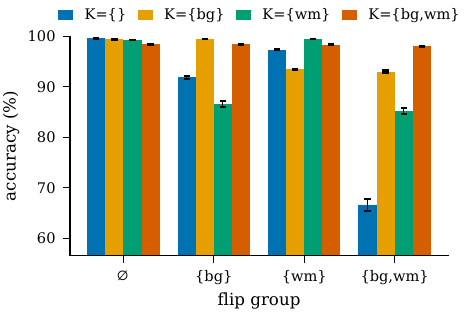}
\caption{E3, $q=0.5$: flip-group accuracy for each covered set (5 seeds). The Whac-a-Mole pattern of \cref{fig:e3}(a) persists.}
\label{fig:e3-q05}
\end{figure}

\paragraph{Whac-a-Mole.} \Cref{tab:e3-wam} gives all covers and the salience of each shortcut; \cref{tab:e3-trans} gives the per-seed paired changes for every coverage transition. A cell ``changes'' if $|\Delta|\ge1$ point with the same sign in all seeds. At $q=1$, 12 of 16 cells change and all 12 have the predicted sign; the 4 non-changing cells are all ``predicted $\downarrow$'' cells and are also negative ($-0.3$ to $-0.9$). The same holds at $q=0.5$ (12/12; \cref{fig:e3-q05}), and $q=1$ vs.\ $q=0.5$ makes little difference on real images, unlike in E1. After coverage, the \emph{signed} salience of a covered shortcut is $\approx0$; the absolute $|\Delta\text{logit}|$ stays at $1.1$ (bg) and $0.6$ (wm) because it is computed between two specific background images or watermark positions and includes image-specific variation. The uncovered shortcut's salience rises once the other is covered ($\hat\lambda_{\rm wm}$: $2.6\to3.3$; $\hat\lambda_{\rm bg}$: $4.4\to5.0$). \emph{Caveat}: plugging ERM saliences into the decision rule ($\sal_c\approx\hat\lambda_{\rm core}/2\approx4.7>\sal_{\rm bg}+\sal_{\rm wm}\approx3.5$) leaves every group margin positive at the population level, so the rule with mean saliences predicts no population-level flips. The observed changes are in the per-example tail, and sign agreement, not magnitude, is the testable prediction here.

\begin{table}[h!]
\centering
\caption{E3: coverage subsets (ResNet-50, 5 seeds). Saliences are given as absolute / signed mean logit change on validation.}
\label{tab:e3-wam}
\footnotesize
\setlength{\tabcolsep}{3.5pt}
\begin{tabular}{clccccccccc}
\toprule
$q$ & $K$ & IID & $\varnothing$ & $\{$bg$\}$ & $\{$wm$\}$ & $\{$bg,wm$\}$ & WGA & $\hat\lambda_{\rm bg}$ & $\hat\lambda_{\rm wm}$ & $\hat\lambda_{\rm core}$ \\
\midrule
-- & $\varnothing$ (ERM) & $98.9\pm0.1$ & $99.6\pm0.0$ & $91.9\pm0.3$ & $97.3\pm0.1$ & $66.5\pm1.3$ & $66.5\pm1.3$ & 4.4 / 4.2 & 2.6 / 2.6 & 9.3 \\
\midrule
1 & $\{$bg$\}$ & $98.9\pm0.0$ & $99.3\pm0.0$ & $99.5\pm0.0$ & $93.5\pm0.3$ & $93.5\pm0.4$ & $93.4\pm0.4$ & 1.1 / $-0.1$ & 3.3 / 3.3 & 12.6 \\
1 & $\{$wm$\}$ & $98.8\pm0.0$ & $99.2\pm0.1$ & $86.2\pm0.3$ & $99.5\pm0.1$ & $85.8\pm0.4$ & $85.8\pm0.4$ & 5.0 / 4.8 & 0.7 / $-0.2$ & 11.2 \\
1 & $\{$bg,wm$\}$ & $98.5\pm0.1$ & $98.3\pm0.1$ & $98.6\pm0.1$ & $98.4\pm0.1$ & $98.3\pm0.1$ & $98.2\pm0.1$ & 1.1 / $-0.1$ & 0.6 / $-0.0$ & 14.1 \\
\midrule
0.5 & $\{$bg$\}$ & $99.0\pm0.0$ & $99.4\pm0.0$ & $99.5\pm0.0$ & $93.5\pm0.1$ & $93.0\pm0.2$ & $93.0\pm0.2$ & 1.2 / $-0.0$ & 3.4 / 3.4 & 12.4 \\
0.5 & $\{$wm$\}$ & $98.8\pm0.0$ & $99.3\pm0.0$ & $86.6\pm0.6$ & $99.5\pm0.0$ & $85.2\pm0.7$ & $85.2\pm0.7$ & 4.9 / 4.8 & 0.7 / $-0.0$ & 11.2 \\
0.5 & $\{$bg,wm$\}$ & $98.5\pm0.1$ & $98.4\pm0.1$ & $98.5\pm0.1$ & $98.3\pm0.1$ & $98.0\pm0.1$ & $98.0\pm0.1$ & 1.1 / 0.0 & 0.6 / 0.1 & 13.8 \\
\bottomrule
\end{tabular}
\end{table}

\begin{table}[h!]
\centering
\caption{E3: change in accuracy (points, mean$\pm$std of per-seed paired differences) per coverage transition. Prediction: covering $j$ lowers the margin of every group $F\not\ni j$ ($\downarrow$) and raises it for $F\ni j$ ($\uparrow$). $^\ast$: $|\Delta|\ge1$ with the same sign in all seeds; every such cell has the predicted sign.}
\label{tab:e3-trans}
\footnotesize
\setlength{\tabcolsep}{4pt}
\begin{tabular}{clccccc}
\toprule
$q$ & transition & $\varnothing$ & $\{$bg$\}$ & $\{$wm$\}$ & $\{$bg,wm$\}$ & $\Delta$WGA \\
\midrule
1 & $\varnothing\to\{$bg$\}$ & $-0.3\pm0.1$ ($\downarrow$) & $+7.6\pm0.3^\ast$ ($\uparrow$) & $-3.8\pm0.3^\ast$ ($\downarrow$) & $+27.0\pm1.1^\ast$ ($\uparrow$) & $+26.9\pm1.1$ \\
1 & $\varnothing\to\{$wm$\}$ & $-0.4\pm0.1$ ($\downarrow$) & $-5.7\pm0.4^\ast$ ($\downarrow$) & $+2.2\pm0.1^\ast$ ($\uparrow$) & $+19.3\pm1.1^\ast$ ($\uparrow$) & $+19.3\pm1.1$ \\
1 & $\{$wm$\}\to\{$bg,wm$\}$ & $-0.9\pm0.1$ ($\downarrow$) & $+12.4\pm0.2^\ast$ ($\uparrow$) & $-1.1\pm0.1^\ast$ ($\downarrow$) & $+12.5\pm0.3^\ast$ ($\uparrow$) & $+12.4\pm0.3$ \\
1 & $\{$bg$\}\to\{$bg,wm$\}$ & $-1.0\pm0.1^\ast$ ($\downarrow$) & $-0.9\pm0.1$ ($\downarrow$) & $+4.8\pm0.2^\ast$ ($\uparrow$) & $+4.8\pm0.3^\ast$ ($\uparrow$) & $+4.8\pm0.3$ \\
\midrule
0.5 & $\varnothing\to\{$bg$\}$ & $-0.2\pm0.0$ ($\downarrow$) & $+7.6\pm0.3^\ast$ ($\uparrow$) & $-3.8\pm0.1^\ast$ ($\downarrow$) & $+26.5\pm1.2^\ast$ ($\uparrow$) & $+26.5\pm1.2$ \\
0.5 & $\varnothing\to\{$wm$\}$ & $-0.3\pm0.1$ ($\downarrow$) & $-5.3\pm0.4^\ast$ ($\downarrow$) & $+2.1\pm0.1^\ast$ ($\uparrow$) & $+18.7\pm1.1^\ast$ ($\uparrow$) & $+18.7\pm1.1$ \\
0.5 & $\{$wm$\}\to\{$bg,wm$\}$ & $-0.9\pm0.1$ ($\downarrow$) & $+11.9\pm0.5^\ast$ ($\uparrow$) & $-1.2\pm0.1^\ast$ ($\downarrow$) & $+12.8\pm0.7^\ast$ ($\uparrow$) & $+12.8\pm0.7$ \\
0.5 & $\{$bg$\}\to\{$bg,wm$\}$ & $-1.0\pm0.1$ ($\downarrow$) & $-1.0\pm0.1^\ast$ ($\downarrow$) & $+4.8\pm0.1^\ast$ ($\uparrow$) & $+5.0\pm0.3^\ast$ ($\uparrow$) & $+5.0\pm0.3$ \\
\bottomrule
\end{tabular}
\end{table}

\paragraph{Validator error.} \Cref{tab:e3-delta} reports the $\delta$ sweep at $q=1$ with and without the consistency term, and at $q=0.5$. Qualitatively the predictions of \cref{thm:validator} hold: partial coverage is fragile, interpolation hurts under partial coverage (with $K=\{$bg$\}$: $47.2$ vs.\ $63.4$ at $\delta=0.05$, $25.5$ vs.\ $45.7$ at $\delta=0.1$) but not under full coverage ($97.8$ vs.\ $96.8$ at $\delta=0.05$). The pre-registered balanced-class tolerances fail quantitatively: $\delta^\star=0.10$ for $K=\{$bg$\}$ ($R_U=\lambda(0.9)$) and $0.5$ for full coverage, whereas $K=\{$bg$\}$ already fails at $\delta=0.05$, and full coverage collapses to the constant ``landbird'' classifier at $\delta\ge0.2$ with the KL term ($\ge0.3$ for aug-only): WGA $78.9$ equals the landbird share and core-follow $21.1$ the waterbird share. Including the class prior $\pi=0.211$ (\cref{rem:prior}, derived after observing this failure) with $p\approx1$ gives $\delta^\star_\pi\approx0.03$ ($K=\{$bg$\}$) and $\approx\pi\approx0.21$ (full coverage), which matches where the curves break. The consistency term amplifies false accepts: aug-only is clearly more tolerant ($96.9$ vs.\ $85.9$ at $\delta=0.1$ and $86.6$ vs.\ $78.9$ at $\delta=0.2$ for full coverage; $60.5$ vs.\ $45.7$ at $\delta=0.1$ for $K=\{$bg$\}$). An interpolating ERM (no weight decay, 60 epochs) reaches WGA $68.3\pm1.2$, core-follow $62.5\pm0.5$ and IID $98.9\pm0.1$ (3 seeds).

\begin{table}[h!]
\centering
\caption{E3: WGA / core-follow vs.\ injected false-accept rate $\delta$. Seeds: 5 for the regularized cells with KL at $q=1$ and for the $\delta=0$ regularized cells at $q=0.5$; 3 otherwise.}
\label{tab:e3-delta}
\footnotesize
\setlength{\tabcolsep}{4pt}
\begin{tabular}{cccccc}
\toprule
& $\delta$ & reg., $K=\{$bg$\}$ & reg., $K=\{$bg,wm$\}$ & interp., $K=\{$bg$\}$ & interp., $K=\{$bg,wm$\}$ \\
\midrule
\multirow{5}{*}{\shortstack{$q=1$\\with KL}}
& 0 & $93.4\pm0.4$ / $90.0\pm0.4$ & $98.2\pm0.1$ / $97.0\pm0.3$ & $94.9\pm0.2$ / $91.9\pm0.5$ & $98.5\pm0.1$ / $97.5\pm0.2$ \\
& 0.05 & $63.4\pm1.7$ / $17.7\pm0.3$ & $96.8\pm0.3$ / $91.5\pm1.0$ & $47.2\pm1.3$ / $13.4\pm0.4$ & $97.8\pm0.1$ / $94.5\pm0.1$ \\
& 0.1 & $45.7\pm2.3$ / $13.2\pm0.9$ & $85.9\pm1.3$ / $68.4\pm4.3$ & $25.5\pm2.7$ / $7.8\pm0.7$ & $84.8\pm0.6$ / $74.6\pm0.9$ \\
& 0.2 & $16.5\pm4.3$ / $5.8\pm1.4$ & $78.9\pm0.0$ / $21.1\pm0.0$ & $10.4\pm1.1$ / $3.8\pm0.3$ & $78.9\pm0.0$ / $21.1\pm0.0$ \\
& 0.3 & $2.0\pm0.9$ / $1.0\pm0.3$ & $78.9\pm0.0$ / $21.1\pm0.0$ & $3.1\pm0.5$ / $1.3\pm0.1$ & $78.9\pm0.0$ / $21.1\pm0.0$ \\
\midrule
\multirow{5}{*}{\shortstack{$q=1$\\aug-only}}
& 0 & $93.0\pm0.5$ / $90.7\pm0.7$ & $98.0\pm0.1$ / $97.2\pm0.1$ & $94.5\pm0.1$ / $91.9\pm0.2$ & $98.3\pm0.1$ / $97.6\pm0.1$ \\
& 0.05 & $69.6\pm1.3$ / $30.9\pm0.3$ & $97.8\pm0.1$ / $94.9\pm0.2$ & $55.6\pm1.3$ / $17.9\pm0.6$ & $98.2\pm0.1$ / $96.8\pm0.1$ \\
& 0.1 & $60.5\pm1.6$ / $18.0\pm0.7$ & $96.9\pm0.3$ / $92.1\pm1.0$ & $31.9\pm3.6$ / $10.3\pm0.4$ & $97.6\pm0.2$ / $94.6\pm0.3$ \\
& 0.2 & $42.2\pm6.1$ / $12.1\pm1.2$ & $86.6\pm0.3$ / $62.8\pm0.5$ & $18.3\pm0.8$ / $5.9\pm0.4$ & $87.2\pm1.1$ / $64.1\pm3.7$ \\
& 0.3 & $19.4\pm5.4$ / $5.8\pm1.5$ & $79.0\pm0.0$ / $21.7\pm0.3$ & $13.3\pm0.2$ / $4.0\pm0.2$ & $78.9\pm0.0$ / $21.1\pm0.0$ \\
\midrule
\multirow{5}{*}{\shortstack{$q=0.5$\\with KL}}
& 0 & $93.0\pm0.2$ / $89.2\pm0.8$ & $98.0\pm0.1$ / $96.7\pm0.3$ & $94.2\pm0.7$ / $91.1\pm0.4$ & $98.4\pm0.1$ / $97.3\pm0.2$ \\
& 0.05 & $63.3\pm1.4$ / $18.2\pm0.1$ & $97.1\pm0.1$ / $91.2\pm0.4$ & $53.5\pm2.9$ / $14.9\pm0.4$ & $97.8\pm0.1$ / $93.3\pm0.2$ \\
& 0.1 & $51.9\pm3.3$ / $14.2\pm0.8$ & $91.5\pm0.9$ / $68.0\pm3.8$ & $34.0\pm2.3$ / $9.7\pm0.5$ & $93.3\pm0.9$ / $75.1\pm4.2$ \\
& 0.2 & $22.2\pm6.1$ / $6.2\pm1.6$ & $79.0\pm0.0$ / $21.3\pm0.1$ & $15.3\pm1.0$ / $4.3\pm0.3$ & $78.9\pm0.0$ / $21.1\pm0.0$ \\
& 0.3 & $10.6\pm1.4$ / $2.7\pm0.3$ & $78.9\pm0.0$ / $21.1\pm0.0$ & $6.0\pm0.2$ / $1.5\pm0.0$ & $78.9\pm0.0$ / $21.1\pm0.0$ \\
\bottomrule
\end{tabular}
\end{table}

\paragraph{CLIP ViT-B/16.} We repeat the coverage subsets with OpenAI CLIP ViT-B/16 weights loaded through timm, either fine-tuned (AdamW, learning rate $10^{-5}$, weight decay $0.1$, 10 epochs) or linearly probed (frozen backbone; the head is trained with the coverage objective, learning rate $10^{-3}$, 20 epochs), $q=1$, 3 seeds (\cref{tab:e3-clip}). In both variants every coverage transition raises WGA, and in all 32 (transition, group) cells the mean change has the predicted sign, although some changes are small (e.g.\ $0.1$ points). The Whac-a-Mole decreases are clear: covering the background lowers $\{$wm$\}$ from $98.4$ to $94.8$ (fine-tuned) and from $96.2$ to $89.5$ (linear probe); covering the watermark lowers $\{$bg$\}$ from $96.2$ to $90.1$ and from $94.5$ to $84.2$.

\begin{table}[h!]
\centering
\caption{E3: CLIP ViT-B/16 coverage subsets, $q=1$, 3 seeds.}
\label{tab:e3-clip}
\footnotesize
\setlength{\tabcolsep}{4pt}
\begin{tabular}{llcccccccc}
\toprule
variant & $K$ & IID & $\varnothing$ & $\{$bg$\}$ & $\{$wm$\}$ & $\{$bg,wm$\}$ & WGA & $\hat\lambda_{\rm bg}$ & $\hat\lambda_{\rm wm}$ \\
\midrule
fine-tune & $\varnothing$ & $99.4\pm0.1$ & $99.8\pm0.0$ & $96.2\pm0.9$ & $98.4\pm0.2$ & $81.4\pm1.0$ & $81.4\pm1.0$ & 2.6 & 2.6 \\
fine-tune & $\{$bg$\}$ & $99.3\pm0.0$ & $99.7\pm0.1$ & $99.7\pm0.1$ & $94.8\pm0.3$ & $94.2\pm0.3$ & $94.2\pm0.3$ & 0.4 & 4.0 \\
fine-tune & $\{$wm$\}$ & $99.2\pm0.1$ & $99.6\pm0.0$ & $90.1\pm1.4$ & $99.7\pm0.1$ & $88.9\pm1.4$ & $88.9\pm1.4$ & 4.1 & 0.2 \\
fine-tune & $\{$bg,wm$\}$ & $99.0\pm0.1$ & $99.0\pm0.0$ & $99.0\pm0.1$ & $99.0\pm0.1$ & $98.8\pm0.1$ & $98.8\pm0.1$ & 0.4 & 0.3 \\
\midrule
linear probe & $\varnothing$ & $99.1\pm0.0$ & $99.8\pm0.0$ & $94.5\pm0.1$ & $96.2\pm0.1$ & $68.0\pm0.7$ & $68.0\pm0.7$ & 5.5 & 5.6 \\
linear probe & $\{$bg$\}$ & $98.8\pm0.0$ & $99.7\pm0.0$ & $99.4\pm0.0$ & $89.5\pm0.2$ & $86.5\pm0.2$ & $86.5\pm0.2$ & 1.5 & 4.9 \\
linear probe & $\{$wm$\}$ & $98.7\pm0.1$ & $99.4\pm0.0$ & $84.2\pm0.2$ & $99.5\pm0.1$ & $82.5\pm0.2$ & $82.5\pm0.2$ & 5.6 & 1.0 \\
linear probe & $\{$bg,wm$\}$ & $98.0\pm0.0$ & $98.2\pm0.0$ & $97.4\pm0.1$ & $97.5\pm0.0$ & $96.6\pm0.0$ & $96.6\pm0.0$ & 1.4 & 0.8 \\
\bottomrule
\end{tabular}
\end{table}

\paragraph{\peel{} discovery.} \Cref{tab:e3-peel} shows the log for seed 0. The other seeds behave identically. In round 1, background ($\widehat\Delta=3.00$--$3.19$), watermark ($1.46$--$1.60$) and core swap ($8.7$--$9.5$) are significant against a largest null of $0.86$--$0.91$. The audit drops core swap ($50/50$ bad, $\bar\delta=1.00$) and keeps the other two ($0/50$, $\bar\delta=0.058$). In round 2 only core swap is significant ($\widehat\Delta=13.8$--$14.9$), and it is dropped again, so the loop stops with bank $\{$bg,wm$\}$ in all 5 seeds. \peel-greedy keeps background in round 1 and watermark in round 2 in all 5 seeds (decreasing salience, \cref{prop:greedy}). The watermark's $\widehat\Delta$ rises once the background is covered (round 1 $\to$ round 2: $1.54\to1.68$, $1.56\to1.74$, $1.60\to1.86$, $1.46\to1.80$, $1.50\to1.87$ for seeds 0--4); this is the Whac-a-Mole effect seen by the discovery statistic. \peel-head finds the same bank in every seed. Without the audit, core swap enters the bank in round 1, round 2 finds nothing new, and WGA falls to $85.6\pm0.5$. Every variant returns the same bank in every seed as the flip-rate test of our first runs, and WGA agrees within $0.25$ points.

\begin{table}[h!]
\centering
\caption{E3: \peel{} discovery log, seed 0. Mean absolute logit-margin change $\widehat\Delta$ on validator-accepted validation edits ($n\approx1000$; core swap $n\approx100$).}
\label{tab:e3-peel}
\footnotesize
\setlength{\tabcolsep}{4pt}
\begin{tabular}{llcccccccc}
\toprule
variant & round & bg & wm & core swap & h-flip & jitter & crop & significant (Holm) & kept \\
\midrule
\peel-full & 1 & 3.04 & 1.54 & 9.39 & 0.85 & 0.36 & 0.86 & core swap, bg, wm & bg, wm \\
\peel-full & 2 & bank & bank & 14.42 & 0.92 & 0.30 & 0.80 & core swap & -- (stop) \\
\midrule
\peel-greedy & 1 & 3.04 & 1.54 & 9.39 & 0.85 & 0.36 & 0.86 & core swap, bg, wm & bg \\
\peel-greedy & 2 & bank & 1.68 & 12.51 & 0.82 & 0.30 & 0.77 & core swap, wm & wm \\
\peel-greedy & 3 & bank & bank & 14.03 & 0.92 & 0.31 & 0.78 & core swap & -- (stop) \\
\bottomrule
\end{tabular}
\end{table}

\paragraph{Certificate.} Discovery uses the IID validation set. It selects the bank and, through it, the trained model, and its audit decides which types are kept. \Cref{thm:cert} needs samples that are independent of both, so the certificate reuses neither. It is computed on a fixed random half of the test birds ($2{,}897$ birds; the split seed was fixed before computing it), which no training, discovery or selection step uses. There, $n_{\rm v}=2{,}897$ IID test composites, $n_t=2{,}897$ (one fresh edit of each composite per bank type; the simulated validator accepts every background and watermark edit), and a fresh audit of $m_t=50$ accepted edits per type ($0/50$ bad) gives $\bar\delta_t=0.091$ at level $\alpha/(3\tau)$, with $\tau=|\text{bank}|=2$ and $\alpha=0.05$. The bound is checked on the other $2{,}897$ birds, whose IID test composites are edited by each type. It is computed for the final models of the reported runs. \Cref{tab:e3-cert} lists the bound and the measured error for \peel-full, \peel-head and, as a post-hoc contrast, ERM with the same bank. The bound holds in all 20 (\peel{} model, edit type) cases and in all 10 ERM cases. The uniform bound over $\mathcal Q$ (the maximum over types) also holds against the largest measured error among $P$, $Q_{\rm bg}$ and $Q_{\rm wm}$ in every case. The bound is non-vacuous but loose: $7.8$--$8.9\%$ vs.\ a measured $1.6$--$1.8\%$ for \peel-full, and $10.2$--$12.3\%$ vs.\ $2.3$--$3.3\%$ for \peel-head. Most of the slack comes from the two Hoeffding terms ($2.7$ and $2.9$ points at $n=2{,}897$) and from the factor $1/(1-\bar\delta_t)$ with $\bar\delta_t=0.091$ from $m=50$. The edited test distributions (random resampling of one type) are much milder than the flip groups, so the certificate says nothing about WGA. \Cref{thm:cert} covers mixtures of single edits, not the composition bg$\circ$wm, and E1 gives a counterexample to summing per-type terms (\cref{app:e1}). We therefore report no bound for the composition.

\begin{table}[h!]
\centering
\caption{E3: \cref{thm:cert} bound vs.\ measured error (\%). The certificate ($\widehat{\err}_P$, $\widehat\Gamma^{\rm acc}$, audit) uses one half of the test birds, and the measured error is on the IID test composites of the other half, edited by each bank type. ERM: post-hoc contrast with the same bank.}
\label{tab:e3-cert}
\footnotesize
\setlength{\tabcolsep}{4pt}
\begin{tabular}{llccccccc}
\toprule
& & & \multicolumn{3}{c}{background swap} & \multicolumn{3}{c}{watermark toggle} \\
\cmidrule(lr){4-6}\cmidrule(lr){7-9}
model & seed & $\widehat{\err}_P$ & $\widehat\Gamma^{\rm acc}$ & bound & measured & $\widehat\Gamma^{\rm acc}$ & bound & measured \\
\midrule
\peel-full & 0 & 1.4 & 1.2 & 8.6 & 1.6 & 0.5 & 7.8 & 1.6 \\
 & 1 & 1.4 & 1.4 & 8.9 & 1.7 & 0.6 & 8.0 & 1.6 \\
 & 2 & 1.5 & 0.8 & 8.3 & 1.6 & 0.4 & 7.8 & 1.6 \\
 & 3 & 1.6 & 1.1 & 8.6 & 1.6 & 0.6 & 8.1 & 1.8 \\
 & 4 & 1.7 & 0.9 & 8.5 & 1.7 & 0.5 & 8.1 & 1.6 \\
\midrule
\peel-head & 0 & 2.8 & 3.2 & 12.2 & 3.1 & 1.7 & 10.5 & 2.5 \\
 & 1 & 2.6 & 3.3 & 12.1 & 2.9 & 1.6 & 10.2 & 2.9 \\
 & 2 & 2.8 & 3.1 & 12.0 & 3.0 & 1.9 & 10.6 & 2.3 \\
 & 3 & 2.9 & 2.9 & 11.9 & 2.9 & 2.0 & 10.9 & 2.3 \\
 & 4 & 2.9 & 3.2 & 12.3 & 3.3 & 1.7 & 10.7 & 2.8 \\
\midrule
ERM & 0 & 1.2 & 6.2 & 13.9 & 5.8 & 2.1 & 9.3 & 2.5 \\
 & 1 & 1.4 & 7.0 & 14.9 & 6.0 & 1.4 & 8.8 & 2.9 \\
 & 2 & 1.3 & 6.2 & 14.0 & 5.8 & 2.7 & 10.1 & 2.8 \\
 & 3 & 1.3 & 5.9 & 13.6 & 5.5 & 2.0 & 9.3 & 2.6 \\
 & 4 & 1.3 & 6.0 & 13.8 & 5.7 & 2.1 & 9.5 & 2.8 \\
\bottomrule
\end{tabular}
\end{table}

\begin{figure}[h!]
\centering
\includegraphics[width=0.45\textwidth]{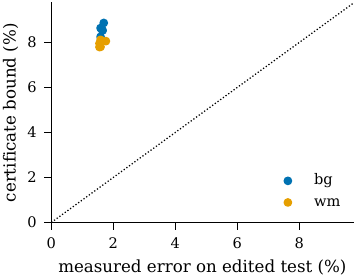}
\caption{E3: certificate vs.\ measured error for the final \peel-full models (5 seeds, both bank types). The certificate is computed on one half of the test birds and checked on the other.}
\label{fig:e3-cert}
\end{figure}

\paragraph{Limitations of E3.} All edits are exact compositing, so validator errors arise only by injection, and the validator is simulated. We did not run a natural validator, generative edits, a generic diffusion-augmentation baseline, or \peel{} and the baselines on CLIP.

\FloatBarrier
\subsection{Licenses and assets}
\label{app:licenses}

CIFAR-10 \citep{krizhevsky2009learning} is distributed by its authors for research use. CUB-200-2011 \citep{wah2011caltech} and its segmentation masks are provided for non-commercial research and educational use. Places365 \citep{zhou2018places} images are provided for research and non-commercial use. Pretrained ResNet-50 weights come from torchvision (BSD-3-Clause license). CLIP ViT-B/16 weights are OpenAI's release (MIT license), loaded through timm (Apache-2.0). We do not redistribute any images; our code renders all composites from the original datasets.

\end{document}